\documentclass[runningheads]{llncs}
\pdfoutput=1   

\usepackage{eccv}

\usepackage{eccvabbrv}
\usepackage{graphicx}
\usepackage{subcaption}
\usepackage{multirow}
\usepackage{gensymb}
\usepackage{xcolor,colortbl}

\usepackage{algorithm}
\usepackage{algpseudocode}
\usepackage{caption}
\usepackage{xspace}
\usepackage{pifont} 
\newcommand{\cmark}{\ding{51}}%
\newcommand{\xmark}{\ding{55}}%

\definecolor{green}{rgb}{0, 1, 0}
\definecolor{red}{rgb}{1,0,0}
\definecolor{yellow}{rgb}{1,1,0}
\colorlet{tabfirst}{green!45}
\colorlet{tabsecond}{red!15}
\colorlet{tabthird}{yellow!30}

\usepackage{booktabs}
\usepackage{wrapfig}

\usepackage[accsupp]{axessibility}  

\usepackage{hyperref}

\usepackage{orcidlink}

\makeatletter
\AtBeginDocument{\let\caption@xlabel\label}
\makeatother

\makeatletter
\newcommand{\maketitlesupplementary}{%
  \begingroup
  \let\oldtitle\@title
  \gdef\@title{\oldtitle\\[0.5ex] Supplementary Material}%
  \maketitle
  \endgroup
}
\makeatother
\begin{document}


\title{Off the Planckian Locus: Using 2D Chromaticity to Improve In-Camera Color} 

\titlerunning{Off the Planckian Locus}

\author{SaiKiran Tedla*~\inst{1,3} \and
Joshua Little*~\inst{1} \\
Hakki C. Karaimer\inst{2} \and
Michael S. Brown\inst{1}}

\authorrunning{S.~Tedla et al.}

\institute{$^{1}$York University \enspace
$^{2}$AI Center-Toronto, Samsung Electronics \enspace
$^{3}$University of Toronto\\
\email{\{tedlasai, jelittle, mbrown\}@yorku.ca} \enspace
\email{\{hakki.k\}@samsung.ca}
}
\maketitle

\renewcommand{\thefootnote}{\fnsymbol{footnote}}
\footnotetext[1]{Equal contribution.}
\renewcommand{\thefootnote}{\arabic{footnote}}

\begin{abstract}
Traditional in-camera colorimetric mapping relies on correlated color temperature (CCT)–based interpolation between pre-calibrated transforms optimized for Planckian illuminants such as CIE A and D65. However, modern lighting technologies such as LEDs can deviate substantially from the Planckian locus, exposing the limitations of relying on conventional one-dimensional CCT for illumination characterization. This paper demonstrates that transitioning from 1D CCT (on the Planckian locus) to a 2D chromaticity space (off the Planckian locus) improves colorimetric accuracy across various mapping approaches. In addition, we replace conventional CCT interpolation with a lightweight multi-layer perceptron (MLP) that leverages 2D chromaticity features for robust colorimetric mapping under non-Planckian illuminants. A lightbox-based calibration procedure incorporating representative LED sources is used to train our MLP. Validated across diverse LED lighting, our method reduces angular reproduction error by 22\% on average in LED-lit scenes, maintains backward compatibility with traditional illuminants, accommodates multi-illuminant scenes, and supports real-time in-camera deployment with negligible additional computational cost. Code and data can be found on the project webpage: \href{https://ccmmlp.github.io}{\texttt{ccmmlp.github.io}}

\end{abstract}

\section{Introduction}
\label{sec:intro}

\begin{figure}[tb]
  \centering
  \includegraphics[width=\textwidth, keepaspectratio]{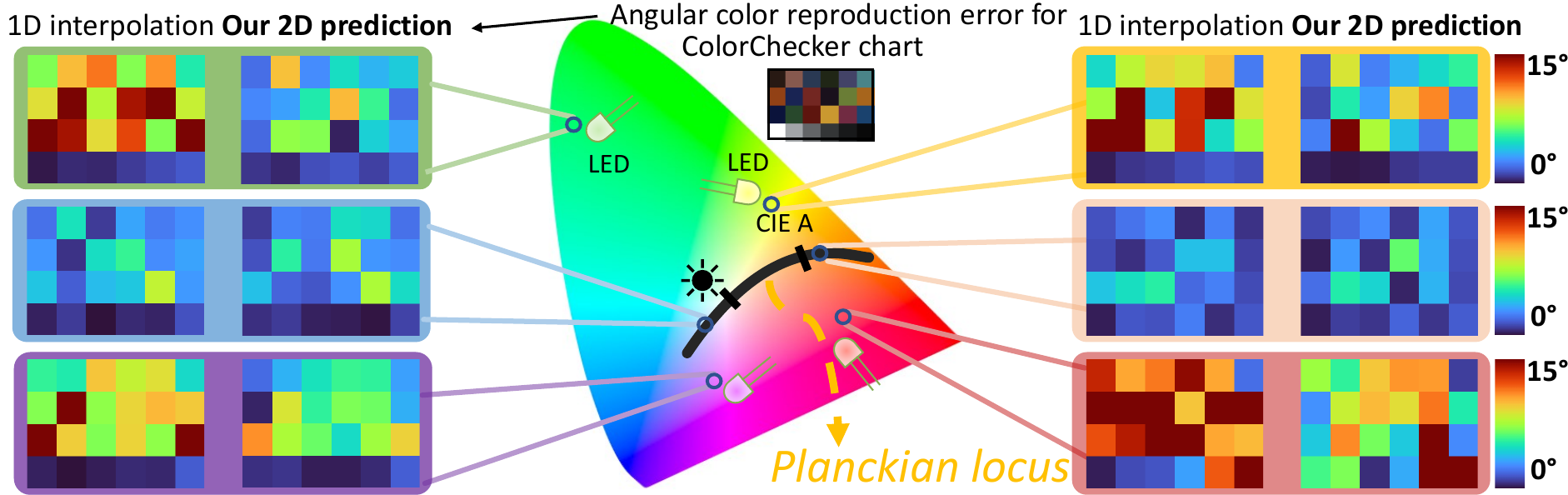}   
  \caption{\textbf{1D Planckian interpolation vs. our 2D chromaticity-based prediction.} Traditional 1D methods perform adequately for light sources whose CCTs lie close to the Planckian locus but are less accurate for light sources (namely, LEDs) that lie off the Planckian locus. Our approach leverages 2D chromaticity coordinates for more robust color reproduction across all illuminant types. Heat maps show angular reproduction error, from $0^\circ$ (dark blue) to $15^\circ$ (dark red), demonstrating comparable performance along the Planckian curve but substantial improvements in LED-dominated regions where existing methods break down.
  }
  
  \label{fig:teaser}
\end{figure}

Digital cameras perform a wide range of in-camera processing routines to convert sensor-specific RGB responses into the final output image~\cite{Delbracio2021MobileCP}. A critical component of the in-camera processing pipeline is the colorimetric mapping stage, which performs illumination correction followed by a color correction stage, also known as color space transformation (CST)\cite{Hakki_CVPR18, Hakki_eccv16}, to map sensor values to a standard color space such as CIE XYZ~\cite{xyz}. The accuracy of this colorimetric mapping impacts color reproduction quality across different lighting conditions.

The current in-camera colorimetric mapping method was developed when early digital cameras primarily captured images illuminated by conventional light sources (e.g., incandescent, fluorescent, and natural/outdoor light). The perceived colors of these early light sources’ spectra are conveniently correlated with blackbody radiators---idealized physical bodies whose spectral distributions depend solely on temperature according to Planck’s law~\cite{wyszecki1982color}. In the CIE xy chromaticity space~\cite{xyz}, these blackbody sources form the Planckian locus, as shown in \cref{fig:teaser}.
Correlated color temperature (CCT) provides a simple 1D representation for characterizing such conventional light sources.
The standard procedure for color correction involves computing the color correction matrix (CCM) using a one-dimensional CCT-based interpolation of pre-calibrated matrices along the Planckian locus.  Typically, two pre-calibrated illuminants are used (1) daylight (6500 K) and (2) incandescent light (CIE A). 

However, the proliferation of LED technology has fundamentally disrupted this framework. Modern artificial illuminants exhibit chromaticity coordinates that deviate significantly from the Planckian locus, resulting in suboptimal color reproduction when using CCT-based interpolation (\cref{fig:teaser}). This mismatch highlights the growing disparity between current camera color-processing assumptions and the spectral characteristics of modern illumination.

This paper proposes shifting from a one-dimensional CCT space to a 2D chromaticity space for CCM estimation. Our approach recognizes that modern illuminants require characterization beyond simple CCT, incorporating camera-specific chromaticity coordinates as input features. Moreover, we employ a low-parameter multi-layer perceptron (MLP) to learn optimal interpolation. By training on an expanded dataset including representative LED illuminants alongside traditional sources, our CCM-MLP achieves superior colorimetric mapping that adapts to contemporary lighting spectral diversity while maintaining backward compatibility.

Our key contributions are as follows: (1) CCM-MLP, a lightweight MLP that predicts CCMs in 2D chromaticity space, replacing traditional CCT-based interpolation with learned transforms capable of handling non-Planckian illuminants; (2) adaptation of traditional and baseline methods to operate in the chromaticity domain; (3) extensive validation demonstrating an average 22\% improvement in color reproduction accuracy across two smartphones and two DSLRs; (4) a spatially varying extension for multi-illuminant scenarios; and (5) a carefully collected laboratory and in-the-wild dataset captured under non-Planckian illuminants for evaluating our method. This work represents a step toward modernizing in-camera color processing in line with the widespread adoption of LED lighting.

\section{Motivation and Related Work}\label{sec:realated_works}

\begin{figure}[t]
  \centering
  \includegraphics[width=1.0\linewidth]{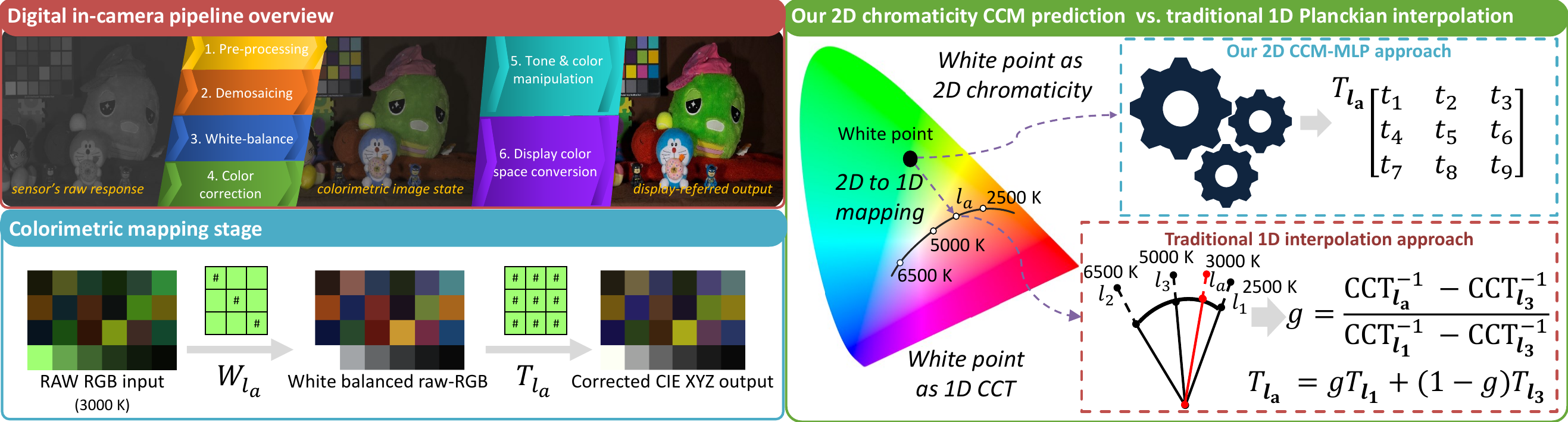}
  \caption{\textbf{Top Left: Camera pipeline overview.} Early stages convert raw sensor responses to colorimetric values, while later stages transform colors to a photo-finished display-referred space~\cite{Hakki_eccv16}. Our work aims to optimize the colorimetric mapping stage.
\textbf{Bottom Left: Colorimetric mapping stage.} When an image is captured under arbitrary illumination, the estimated illumination is used to remove color cast via a diagonal white balance $3\!\times\!3$ matrix ($W_{l_a}$), then mapped to CIE XYZ with an illumination-specific $3\!\times\!3$ CCM ($T_{l_a}$).
\textbf{Right: 2D chromaticity CCM prediction vs. 1D CCT interpolation.} CCM interpolation approaches (e.g., $T_{l_a} = gT_{2500 K} + (1-g)T_{6500K}$) assume CCM parameters can be derived from a CCT-based interpolation, given calibrated illuminants along the Planckian Locus. CCT-based interpolation works poorly for non-Planckian illuminants. We propose a 2D CCM-MLP that takes chromaticity coordinates as input and directly predicts the full $3\!\times\!3$ CCM.}
  \label{fig:cst_interpolation}
\end{figure}

\noindent{\textbf{Motivation.}}
The colorimetric mapping stage of the digital camera pipeline (top/middle of \cref{fig:cst_interpolation} consists of two main steps: white balance (WB) via a diagonal correction and color balance (CB) via a CCM. In the camera pipeline, WB is typically modeled in the 2D raw chromaticity space~\cite{Hakki_eccv16}. Other software, such as Lightroom~\cite{lightroom}, allows 2D WB control via temperature and tint~\cite {Ohno} sliders. While estimated WB values are represented in 2D chromaticity space, the color-balancing stage of the camera pipeline projects 2D chromaticities onto a 1D CCT and utilizes a CCT-based interpolation~\cite{AdobeDNG} to derive the CCM (see Supplementary \cref{sec:cct_interpolation} for details on this procedure).

CCT-based interpolation assumes that conventional light sources lie close to the Planckian locus, however, LEDs and tunable lighting violate the core assumption, as their chromaticity coordinates often deviate significantly from the Planckian locus~\cite{Murdoch,Boyce}. This divergence arises from inherent LED properties, including discontinuous spectra~\cite{Khanna} and narrow-band primaries~\cite{Schubert}. 

When illumination chromaticity coordinates fall far from the Planckian locus, CCT calculations become ill-defined~\cite{Ohno}, and subsequent CCT-based interpolation leads to errors exceeding just-noticeable color differences~\cite{Buzzelli}. Karaimer and Brown~\cite{Hakki_CVPR18} introduced an additional calibration point along the Planckian locus to improve interpolation, an approach later integrated into the Adobe DNG specification~\cite{AdobeDNG}. However, because it still relies on CCT-based interpolation, this method cannot represent light sources that fall off the Planckian locus. Other research has explored improved illumination estimation~\cite{gcc,Zhan,Dongyoung_2024, Juarez} and post-processing strategies~\cite{abc_chiu,deepwb, Finlayson, Hakki_CVPR18}, yet these approaches do not address the fundamental limitation of CCT-based interpolation in the colorimetric mapping stage. To overcome this, we propose predicting the CCM directly in the 2D chromaticity space using a CCM-MLP, as illustrated in \cref{fig:cst_interpolation} (bottom).

In the following, we present a short review of relevant areas for in-camera color reproduction.

\noindent{\textbf{Color constancy.}}~~Early color-correction stages in camera ISP are motivated by the human visual system's color constancy ability to perceive object colors similarly under different illuminations. Computational color constancy generates sensor responses that are invariant to scene illumination in two steps: estimating scene illumination in raw sensor space, then removing the color cast. 

\noindent{\bf{White balance.}}~~Most color constancy works focus on illumination estimation using classical statistics~\cite{BUCHSBAUM, edge_cc, cbcorr, gijsenij2012},
gamut-based methods~\cite{Forsyth1990, finlayson_2013}, and deep-learning approaches~\cite{Lou2015,barron2015convolutional,Lo2021,deepwb}. After illuminant estimation, correction typically applies a diagonal $3\times3$ matrix to the raw-RGB image. Simple WB correction only ensures proper correction of neutral colors (\(R=G=B\) after correction).

\noindent{\textbf{Color balance.}}~~Additional correction is required for non-neutral colors. Thus, diagonal correction provides white balance (WB), while full correction is color balance (CB). Diagonal correction suffices for particular illuminations (e.g., broadband) to achieve full CB, but many illuminations require full color balancing~\cite{cheng}. Camera ISPs apply an illumination-specific CCM~\cite{Hakki_CVPR18} after WB to correct non-neutral colors. Adobe DNG implements this by mapping illuminants to CCT, then interpolating between calibrated $3\times 3$ CCM's that map to CIE XYZ. Initially, DNG supported two CCM's at high ($\sim$6500 K) and low ($\sim$2500 K) CCT. Karaimer and Brown~\cite{Hakki_CVPR18} showed that an additional calibration point ($\sim$4300 K) improves reproduction, resulting in DNG now supporting three CCMs. Other higher-dimensional mappings, such as polynomial~\cite{cheung} and root-polynomial~\cite{Finlayson}, could be used in place of the $3\times 3$ transformation when invertibility is not required. 

\noindent{\bf{Direct mapping.}} Recent methods such as CCCNN~\cite{cccnn} and EXPINV~\cite{expinv} learn an illuminant-dependent colorimetric mapping by training MLPs that take raw pixel values and illuminant CCT as input to predict CIE XYZ values directly. However, these networks require expensive per-pixel evaluation at test time and rely on access to camera spectral sensitivities during training. Their pixel-wise formulation introduces spatial inconsistencies and color casts, as we demonstrate in \cref{sec:lightbox_results}. Finally, their dependence on 1D CCT illumination coordinates limits these networks' ability to model non-Planckian illuminants. In contrast, our proposed CCM-MLP predicts a global CCM in a single forward pass, enforces spatial consistency, requires no spectral sensitivity data, and generalizes better across diverse light sources by operating in 2D chromaticity space.

\section{Method}
\begin{figure*}[t]  
    \centering
    \includegraphics[width=\textwidth]{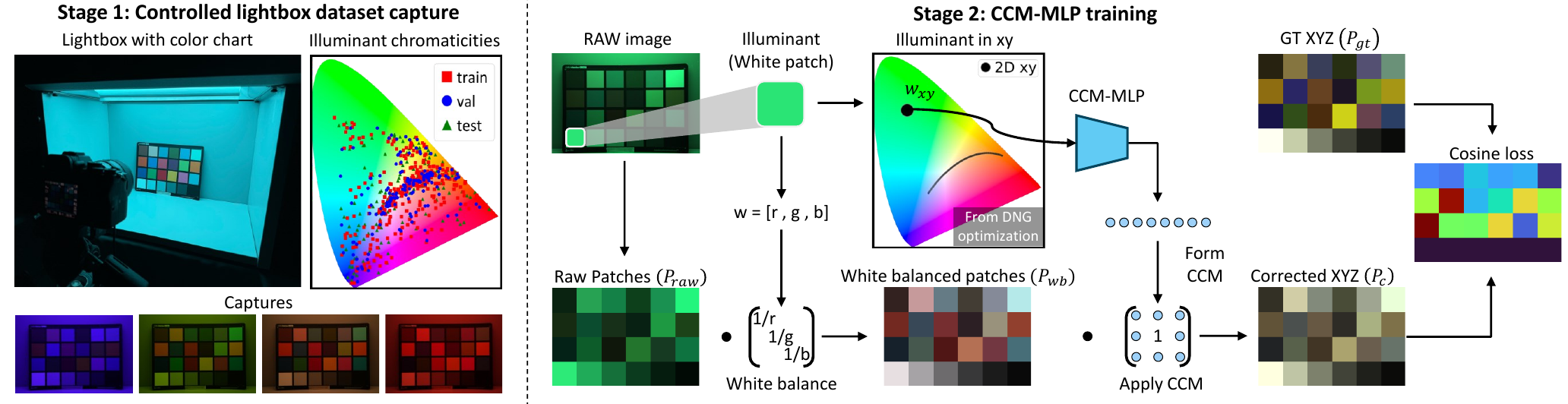}
    \caption{\textbf{Our 2D chromaticity CCM prediction framework.} \textit{(Stage 1) Controlled lightbox dataset capture:} A ColorChecker~\cite{xrite} target is captured under diverse illumination, including traditional both Planckian an LED illuminants covering  on-locus and off-locus illuminants. (Stage 2) CCM-MLP training: Our MLP takes 2D chromaticity coordinates as input and predicts the color correction matrix. To train CCM-MLP, we begin with raw color patches $P_{\text{raw}}$ and apply white balancing using the estimated illuminant $w$, resulting in white balanced patches $P_{\text{wb}}$. We then convert the illuminant $w$ into chromaticity coordinates $w_{xy}$ via the DNG optimization procedure (see Supplementary \cref{sec:cct_interpolation} ). The chromaticity $w_{xy}$ serves as input to our CCM-MLP, which predicts the CCM applied to $P_{\text{wb}}$. Finally, we train with a cosine loss between the corrected patches $P_c$ and the ground-truth patches $P_{\text{gt}}$ in XYZ space. 
}    
    \label{fig:setup}
    
\end{figure*}

We aim to learn a mapping from illuminant chromaticity to a CCM using a two-stage method, visualized in \cref{fig:setup}, involving calibration data capture within a lightbox (\cref{sec:data_capture}) and training a lightweight CCM-MLP (\cref{sec:model}). Building a drop-in replacement for the color correction stage allows us to maintain the two-stage approach~\cite{AdobeDNG} that currently dominates in-camera pipelines while keeping training requirements down.

\subsection{Data Capture} 
\label{sec:data_capture}
Traditional color correction calibration methods calibrate CCMs by capturing color charts under standard illuminants such as CIE A or D65. However, we require our method to handle non-Planckian illuminants that a CCT cannot adequately describe. The first step in our process is capturing calibration data across diverse illuminants,  providing broad coverage of the xy chromaticity space.

We use a modified GTI lightbox~\cite{gti} to capture a ColorChecker~\cite{xrite} target under a wide range of lighting conditions using two smartphones and two DSLRs (Samsung S22, Google Pixel 9 Pro, Sony $\alpha$57, and Canon EOS 5D Mark III). This modified lightbox replaces the original light sources with Telelumen multispectral LED lighting~\cite{lightbox}, which consists of seven narrow-band LEDs that can be adjusted to match a target spectral power distribution (SPD). Using the Telelumen lightbox, we capture the color chart under many SPDs, resulting in a large calibration dataset. 

We capture each color chart under 790 different illuminant SPDs sourced from two methods. First, we utilize 390 illuminants from Punnappurath et al.~\cite{Hoang} that estimate various light sources, including sunlight, fluorescent, incandescent, and LED. We use the original splits of 195 training, 78 validation, and 117 test illuminants. Then, we introduce a set of 400 illuminants generated by linearly combining the Telelumen LED SPDs with weights randomly sampled from a Dirichlet distribution (see Supplementary \cref{sec:lightbox_supp}  for details). Using this distribution results in an even spread of illuminants across the chromaticity space and allows us to sample more illuminants farther from the Planckian Locus. We randomly split the dataset into 200 train, 80 validation, and 120 test illuminants. We combine both datasets for our calibration, resulting in 395 train, 158 validation, and 237 test illuminants.

\subsection{Model}
\label{sec:model}
We propose a model, CCM-MLP, that maps a white point chromaticity $w_{xy}$ to a CCM $T$. Following Karaimer and Brown~\cite{Hakki_CVPR18}, we use the setting of mapping white-balanced raw images into the XYZ canonical space, and assume the raw image $I_{raw}$ and ground-truth white point chromaticity $w = [r/g, b/g]$ are known.  Following previous work~\cite{Hakki_CVPR18}, we use the ground-truth white point from the color chart white patch and compute the white-balanced raw image $I_{wb}$ by applying a diagonal correction~\cite{barron2015convolutional}. For our model, we convert the raw-space chromaticity $w$ to its xy-space counterpart $w_{xy}$ using the DNG~\cite{AdobeDNG} optimization procedure described in Supplementary \cref{sec:cct_interpolation}. This algorithm is fast, commonly implemented on modern phones~\cite{Hakki_CVPR18}, and executes in an average of $250\ \mu s$ on an iPhone 16.

CCM-MLP is a ReLU~\cite{relu} MLP with one hidden layer of 32 neurons and a linear output layer of eight neurons, trained with the Adam optimizer~\cite{adam}. Since the angular-error loss defines the CCM only up to scale, we fix its central element to 1 to remove this ambiguity, leaving eight free entries for the network to predict. These are reshaped into the $3\times3$ CCM. If invertibility is not a requirement, we also discuss predicting polynomial and root-polynomial mappings in~\cref{sec:results}.

To train our method, we extract the mean color of the 24 patches in the raw image $P_{raw} \in \mathbb{R}^{24 \times 3}$ by averaging across an $11 \times 11$ pixel region per patch. We then apply a diagonal white balance to each patch using the white patch \( w \), resulting in white-balanced patches \( P_{\mathrm{wb}} \).
 For a given illumination, the model aims to predict a $3 \times 3$ CCM that, when applied to $P_{wb}$ minimizes the angular error between the corrected XYZ patches $P_c\in \mathbb{R}^{24\times 3}$ and the ground-truth XYZ $P_{gt}\in \mathbb{R}^{24\times 3}$.

We utilize our calibration dataset to train the model and use the ground-truth XYZ values provided by X-Rite~\cite{xrite} for the color chart. Following Karaimer and Brown~\cite{Hakki_CVPR18}, we aim to minimize the angular error of the corrected XYZ colors and the ground-truth XYZ colors. We found direct angular error loss was unstable during optimization; thus, we follow other works~\cite{clip, wang2018cosface} and use a cosine loss, which still minimizes angular errors~\cite{wang2018cosface}. We define the loss function as the average cosine loss over all color patches:

{\setlength{\abovedisplayskip}{8pt}%
 \setlength{\abovedisplayshortskip}{0pt}%
 \begin{equation}
    L(P_c, P_{gt}) = \frac{1}{N} \sum_{i=1}^{N} \left(1 - \cos\left(\hat{P_c^i}, \hat{P_{gt}^i}\right)\right),
 \end{equation}
}
where \( \hat{P_c^i} \) and \( \hat{P_{gt}^i} \) are the normalized colors at index \( i \).

\section{Evaluation}
\label{sec:results}
 First, we evaluate 1D CCT vs.~2D chromaticity-based approaches using the lightbox dataset described in~\cref{sec:data_capture}. Next, we test our lightbox calibration across a set of in-the-wild scenes. Finally, we evaluate on LSMI~\cite{lsmi} under both single- and multi-illuminant settings, and the NUS~\cite{nus} dataset.

\subsection{Evaluation Details}

\noindent{\bf{Baselines.}}~For comparison, we implement a variety of state-of-the-art baselines. First, to contextualize the upper bound of CCM prediction, we compute the Oracle, the best-fit \(3 \times 3\) matrix that minimizes the angular error given the full color chart. Then, we compare with the standard 2-CCM interpolation described by the DNG~\cite{AdobeDNG}, which uses matrices recalibrated by capturing the color chart under Standard Illuminant A (2850 K) and D65 (6500 K) lighting within the lightbox. We also compare to the 3-CCM~\cite{Hakki_CVPR18} interpolation with an additional calibrated CCM at 4250 K. Next, we implement a nearest neighbors (NN)~\cite{nearestneighbors} baseline that maps each training white point with its best-fit \( 3 \times 3 \) CCM (minimizing angular error on the color chart). At test time, given a white point coordinate, we retrieve the nearest illuminant by L2 distance and apply its corresponding matrix. Finally, we compare direct MLP methods, CCCNN~\cite{cccnn} and EXPINV~\cite{expinv}, that perform per-pixel correction given an input color and white point.

\noindent{\bf{1D CCT vs 2D Chromaticity.}}~To demonstrate the importance of using 2D chromaticity coordinates, we also compare our CCM-MLP with a 1D CCT-based variant. Additionally, we adapt the 1D CCT-based baseline methods (NN, EXPINV, and CCCNN) into 2D chromaticity-based versions to show the benefits of our approach.

\noindent{\bf{Polynomial mappings.}}~To evaluate the effect of higher-dimensional transforms, we extend our CCM-MLP to output polynomial~\cite{cheung} and root-polynomial \cite{Finlayson} mappings. We compare standard linear \( 3 \times 3 \) transforms against polynomial and root-polynomial mappings of varying sizes and find that root-polynomial \( 3 \times 6 \) mappings yield the lowest angular error (see \cref{tab:ablation}). We denote the $3 \times 6$ root-polynomial version as CCM-MLP (RP) in all subsequent tables.

\noindent{\bf{Training details.}}~We use the dataset splits in~\cref{sec:data_capture} to train and evaluate all methods except CCCNN~\cite{cccnn} and EXPINV~\cite{expinv}, which learn a per-pixel mapping; following their original papers, we train them on a larger simulated dataset. Additionally, we adapt EXPINV with the CCT conditioning utilized in CCCNN~\cite{cccnn}. For simulation, we measure each camera’s spectral sensitivities using CamspecV2~\cite{camspec}, collect illumination SPDs from our lightbox capture setup (see ~\cref{sec:data_capture}), and utilize 1269 reflectance spectra from the Munsell Book of Color~\cite{munsell} measured by Chromaxion~\cite{chromaxion}. Finally, we synthesize data using the image formation model described in CCCNN~\cite{cccnn} and EXPINV~\cite{expinv}.

All MLP methods are trained for 100,000 iterations using the Adam~\cite{adam} optimizer with a learning rate of 0.001. Our CCM-MLP is trained using the cosine loss described in \cref{sec:model}, while CCCNN~\cite{cccnn} and EXPINV~\cite{expinv} are trained with the recommended \(\Delta E\) loss. Spectral sensitivities are unavailable for the LSMI~\cite{lsmi} and NUS~\cite{nus} datasets, so CCCNN and EXPINV are excluded from those evaluations.

\begin{table*}[tb]
\caption{\textbf{Quantitative evaluation on lightbox dataset.} We compare angular error and $\Delta E_{2000}$ of traditional 2-CCM/3-CCM interpolation~\cite{AdobeDNG, Hakki_CVPR18}, NN~\cite{nearestneighbors}, CCCNN~\cite{cccnn}, EXPINV~\cite{expinv}, and our CCM-MLP. For NN, CCCNN, EXPINV, and CCM-MLP,  we evaluate both 1D CCT implementations and our proposed 2D chromaticity adaptations (highlighted in blue). Methods Marked (RP) predict a root-polynomial~\cite{Finlayson} mapping.}

\centering

\small
\setlength{\tabcolsep}{1pt} 

\resizebox{\textwidth}{!}{%
\begin{tabular}{@{}llcc *{4}{cccc} @{}}
\toprule
& & & & \multicolumn{16}{c}{\textbf{Angular Error ($\boldsymbol{\degree}$)}} \\
\cmidrule(lr){5-20}
\multirow[b]{2}{*}[-6pt]{\textbf{Method}} &
\multirow[b]{2}{*}[-6pt]{\textbf{Dim}} & 
\multirow[b]{2}{*}[-6pt]{\shortstack{\textbf{Size}\\\textbf{(KB)}}} & 
\multirow[b]{2}{*}[-6pt]{\shortstack{\textbf{MACs}\\\textbf{(M)}}} & 
\multicolumn{4}{c}{\textbf{Sony}} & 
\multicolumn{4}{c}{\textbf{Canon}} & 
\multicolumn{4}{c}{\textbf{Pixel}} & 
\multicolumn{4}{c}{\textbf{Samsung}} \\
\cmidrule(lr){5-8} \cmidrule(lr){9-12} \cmidrule(lr){13-16} \cmidrule(lr){17-20}
& & & & Mean & 25\% & 50\% & 90\% & Mean & 25\% & 50\% & 90\% & Mean & 25\% & 50\% & 90\% & Mean & 25\% & 50\% & 90\% \\
\midrule
Oracle                                        & 1D &                      0.00 &                      0.01 &                      1.71 &                      0.54 &                      1.15 &                      3.68 &                      2.06 &                      0.64 &                      1.31 &                      4.65 &                      2.79 &                      0.95 &                      2.00 &                      5.94 &                      2.58 &                      0.84 &                      1.83 &                      5.47 \\
2-CCM~\cite{AdobeDNG}                         & 1D &  \cellcolor{tabfirst}0.07 &  \cellcolor{tabfirst}0.01 &                      3.31 & \cellcolor{tabsecond}0.80 &                      1.98 &                      7.91 &                      3.86 &                      0.96 &                      2.31 &                      9.59 &                      4.56 &                      1.48 &                      3.25 &                      9.94 &                      4.51 &                      1.53 &                      3.19 &                      9.95 \\
3-CCM~\cite{Hakki_CVPR18}                     & 1D & \cellcolor{tabsecond}0.11 &  \cellcolor{tabfirst}0.01 &                      3.31 & \cellcolor{tabsecond}0.80 &                      2.01 &                      7.91 &                      3.86 &                      0.94 &                      2.32 &                      9.68 &                      4.53 &                      1.48 &                      3.22 &                      9.95 &                      4.51 &                      1.51 &                      3.18 &                      9.94 \\
NN~\cite{nearestneighbors}                    & 1D &                      15.43 &  \cellcolor{tabfirst}0.01 &                      3.93 &                      0.94 &                      2.33 &                      9.59 &                      4.68 &                      1.05 &                      2.74 &                      11.62 &                      5.24 &                      1.54 &                      3.56 &                      11.73 &                      5.48 &                      1.82 &                      3.59 &                      12.27 \\
CCCNN~\cite{cccnn}                            & 1D &                      68.51 & \cellcolor{tabsecond}13.39 &                      3.28 &                      1.09 &                      2.13 &                      7.38 &                      3.58 &                      1.19 &                      2.42 &                      8.02 &                      4.39 &                      1.47 &                      2.96 &                      9.67 &                      4.71 &                      1.76 &                      3.35 &                      10.00 \\
EXPINV~\cite{expinv}                          & 1D &                      68.53 & \cellcolor{tabsecond}13.39 &                      3.13 &                      0.91 &                      1.86 &                      7.42 &                      3.61 &                      1.08 &                      2.33 &                      8.46 &                      4.31 &  \cellcolor{tabthird}1.26 &                      2.85 &                      9.88 &                      4.68 &                      1.43 &                      3.31 &                      10.40 \\
CCM-MLP                                       & 1D &  \cellcolor{tabthird}1.41 &  \cellcolor{tabfirst}0.01 &                      3.09 &                      0.91 &                      1.85 &                      7.35 &                      3.52 &                      0.98 &                      2.06 &                      8.60 &                      4.11 &                      1.37 &                      2.62 &                      9.55 &                      4.05 &                      1.29 &  \cellcolor{tabthird}2.60 &                      9.42 \\
CCM-MLP(RP)                                   & 1D &                      2.57 &  \cellcolor{tabfirst}0.01 &  \cellcolor{tabthird}2.82 &                      0.85 &  \cellcolor{tabthird}1.73 &                      6.65 &                      3.26 &                      0.96 &                      2.05 &                      7.68 &  \cellcolor{tabthird}3.77 &                      1.34 &  \cellcolor{tabthird}2.45 &                      8.57 &  \cellcolor{tabthird}3.70 &  \cellcolor{tabthird}1.21 & \cellcolor{tabsecond}2.43 &  \cellcolor{tabthird}8.52 \\
\cellcolor{blue!30}NN~\cite{nearestneighbors} & \cellcolor{blue!30}2D &                      16.97 &  \cellcolor{tabfirst}0.01 &                      3.24 &  \cellcolor{tabthird}0.82 &                      1.79 &                      7.73 &                      3.56 &  \cellcolor{tabthird}0.90 &  \cellcolor{tabthird}2.03 &                      8.51 &                      4.30 &                      1.37 &                      2.84 &                      9.81 &                      4.15 &                      1.33 &                      2.72 &                      9.48 \\
\cellcolor{blue!30}CCCNN~\cite{cccnn}         & \cellcolor{blue!30}2D &                      69.01 &  \cellcolor{tabthird}13.48 &                      2.89 &                      0.97 &                      1.99 &                      6.43 &  \cellcolor{tabthird}3.00 &                      1.13 &                      2.20 & \cellcolor{tabsecond}6.19 &                      4.00 &                      1.50 &                      2.93 &                      8.76 &                      4.17 &                      1.78 &                      3.32 &  \cellcolor{tabthird}8.52 \\
\cellcolor{blue!30}EXPINV~\cite{expinv}       & \cellcolor{blue!30}2D &                      69.03 &  \cellcolor{tabthird}13.48 & \cellcolor{tabsecond}2.66 &                      0.87 &                      1.78 & \cellcolor{tabsecond}5.97 &                      3.11 &                      0.94 &                      2.26 &                      6.89 &                      3.84 &                      1.32 &                      2.71 &  \cellcolor{tabthird}8.41 &                      4.14 &                      1.49 &                      3.05 &                      8.86 \\
\cellcolor{blue!30}CCM-MLP                    & \cellcolor{blue!30}2D &                      1.54 &  \cellcolor{tabfirst}0.01 & \cellcolor{tabsecond}2.66 &  \cellcolor{tabthird}0.82 & \cellcolor{tabsecond}1.70 &  \cellcolor{tabthird}6.14 & \cellcolor{tabsecond}2.89 & \cellcolor{tabsecond}0.87 & \cellcolor{tabsecond}1.85 &  \cellcolor{tabthird}6.67 & \cellcolor{tabsecond}3.60 & \cellcolor{tabsecond}1.22 & \cellcolor{tabsecond}2.38 & \cellcolor{tabsecond}8.18 & \cellcolor{tabsecond}3.60 & \cellcolor{tabsecond}1.20 & \cellcolor{tabsecond}2.43 & \cellcolor{tabsecond}8.23 \\
\cellcolor{blue!30}CCM-MLP(RP)                & \cellcolor{blue!30}2D &                      2.70 &  \cellcolor{tabfirst}0.01 &  \cellcolor{tabfirst}2.36 &  \cellcolor{tabfirst}0.71 &  \cellcolor{tabfirst}1.44 &  \cellcolor{tabfirst}5.53 &  \cellcolor{tabfirst}2.60 &  \cellcolor{tabfirst}0.71 &  \cellcolor{tabfirst}1.58 &  \cellcolor{tabfirst}6.17 &  \cellcolor{tabfirst}3.27 &  \cellcolor{tabfirst}1.13 &  \cellcolor{tabfirst}2.14 &  \cellcolor{tabfirst}7.47 &  \cellcolor{tabfirst}3.28 &  \cellcolor{tabfirst}1.17 &  \cellcolor{tabfirst}2.20 &  \cellcolor{tabfirst}7.25 \\
\cmidrule{5-20}
\multicolumn{4}{c}{} & \multicolumn{16}{c}
{\textbf{$\boldsymbol{\Delta E_{2000}}$ }} \\

\cmidrule{5-20}
Oracle                                        & 1D &                      0.00 &                      0.01 &                      5.37 &                      2.91 &                      4.70 &                      9.49 &                      5.83 &                      3.22 &                      5.20 &                      10.17 &                      7.07 &                      4.55 &                      6.41 &                      12.14 &                      6.51 &                      4.09 &                      5.94 &                      11.06 \\
2-CCM~\cite{AdobeDNG}                         & 1D &  \cellcolor{tabfirst}0.07 &  \cellcolor{tabfirst}0.01 &                      7.21 &                      3.78 &                      6.06 &                      13.18 &                      8.15 &                      4.41 &                      6.63 &                      15.08 &                      9.18 &                      5.68 &                      7.91 &                      15.74 &                      8.61 &                      5.28 &                      7.64 &                      14.76 \\
3-CCM~\cite{Hakki_CVPR18}                     & 1D & \cellcolor{tabsecond}0.11 &  \cellcolor{tabfirst}0.01 &                      7.20 &                      3.77 &                      6.02 &                      13.19 &                      8.13 &                      4.40 &                      6.65 &                      15.10 &                      9.16 &                      5.68 &                      7.93 &                      15.65 &                      8.62 &                      5.28 &                      7.65 &                      14.76 \\
NN~\cite{nearestneighbors}                    & 1D &                      15.43 &  \cellcolor{tabfirst}0.01 &                      8.02 &                      4.00 &                      6.63 &                      15.03 &                      9.21 &                      4.57 &                      7.20 &                      17.91 &                      9.53 &                      5.57 &                      8.07 &                      17.03 &                      9.49 &                      5.36 &                      8.18 &                      16.81 \\
CCCNN~\cite{cccnn}                            & 1D &                      68.51 & \cellcolor{tabsecond}13.39 &                      7.00 & \cellcolor{tabsecond}3.53 &  \cellcolor{tabthird}5.90 &                      12.55 &                      7.93 &                      4.16 &                      6.53 &                      14.42 &                      8.36 &                      4.91 &                      7.40 &                      14.39 &                      8.45 &                      5.05 &                      7.96 &                      13.93 \\
EXPINV~\cite{expinv}                          & 1D &                      68.53 & \cellcolor{tabsecond}13.39 &                      7.03 &  \cellcolor{tabthird}3.55 &                      5.98 &                      12.42 &                      7.76 &                      3.99 &                      6.75 &                      14.25 &                      8.09 &                      4.83 &                      7.14 &                      13.97 &                      8.17 &                      5.07 &                      7.29 &                      13.98 \\
CCM-MLP                                       & 1D &  \cellcolor{tabthird}1.41 &  \cellcolor{tabfirst}0.01 &                      7.09 &                      3.86 &                      6.20 &                      12.56 &                      7.81 &                      4.25 &                      6.87 &                      14.35 &                      8.48 &                      4.99 &                      7.45 &                      14.58 &                      8.14 &                      4.65 &                      7.51 &                      14.10 \\
CCM-MLP(RP)                                   & 1D &                      2.57 &  \cellcolor{tabfirst}0.01 &                      7.60 &                      4.97 &                      6.83 &                      12.60 &                      8.28 &                      5.36 &                      7.25 &                      14.01 &                      8.19 & \cellcolor{tabsecond}4.60 & \cellcolor{tabsecond}6.89 &                      14.82 &                      8.39 &                      4.84 &  \cellcolor{tabfirst}7.09 &                      14.65 \\
\cellcolor{blue!30}NN~\cite{nearestneighbors} & \cellcolor{blue!30}2D &                      16.97 &  \cellcolor{tabfirst}0.01 &                      7.29 &                      3.60 &                      5.99 &                      13.17 &                      7.79 & \cellcolor{tabsecond}3.92 & \cellcolor{tabsecond}6.37 &                      14.21 &                      8.74 &                      5.26 &                      7.35 &                      15.30 &                      8.37 &                      4.80 &  \cellcolor{tabthird}7.16 &                      14.71 \\
\cellcolor{blue!30}CCCNN~\cite{cccnn}         & \cellcolor{blue!30}2D &                      69.01 &  \cellcolor{tabthird}13.48 & \cellcolor{tabsecond}6.63 &                      3.69 &  \cellcolor{tabfirst}5.62 & \cellcolor{tabsecond}11.40 &  \cellcolor{tabfirst}6.78 &  \cellcolor{tabfirst}3.58 &  \cellcolor{tabfirst}6.07 &  \cellcolor{tabfirst}11.61 &                      8.07 &                      4.88 &                      7.42 & \cellcolor{tabsecond}12.91 &  \cellcolor{tabthird}7.76 &                      4.86 &                      7.66 &  \cellcolor{tabfirst}11.96 \\
\cellcolor{blue!30}EXPINV~\cite{expinv}       & \cellcolor{blue!30}2D &                      69.03 &  \cellcolor{tabthird}13.48 &  \cellcolor{tabfirst}6.48 &  \cellcolor{tabfirst}3.45 & \cellcolor{tabsecond}5.84 &  \cellcolor{tabfirst}11.03 & \cellcolor{tabsecond}7.07 &                      3.98 &                      6.52 & \cellcolor{tabsecond}11.93 & \cellcolor{tabsecond}7.77 &  \cellcolor{tabthird}4.74 &  \cellcolor{tabthird}7.10 &  \cellcolor{tabfirst}12.63 &  \cellcolor{tabfirst}7.68 &  \cellcolor{tabfirst}4.37 &                      7.51 & \cellcolor{tabsecond}12.20 \\
\cellcolor{blue!30}CCM-MLP                    & \cellcolor{blue!30}2D &                      1.54 &  \cellcolor{tabfirst}0.01 &  \cellcolor{tabthird}6.73 &                      3.66 &                      6.07 &                      11.58 &  \cellcolor{tabthird}7.17 &  \cellcolor{tabthird}3.96 &  \cellcolor{tabthird}6.41 &  \cellcolor{tabthird}12.55 &  \cellcolor{tabthird}7.95 &                      4.78 &                      7.21 &  \cellcolor{tabthird}13.26 & \cellcolor{tabsecond}7.74 & \cellcolor{tabsecond}4.38 &                      7.29 &  \cellcolor{tabthird}13.11 \\
\cellcolor{blue!30}CCM-MLP(RP)                & \cellcolor{blue!30}2D &                      2.70 &  \cellcolor{tabfirst}0.01 &                      7.16 &                      4.57 &                      6.48 &  \cellcolor{tabthird}11.56 &                      7.47 &                      4.28 &                      6.83 &                      12.68 &  \cellcolor{tabfirst}7.74 &  \cellcolor{tabfirst}4.50 &  \cellcolor{tabfirst}6.64 &                      13.84 &                      8.08 &  \cellcolor{tabthird}4.56 & \cellcolor{tabsecond}7.10 &                      13.97 \\
\bottomrule

\end{tabular}
}

\vspace{-10pt}
\label{tab:main_results}
\end{table*}

\begin{wrapfigure}{r}{0.32\textwidth}
    \centering
    \vspace{0pt}
        \caption{We visualize the chromaticities of our train (lightbox) and in-the-wild illuminants, which include a few out-of-distribution illuminants (circled purple).}
        \vspace{-4pt}
    \includegraphics[width=0.7\linewidth]{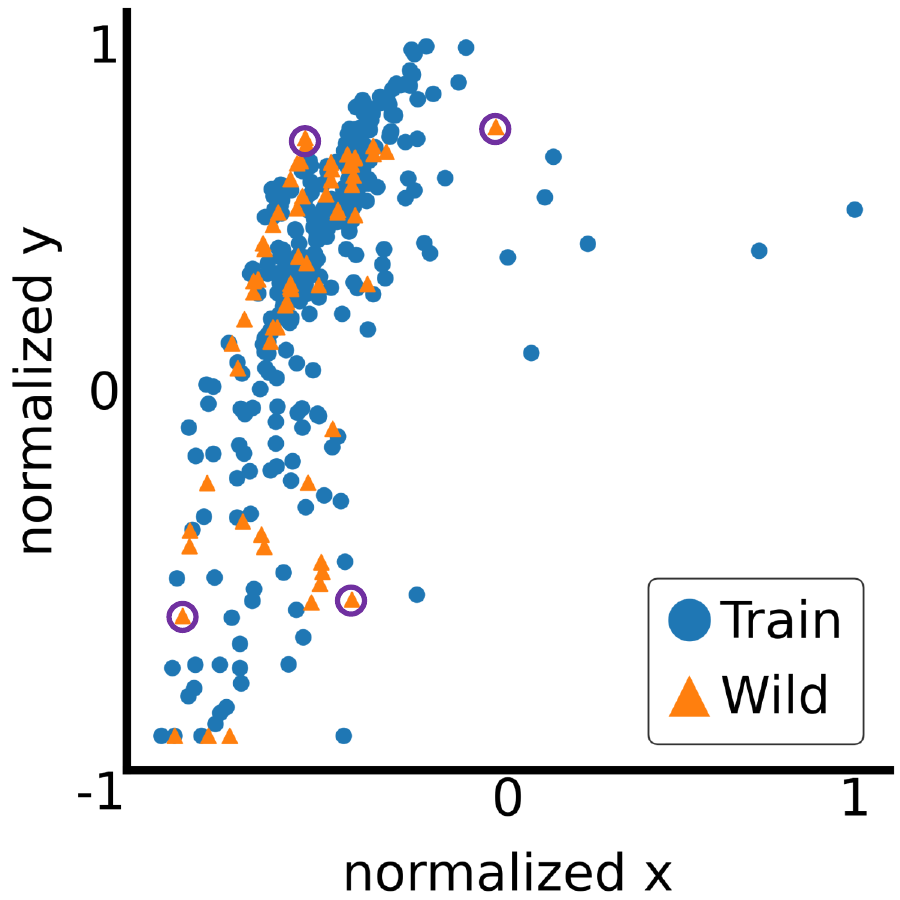}
    \vspace{-25pt}
    \makeatletter\cref@label{fig:precon_plot}\makeatother
\end{wrapfigure}

\noindent{\bf{Noise augmentation.}} When queried with out-of-domain chromaticities not seen in our training set, our 2D CCM-MLP can sometimes fail to generalize. As illustrated in ~\cref{fig:precon_plot}, the training distribution does not contain all in-the-wild illuminants; these chromaticities are outside of the range of our lightbox LEDs. To let our models generalize beyond this range, we add Gaussian noise to the input chromaticity coordinates during training~\cite{ling2025stochastic}. However, a lightbox with a broader range of narrow-band LEDs, or synthetic data augmentation discussed in Supplementary \cref{sec:synthesize}, could provide a similar coverage. For CCM-MLP, we add noise with standard deviation $\sigma_0 = 0.05$ to the normalized input coordinates for all training iterations. We show the positive effects of this augmentation in our ablation study (\cref{tab:ablation}). Finally, we exclude this augmentation for direct-prediction MLP baselines (EXPINV~\cite{expinv}, CCCNN~\cite{cccnn}) as it can degrade their performance (see Supplementary \cref{sec:supp_model_abation}).

\noindent{\bf{Metrics.}}~For evaluation, we utilize angular error following previous work~\cite{Hakki_CVPR18}. We also compare \(\Delta E_{2000}\)  after applying a uniform scaling to all colors to match the Y channel of the third gray patch (bottom row of color chart). Finally, we report model parameter sizes (in KB) and inference costs (in MACs) when applied to a $1080 \times 720$ image.

\subsection{Lightbox Dataset Results}
\label{sec:lightbox_results}
We compare all methods quantitatively on our dataset in \cref{tab:main_results} and qualitatively in \cref{fig:lightbox_results} (additional visualizations in Supplementary \cref{sec:lightbox_supp}). We find that \textit{all methods benefit from using a 2D chromaticity coordinate} instead of a 1D CCT, which confirms our motivation for this work. Additionally, our 2D CCM-MLP surpasses traditional interpolation and nearest-neighbor methods, providing a strong solution that integrates seamlessly into the existing DNG~\cite{AdobeDNG} pipeline, which relies on linear transformations.

\begin{figure}[!t]
  \centering
  \includegraphics[width=1.0\linewidth]{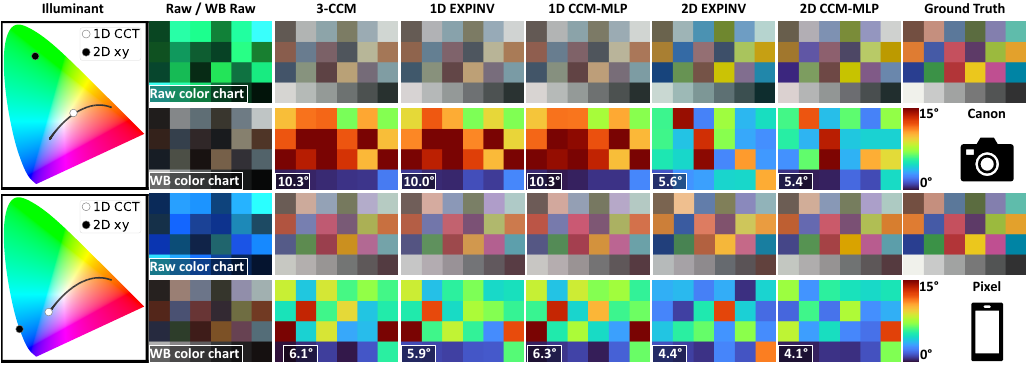}
\caption{\textbf{Colorimetric mapping results on laboratory dataset.} Results for two non-Planckian LED sources: green LED (top half) and blue LED (bottom half) illuminants, captured using Canon and Pixel cameras. Each half starts with the illuminant's CCT and xy chromaticity and contains two rows: the top row shows the raw-RGB color chart and display-referred renderings; the bottom row shows the white-balanced color chart and error maps. Methods: 3-CCM~\cite{Hakki_CVPR18}, 1D EXPINV~\cite{expinv}, 1D CCM-MLP, 2D EXPINV~\cite{expinv}, 2D CCM-MLP, and ground truth.} 
  \label{fig:lightbox_results}
\end{figure}

\begin{figure}[!t]
  \centering
  \includegraphics[width=1.0\linewidth]{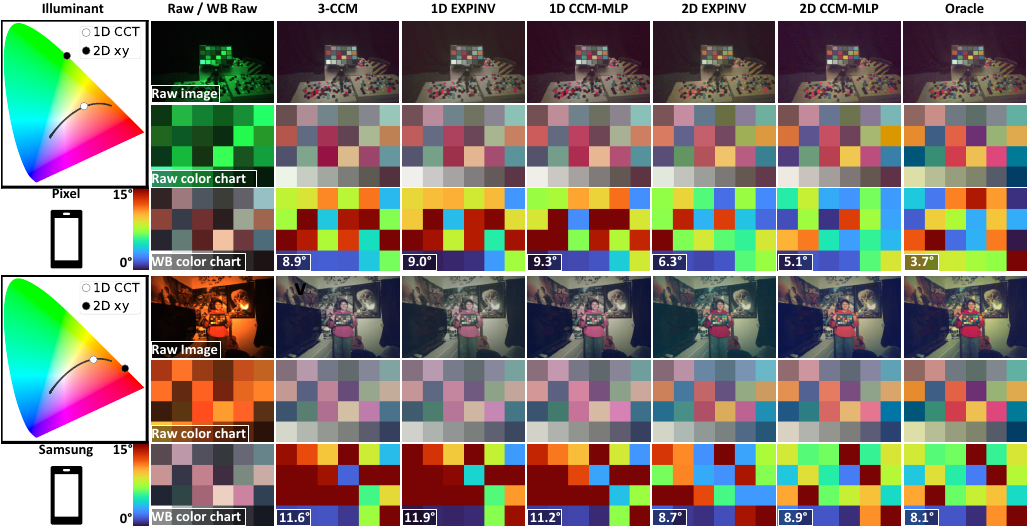}

\caption{\textbf{Colorimetric mapping results on in-the-wild captures.} Results for two non-Planckian LED sources: green LED (top half) and orange LED (bottom half) illuminants, captured using Pixel and Samsung cameras. Each half starts with the illuminant's CCT and xy chromaticity and contains three rows: the top row shows raw-RGB image, and display-referred renderings; the middle row shows raw-RGB color chart and display-referred renderings; the bottom row shows white-balanced color chart and angular error maps. Methods: 3-CCM~\cite{Hakki_CVPR18}, 1D EXPINV~\cite{expinv}, 1D CCM-MLP, 2D EXPINV~\cite{expinv}, 2D CCM-MLP, and Oracle.}
  \label{fig:in-the-wild}
\end{figure}

\noindent{\bf{In-the-wild results.}}
\label{sec:in-the-wild}
After calibrating the CCM prediction methods with lightbox images, we evaluate on a smaller set of 150 in-the-wild photos captured with both smartphones (Pixel and Samsung). As shown in \cref{fig:in-the-wild}, methods that rely on 2D chromaticity deliver qualitatively superior results, enabling more accurate color correction under challenging LED illumination across diverse regions containing skin tones, posters, and craft materials. We also observe that EXPINV~\cite{expinv}, a per-pixel prediction method, exhibits color casts when applied to full-size images. Additional comparisons are provided in Supplementary \cref{sec:lightbox_supp}.

\begin{figure}[t]
\centering
\includegraphics[width=0.8\linewidth]{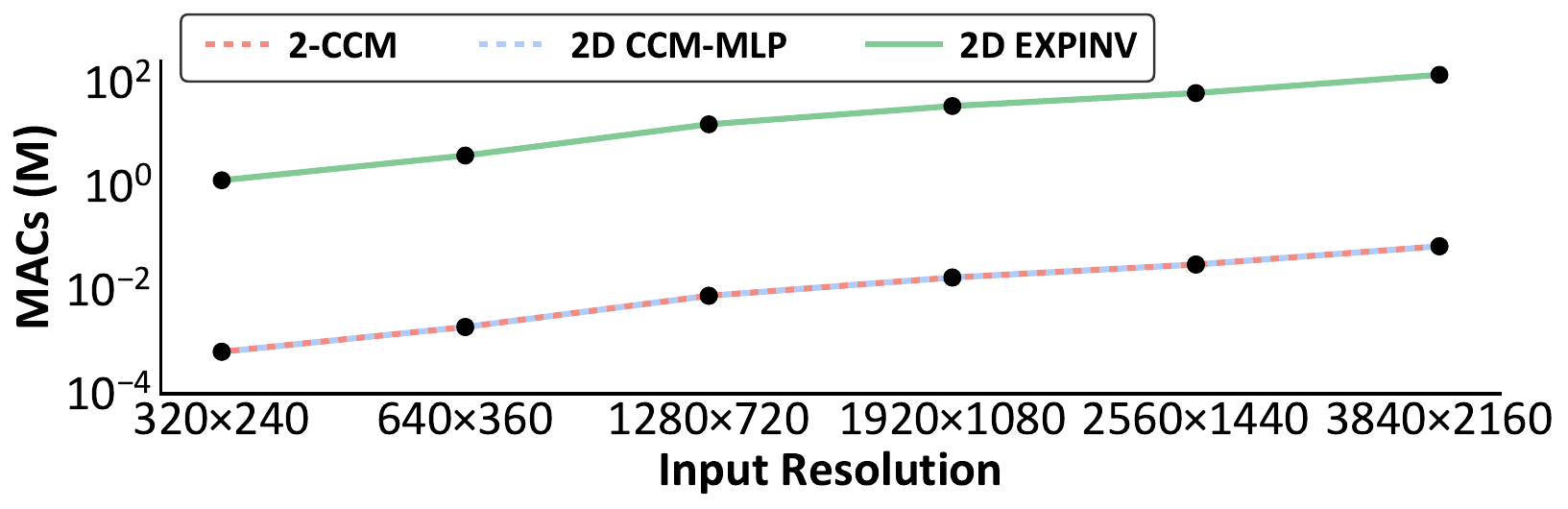}
\caption{\textbf{Computational cost comparison.}
We plot the multiply–accumulate operations (MACs, log scale) at different resolutions for 2-CCM \cite{AdobeDNG} interpolation, EXPINV \cite{expinv}, and our 2D CCM-MLP. Our CCM-MLP has nearly the same computational cost as traditional methods, since the CCM is computed once and applied to the entire image. In contrast, approaches such as EXPINV must evaluate an MLP at every pixel.}
\label{fig:performance}
\end{figure}

\noindent{\bf{Comparing CCM-MLP with per-pixel MLPs.}} We find our 2D CCM-MLP outperforms 1D MLP-based direct prediction baselines, CCCNN~\cite{cccnn} and EXPINV~\cite{expinv}. Additionally, highlighting the strength of utilizing 2D chromaticity input, our 2D chromaticity-based variants of CCCNN~\cite{cccnn} and EXPINV~\cite{expinv} achieve lower \(\Delta E_{2000}\) in many cases and surpass CCM-MLP under challenging illumination conditions (higher percentiles). This performance gain arises because per-pixel MLPs can model more complex mappings that depend on both input color and white point, whereas our method predicts a global transformation that depends only on the white point. However, per-pixel methods remain unsuitable for two reasons. First, CCM-MLP has a footprint of 1.5 KB and 0.01 MMACs, compared to 69 KB and 13 MMACs for per-pixel methods. CCM-MLP is efficient because it predicts a single global transform. In~\cref{fig:performance}, we show that the computational cost of a per-pixel MLP far exceeds our CCM-MLP at various input resolutions. Second, as shown in~\cref{fig:in-the-wild}, applying per-pixel MLP methods to full-size images can introduce color casts, likely because each pixel is transformed independently, leading to spatial inconsistencies. In contrast, CCM-MLP applies a single transformation across the image, ensuring spatial consistency.

\begin{table*}[b!]

\centering
\caption{\textbf{Ablation of CCM-MLP.}
We evaluate CCM-MLP on Pixel camera images from both the lightbox and in-the-wild scenes. We test 1D and 2D input coordinates with and without augmentation. Our results show that using xy chromaticity yields the lowest angular error among all white point parameterizations. Finally, we extend our analysis to higher-dimensional mappings using the xy chromaticity white point.}
\label{tab:ablation}

\resizebox{0.8\linewidth}{!}{%
\begin{tabular}{@{}lcccccc@{\hskip 3pt}}
\toprule
\multirow{2}{*}[-6pt]{\shortstack{\textbf{Input}\\\textbf{coord.}}} &
\multirow{2}{*}[-10pt]{\textbf{Mapping}} &
\multirow{2}{*}[-6pt]{\shortstack{\textbf{Noise}\\\textbf{Aug.}}} &
\multicolumn{2}{c}{\textbf{Lightbox}} &
\multicolumn{2}{c}{\textbf{In-the-wild}} \\
\cmidrule(lr){4-5} \cmidrule(lr){6-7}
& & & \textbf{Ang. ($\boldsymbol{\degree}$)} & \textbf{$\boldsymbol{\Delta E}$} &
\textbf{Ang. ($\boldsymbol{\degree}$)} & \textbf{$\boldsymbol{\Delta E}$} \\
\midrule

1D (CCT) & Linear $(3 \times 3)$ & \xmark &                      4.11 &                      8.48 &                      3.68 &  \cellcolor{tabthird}7.05 \\
 \midrule
\multirow{2}{*}{2D (raw)}& Linear $(3 \times 3)$ & \xmark &                      3.58 &                      7.99 &                      4.02 &                      7.81 \\
 & Linear $(3 \times 3)$ & \cmark &                      3.75 &                      8.05 &                      4.14 &                      8.33 \\
 \midrule

\multirow{6}{*}{2D (xy)} & Linear $(3 \times 3)$ & \xmark &                      \cellcolor{tabthird}3.53 &                      \cellcolor{tabthird}7.92 &                      3.64 &                      7.33 \\
 & Linear $(3 \times 3)$ & \cmark &                      3.60 &                      7.95 & \cellcolor{tabsecond}3.32 &  \cellcolor{tabfirst}6.71 \\

 & Polynomial $(3 \times 9)$ & \cmark &                      3.67 &                      8.78 &                      3.67 &                      8.78 \\
 & Root Polynomial $(3 \times 6)$ & \xmark & \cellcolor{tabfirst}3.13 &  \cellcolor{tabfirst}7.45 &  \cellcolor{tabthird}3.40 &                       7.15 \\
 & Root Polynomial $(3 \times 6)$ & \cmark &  \cellcolor{tabsecond}3.27 &  \cellcolor{tabsecond}7.73 &  \cellcolor{tabfirst}3.07 & \cellcolor{tabsecond}6.77 \\

\bottomrule
\end{tabular}
}
\end{table*}

\noindent{\bf{Ablation of CCM-MLP.}} In~\cref{tab:ablation}, we ablate three design choices. First, we compare specifying the whitepoint in 2D xy chromaticity, which yields better performance than using 1D CCT space or 2D raw space. Second, we find that our noise augmentation strategy is crucial for our method to generalize to out-of-domain illuminant chromaticities. Finally, we compare the performance of predicting different transforms and find that linear and root-polynomial mappings yield the best results. We further ablate CCM-MLP with varying layer and node counts in Supplementary \cref{sec:supp_model_abation}.

\noindent\textbf{Sensitivity to imperfect input white point.}~Our experiments assume that the ground-truth illuminant is known to isolate the performance of various CCM prediction methods. In practice, this assumption does not hold, as white-balance estimation modules~\cite{ccmnet, Afifi2025} are imperfect. To better assess real-world performance, {we evaluate our method (trained with ground truth white points) alongside several baselines on our lightbox dataset with controlled white point offsets, and on the in-the-wild dataset using white points derived from an off-the-shelf cross-camera white point estimator~\cite{ccmnet}.

\noindent\textit{Lightbox dataset.} For each illuminant, an offset of $0\degree$, $1\degree$, $2\degree$, $3\degree$, or $10\degree$ was applied in random directions. Our results show CCM-MLP achieves the lowest mean angular error when the white point is offset by $0\degree$ -- $2\degree$, which falls within the typical range of white point estimators such as CCMNet~\cite{ccmnet} and Time-Aware White Balance~\cite{Afifi2025}, which achieve $1.01\degree$ -- $2.23\degree$  mean and $0.60\degree$ -- $1.71\degree$ median angular error on datasets that include LED illuminants. When the white point offset is large ($3\degree$,$10\degree$), our 2D CCM-MLP performs similarly to 1D CCM-MLP, but still outperforms 2-CCM interpolation.

\noindent\textit{In-the-wild dataset.} 
We apply CCMNet~\cite{ccmnet}, a cross-camera illumination estimation method, to our in-the-wild data as a worst-case illumination estimate, while masking the color chart in each image. On the in-the-wild dataset, CCMNet recovers the illuminant with a mean angular error of $3.21\degree$. Again, we find that all methods benefit from using 2D chromaticity input.

\begin{table*}[h!]
\caption{\textbf{White-point error ablation.}  Angular and $\Delta E$ error for 2-CCM, 3-CCM, CCCNN, EXPINV, NN, and CCM-MLP under varying white point offsets on the lightbox and in-the-wild datasets~(Pixel). Each offset is applied in a random direction; for in-the-wild, we report results with ground truth and  CCMNet~\cite{ccmnet} estimated whitepoints ($3.21\degree$ estimation error). 2D-input methods are highlighted in blue. CCM-MLP performs best at $0\degree$--$2\degree$; 1D and 2D methods are comparable under large offsets ($\geq 3\degree$).}
\centering
\small
\setlength{\tabcolsep}{2pt} 
\resizebox{0.95\textwidth}{!}{
\begin{tabular}{lc cccccccccc cc cc}
\toprule
& & \multicolumn{10}{c}{\textbf{White-point offset:}}& \multicolumn{4}{c}{\textbf{In-the-wild}} \\ 
\cmidrule(lr){3-12}\cmidrule(lr){13-16}

\multirow[b]{2}{*}{\textbf{Method}\rule{0pt}{3.1em}}
& 
\multirow[b]{2}{*}{\textbf{Dim}\rule{0pt}{3.1em}}
&
\multicolumn{2}{c}{\textbf{0\degree}} &
\multicolumn{2}{c}{\textbf{1\degree}} &
\multicolumn{2}{c}{\textbf{2\degree}} &
\multicolumn{2}{c}{\textbf{3\degree}} &
\multicolumn{2}{c}{\textbf{10\degree}} &
\multicolumn{2}{c}{\textbf{0\degree}} &
\multicolumn{2}{c}{\textbf{CCMNet\cite{ccmnet}}} \\
\cmidrule(lr){3-4}\cmidrule(lr){5-6}\cmidrule(lr){7-8}\cmidrule(lr){9-10}\cmidrule(lr){11-12}\cmidrule(lr){13-14}\cmidrule(lr){15-16}
&  &
\textbf{Ang.} & $\boldsymbol{\Delta E}$ &
\textbf{Ang.} & $\boldsymbol{\Delta E}$ &
\textbf{Ang.} & $\boldsymbol{\Delta E}$ &
\textbf{Ang.} & $\boldsymbol{\Delta E}$ &
\textbf{Ang.} & $\boldsymbol{\Delta E}$ &
\textbf{Ang.} & $\boldsymbol{\Delta E}$ &
\textbf{Ang.} & $\boldsymbol{\Delta E}$ \\
\midrule
2-CCM                       & 1D &                      4.56 &                      9.18 &                      5.08 &                      9.80 &                      6.28 &                      11.30 &                      7.87 &                      13.16 &                      21.05 &                      26.87 &                      3.86 &                      7.04 &                      6.23 & \cellcolor{tabsecond}10.51 \\
3-CCM                       & 1D &                      4.53 &                      9.16 &                      5.05 &                      9.78 &                      6.25 &                      11.27 &                      7.84 &                      13.13 &                      20.97 &                      26.80 &                      3.81 &                      7.00 &                      6.14 &  \cellcolor{tabfirst}10.42 \\
NN~\cite{nearestneighbors} & 1D &                      5.24 &                      9.53 &                      6.07 &                      10.75 &                      6.65 &                      11.72 &                      7.90 &                      13.44 &                      20.27 &                      26.99 &                      5.43 &                      9.53 &                      7.47 &                      12.69 \\
CCCNN~\cite{cccnn}         & 1D &                      4.39 &                      8.36 &                      4.93 &                      9.10 &                      5.91 &                      10.62 &  \cellcolor{tabfirst}7.13 &  \cellcolor{tabthird}12.31 &  \cellcolor{tabfirst}17.21 &  \cellcolor{tabthird}23.84 &                      3.58 &                      7.06 &                      7.96 &                      14.67 \\
EXPINV~\cite{expinv}       & 1D &                      4.31 &                      8.09 &                      4.88 &                      9.05 &                      5.98 &                      10.67 &  \cellcolor{tabthird}7.32 &                      12.42 & \cellcolor{tabsecond}17.29 & \cellcolor{tabsecond}23.80 &  \cellcolor{tabthird}3.35 &                      6.82 &  \cellcolor{tabfirst}5.54 &                      11.03 \\
CCM-MLP                    & 1D &                      4.11 &                      8.48 &                      4.61 &                      9.17 &                      5.86 &                      10.86 &                      7.46 &                      12.82 &                      19.88 &                      25.93 &                      3.68 &                      7.05 &                      6.09 &                      11.01 \\
CCM-MLP(RP)                & 1D &  \cellcolor{tabthird}3.77 &                      8.19 &  \cellcolor{tabthird}4.42 &                      9.00 &  \cellcolor{tabthird}5.80 &                      10.72 &                      7.52 &                      12.71 &                      18.50 &                      24.91 &                      3.45 &                      7.63 &                      6.04 &                      11.37 \\\cellcolor{blue!30}NN~\cite{nearestneighbors} &\cellcolor{blue!30}2D &                      4.30 &                      8.74 &                      5.27 &                      9.94 &                      7.03 &                      12.03 &                      9.15 &                      14.33 &                      22.68 &                      29.08 &                      5.06 &                      9.13 &                      7.45 &                      12.61 \\
\cellcolor{blue!30}CCCNN~\cite{cccnn}         &\cellcolor{blue!30}2D &                      4.00 &                      8.07 &                      4.68 &                      8.86 &                      5.86 &  \cellcolor{tabthird}10.41 & \cellcolor{tabsecond}7.27 &  \cellcolor{tabfirst}12.20 &                      17.44 &                      23.94 &                      3.47 &                      6.79 &                      10.77 &                      15.73 \\\cellcolor{blue!30}EXPINV~\cite{expinv}       &\cellcolor{blue!30}2D &                      3.84 & \cellcolor{tabsecond}7.77 &                      4.62 & \cellcolor{tabsecond}8.70 &                      5.95 & \cellcolor{tabsecond}10.40 &                      7.40 & \cellcolor{tabsecond}12.23 &  \cellcolor{tabthird}17.31 &  \cellcolor{tabfirst}23.67 &  \cellcolor{tabfirst}3.07 &  \cellcolor{tabfirst}6.42 & \cellcolor{tabsecond}5.84 &  \cellcolor{tabthird}10.64 \\
\cellcolor{blue!30}CCM-MLP                    & \cellcolor{blue!30}2D & \cellcolor{tabsecond}3.60 &  \cellcolor{tabthird}7.95 & \cellcolor{tabsecond}4.30 &  \cellcolor{tabthird}8.82 & \cellcolor{tabsecond}5.76 &                      10.61 &                      7.54 &                      12.64 &                      20.48 &                      25.96 & \cellcolor{tabsecond}3.32 & \cellcolor{tabsecond}6.71 &                      6.01 &                      10.88 \\
\cellcolor{blue!30}CCM-MLP(RP)                &\cellcolor{blue!30}2D &  \cellcolor{tabfirst}3.27 &\cellcolor{tabfirst}7.74 &  \cellcolor{tabfirst}4.03 &  \cellcolor{tabfirst}8.59 &  \cellcolor{tabfirst}5.59 &  \cellcolor{tabfirst}10.39 &                      7.42 &                      12.42 &                      18.31 &                      24.36 &  \cellcolor{tabfirst}3.07 &  \cellcolor{tabthird}6.77 &  \cellcolor{tabthird}5.88 &                      10.79 \\
\bottomrule
\end{tabular}
}

\label{tab:wpe}

\end{table*}

\begin{figure}[tb]
  \centering
  \includegraphics[width=\linewidth]{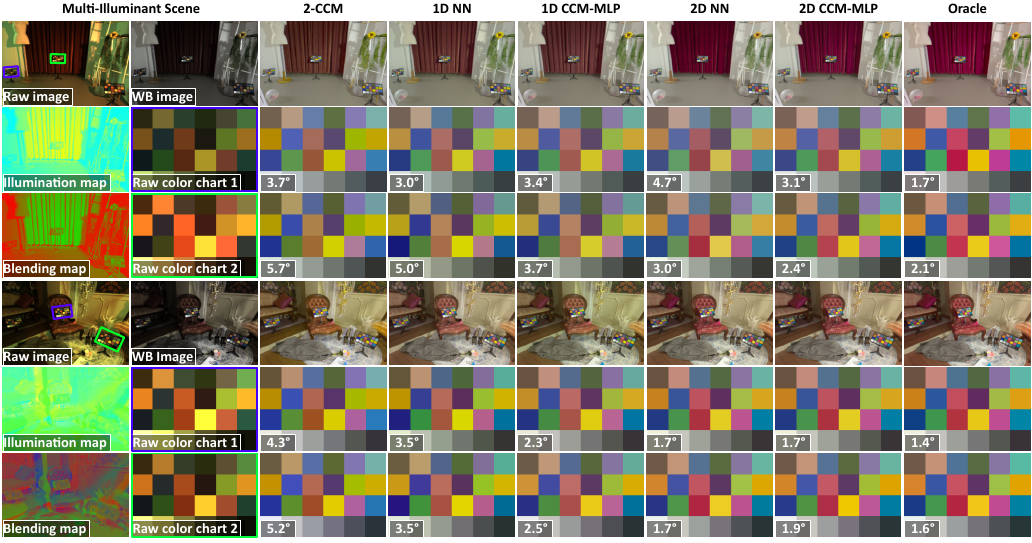}%
\caption{\textbf{Colorimetric mapping results on multi-illuminant data from LSMI~\cite{lsmi}.} We visualize two multi-illuminant scenes from the Galaxy camera. \textbf{Row structure for each half:} Top row displays raw-RGB image, white-balanced raw-RGB image, and display-referred renderings; middle row shows contrast-stretched illumination map, first raw-RGB color chart, and display-referred renderings; bottom row presents blending map, second raw-RGB color chart, and display-referred renderings. \textbf{Methods from left to right:} 2-CCM~\cite{AdobeDNG}, 1D NN~\cite{nearestneighbors}, 1D CCM-MLP, 2D NN~\cite{nearestneighbors}, 2D CCM-MLP, and Oracle. Angular errors are displayed at the bottom left of each color chart image.}%
  \label{fig:multi_illuminant}
\end{figure}

\begin{table}[t]
\centering
\caption{\textbf{LSMI results under single and mixed-illuminant settings.} Comparison of 2D chromaticity-based CCM-MLP and NN~\cite{nearestneighbors} (highlighted blue) with baselines on Galaxy and Sony data.} 
\label{tab:lsmi}
\resizebox{0.8\textwidth}{!}{%
\begin{tabular}{l c cc cc c cc cc}
\toprule
\rowcolor{blue!10} 
& & \multicolumn{4}{c}{\textbf{Single-illuminant}} & & \multicolumn{4}{c}{\textbf{Multi-illuminant}} \\
\cmidrule(lr){3-6} \cmidrule(lr){8-11}
& & \multicolumn{2}{c}{\textbf{Galaxy}} & \multicolumn{2}{c}{\textbf{Sony}} & & \multicolumn{2}{c}{\textbf{Galaxy}} & \multicolumn{2}{c}{\textbf{Sony}} \\
\cmidrule(lr){3-4} \cmidrule(lr){5-6} \cmidrule(lr){8-9} \cmidrule(lr){10-11}
\textbf{Method} & \textbf{Dim} & \textbf{Ang.} & \textbf{$\Delta E$} & \textbf{Ang.} & \textbf{$\Delta E$} & & \textbf{Ang.} & \textbf{$\Delta E$} & \textbf{Ang.} & \textbf{$\Delta E$} \\
\midrule
Oracle                     & 1D &                      2.38 &                      4.90 &                      1.59 &                      3.52 &   &                      2.06 &                      3.92 &                      1.44 &                      3.01 \\
2-CCM~\cite{AdobeDNG}      & 1D &                      5.00 &                      8.43 &                      8.40 &                      9.80 &   &                      3.88 &                      6.87 &                      8.14 &                      9.08 \\
NN~\cite{nearestneighbors} & 1D &                      4.46 &                      7.40 &                      3.07 &                      5.19 &   &                      3.23 &                      5.42 &                      2.43 &                      4.26 \\
CCM-MLP                    & 1D &                      3.77 &  \cellcolor{tabthird}6.30 &                      2.70 &                      4.55 &   &                      2.52 &  \cellcolor{tabthird}4.28 &                      1.88 & \cellcolor{tabsecond}3.40 \\
CCM-MLP (RP)                & 1D &  \cellcolor{tabthird}3.60 &                      6.54 & \cellcolor{tabsecond}2.37 &  \cellcolor{tabthird}4.45 &   &  \cellcolor{tabthird}2.35 &                      4.62 & \cellcolor{tabsecond}1.74 &  \cellcolor{tabthird}3.55 \\
\cellcolor{blue!30}NN~\cite{nearestneighbors} & \cellcolor{blue!30}2D &                      3.98 &                      6.56 &  \cellcolor{tabthird}2.48 & \cellcolor{tabsecond}4.39 &   &                      2.83 &                      4.76 &                      1.93 &                      3.61 \\
\cellcolor{blue!30}CCM-MLP                    & \cellcolor{blue!30}2D & \cellcolor{tabsecond}3.49 &  \cellcolor{tabfirst}5.82 &                      2.49 &  \cellcolor{tabfirst}4.22 &   & \cellcolor{tabsecond}2.26 &  \cellcolor{tabfirst}3.88 &  \cellcolor{tabthird}1.82 &  \cellcolor{tabfirst}3.28 \\
\cellcolor{blue!30}CCM-MLP (RP)                & \cellcolor{blue!30}2D &  \cellcolor{tabfirst}3.37 & \cellcolor{tabsecond}6.12 &  \cellcolor{tabfirst}2.31 &                      4.50 &   &  \cellcolor{tabfirst}2.14 & \cellcolor{tabsecond}4.20 &  \cellcolor{tabfirst}1.69 &                      3.63 \\
\bottomrule
\end{tabular}
}
\end{table}

\subsection{LSMI Dataset Results}

We evaluate CCM-MLP on both single- and multi-illuminant scenes from the Large-Scale Multi-Illuminant (LSMI) dataset~\cite{lsmi}. LSMI was originally proposed for white-balance; we repurpose it for colorimetric evaluation by utilizing the color charts in each scene. LSMI consists of 2,700 scenes captured using one smartphone and two cameras. Each scene contains multiple color charts and two to three distinct illuminants, providing both single- and multi-illuminant images.

\noindent{\bf{Single-illuminant results.}} We train our CCM-MLP and all baseline models on single-illuminant images from the LSMI dataset, since we do not have access to lightbox captures. We filter the LSMI dataset to include scenes where all patches on all color charts are not clipped, resulting in 1504 images for Galaxy and 1414 for Sony. We then split the data into train, validation, and test sets using 50\%, 20\%, and 30\%  of the images, respectively. For this method, we are unable to compare with CCCNN~\cite{cccnn} and EXPINV~\cite{expinv} because these methods require access to the cameras' spectral sensitivities, which are not provided with the LSMI dataset. Summarized results comparing our 2D CCM-MLP with 2-CCM interpolation (using DNG CCMs), nearest-neighbor (NN), and 1D CCM-MLP baselines are presented in \cref{tab:lsmi} (left). We find that our 2D CCM-MLP achieves the highest performance on this dataset. Additional details and results are provided in Supplementary \cref{sec:lsmi_supp}.

\noindent{\bf{Multi-illuminant results.}}
Assuming access to per-pixel illuminant information and blending maps that quantify each illuminant's contribution, we can apply our MLP trained on single-illuminant images to multi-illuminant scenes. Specifically, for each illuminant, we predict and apply the CCM to the white-balanced image to generate multiple corrected versions. We then merge these corrected images using the blending map as weights. A detailed description of this procedure is provided in Supplementary \cref{sec:lsmi_supp}. In \cref{tab:lsmi} (right), we present quantitative comparisons of our method.
Additionally, in \cref{fig:multi_illuminant}, we present qualitative results on 2- and 3-illuminant scenes from the LSMI dataset. We observe that our 2D CCM-MLP outperforms other methods quantitatively and qualitatively, demonstrating the benefits of using a 2D chromaticity representation in multi-illuminant scenarios. Additional details and results are provided in Supplementary \cref{sec:lsmi_supp}.

\begin{table*}
\centering
\small
\setlength{\tabcolsep}{1.05pt} 
\caption{\textbf{Quantitative evaluation on the NUS dataset.} We compare the 2-CCM \cite{AdobeDNG} and NN \cite{nearestneighbors} methods against our proposed CCM-MLP using angular error and $\Delta E_{2000}$. We find that the difference between 1D and 2D CCM-MLPs is small, likely because the dataset's light sources do not strongly deviate from the Planckian Locus. Methods using our proposed 2D chromaticity input are highlighted in blue.}
\resizebox{\textwidth}{!}{
\begin{tabular}{l@{\hskip 2pt}ccccccccccc}
\toprule
\multirow[b]{2}{*}[-6pt]{\textbf{Method}} & \multirow[b]{2}{*}[-6pt]{\textbf{Dim}}
& \multirow[b]{2}{*}[-6pt]{\shortstack{\textbf{Size} \\ \textbf{(KB)}}}
& \multirow[b]{2}{*}[-6pt]{\shortstack{\textbf{MACs} \\ \textbf{(M)}}}
& \multicolumn{2}{c}{\textbf{Nikon}}
& \multicolumn{2}{c}{\textbf{Olympus}}
& \multicolumn{2}{c}{\textbf{Panasonic}}
& \multicolumn{2}{c}{\textbf{Samsung}} \\
\cmidrule(lr){5-6}\cmidrule(lr){7-8}\cmidrule(lr){9-10}\cmidrule(lr){11-12}
& & & 
& \textbf{Ang.} & \textbf{$\Delta E$}
& \textbf{Ang.} & \textbf{$\Delta E$}
& \textbf{Ang.} & \textbf{$\Delta E$}
& \textbf{Ang.} & \textbf{$\Delta E$} \\
\midrule
Oracle                         & 1D &                      0.00 &                      0.01 &                      1.00 &                      2.29 &                      1.16 &                      2.60 &                      1.21 &                      2.67 &                      0.92 &                      2.08 \\
2-CCM~\cite{AdobeDNG}          & 1D &  \cellcolor{tabfirst}0.07 &  \cellcolor{tabfirst}0.01 &                      1.65 &                      2.88 &                      2.10 &                      3.50 &                      2.11 &                      3.38 &                      1.69 &                      2.83 \\
NN~\cite{nearestneighbors}     & 1D &                      3.67 &  \cellcolor{tabfirst}0.01 &                      1.38 &  \cellcolor{tabthird}2.57 &                      1.67 &                      3.07 &                      1.53 & \cellcolor{tabsecond}2.91 &                      1.31 &                      2.42 \\
CCM-MLP                        & 1D & \cellcolor{tabsecond}1.41 &  \cellcolor{tabfirst}0.01 &                      1.29 & \cellcolor{tabsecond}2.47 &                      1.47 &  \cellcolor{tabthird}2.89 &                      1.45 &  \cellcolor{tabfirst}2.80 &  \cellcolor{tabthird}1.13 &  \cellcolor{tabthird}2.22 \\
CCM-MLP(RP) & 1D &                      2.57 &  \cellcolor{tabfirst}0.01 & \cellcolor{tabsecond}1.12 &                      2.84 & \cellcolor{tabsecond}1.21 &                      3.20 & \cellcolor{tabsecond}1.18 &                      3.28 & \cellcolor{tabsecond}1.02 & \cellcolor{tabsecond}2.05 \\
\cellcolor{blue!30}NN~\cite{nearestneighbors}     &\cellcolor{blue!30}2D &                      4.04 &  \cellcolor{tabfirst}0.01 &                      1.41 &                      2.62 &                      1.64 &                      3.06 &                      1.52 & \cellcolor{tabsecond}2.91 &                      1.37 &                      2.47 \\
\cellcolor{blue!30}CCM-MLP                        &\cellcolor{blue!30}2D &  \cellcolor{tabthird}1.54 &  \cellcolor{tabfirst}0.01 &  \cellcolor{tabthird}1.28 &  \cellcolor{tabfirst}2.46 &  \cellcolor{tabthird}1.46 & \cellcolor{tabsecond}2.87 &  \cellcolor{tabthird}1.44 &  \cellcolor{tabfirst}2.80 &                      1.15 &                      2.23 \\
\cellcolor{blue!30}CCM-MLP(RP)  &\cellcolor{blue!30}2D &                      2.70 &  \cellcolor{tabfirst}0.01 &  \cellcolor{tabfirst}1.11 &                      3.17 &  \cellcolor{tabfirst}1.20 &  \cellcolor{tabfirst}2.86 &  \cellcolor{tabfirst}1.16 &  \cellcolor{tabthird}3.25 &  \cellcolor{tabfirst}1.01 &  \cellcolor{tabfirst}2.03 \\

\bottomrule
\end{tabular}
}
\label{tab:nus}
\vspace{-3pt}
\end{table*}
\subsection{NUS Dataset Results}
\label{sec:nus_dataset}
We compare on four cameras from the NUS~\cite{nus} dataset. We report quantitative results for four cameras in Table~\ref{tab:nus}. Similarly to our lightbox dataset, the 2D CCM-MLP outperforms the 2-CCM baseline but performs equally well as the 1D CCM-MLP. This is due to the illuminant distribution of the NUS dataset, which is dominated by illuminants that can be effectively modeled with CCT. These results indicate that our method matters most in the presence of LED illuminants.

\section{Conclusion}
\label{ref:conclusion}
This paper demonstrates that shifting from conventional 1D CCT-based interpolation to 2D chromaticity-based neural prediction can improve in-camera colorimetric accuracy. Additionally, our investigation finds that \textit{using the 2D chromaticity space improves accuracy for all methods} by capturing illuminant characteristics that are lost in CCT parameterization. Moreover, our CCM-MLP leverages this representation effectively while remaining lightweight and efficient. We demonstrate that our 2D CCM-MLP outperforms existing methods on a newly introduced lightbox dataset, on in-the-wild scenes, and on both single- and mixed-illuminant data from the LSMI~\cite{lsmi} dataset. Our approach has a computational cost nearly identical to current CCT-based interpolation methods, making it suitable for industry adoption and laying a foundation for next-generation camera color processing robust to illumination diversity.

\noindent{\bf{Limitations.}} Our method shares constraints with other color correction methods. First, it depends on imperfect white-balance estimators, whose errors propagate into all downstream tasks. Second, our work is limited by metamers \cite{Logvinenko2015}---distinct spectra that produce identical camera responses. In our setting with LED illuminants, we observe an increase in metamers due to their narrow-band characteristics~\cite{tedla2022}. Finally, our advantage over baseline methods diminishes near the Planckian locus (see Section~\ref{sec:nus_dataset}), indicating that our method should be interpreted as an improvement focused on modern artificial lighting rather than for all illuminants. Please see Supplementary \cref{sec:supp_limitations} for more discussion.

\section*{Acknowledgements}
We thank Trevor Canham, Jason J. Yu, and Konstantinos Derpanis for their valuable discussions and feedback throughout this project. 

\bibliographystyle{splncs04}
\bibliography{main}

\clearpage

\title{Off the Planckian Locus: Using 2D Chromaticity to Improve In-Camera Color}
\titlerunning{Off the Planckian Locus}
\author{}
\authorrunning{S.~Tedla et al.}
\institute{}
\maketitlesupplementary

\renewcommand{\thesection}{S\arabic{section}}
\renewcommand{\thefigure}{S\arabic{figure}}
\renewcommand{\thetable}{S\arabic{table}}
\renewcommand{\theequation}{S\arabic{equation}}
\setcounter{section}{0}
\setcounter{figure}{0}
\setcounter{table}{0}

In Section~\ref{sec:extensions_supp}, we present additional extensions, ablations, and limitations of our method. Sections~\ref{sec:lightbox_supp}–\ref{sec:nus_supp} include further results and details on our lightbox and in-the-wild datasets, the LSMI~\cite{lsmi} dataset, and the NUS~\cite{nus} dataset. Finally, Section~\ref{sec:cct_interpolation} provides additional background on CCT-based interpolation and its implementation.

\section{Extensions and Additional Discussion} 
\label{sec:extensions_supp}

In this section, we elaborate on several key aspects of our method. We first detail an alternative data synthesis process for efficiently generating a large-scale laboratory dataset. Next, we describe the conversion of our MLP model into a computationally efficient look-up table (LUT) suitable for mobile deployment, and we add results utilizing alternative polynomial and root-polynomial mappings. We then present a series of ablation studies to analyze the impact of different architectural choices and conclude with a discussion on the limitations of our approach.

\subsection{Synthesizing Data} 
\label{sec:synthesize}
Cycling through all illuminants in the lightbox and capturing each image takes approximately 45 minutes per camera. However, we can speed up our capture process by using a simple technique that takes advantage of the linearity when capturing RAW sensor data~\cite{Hoang}. The Telelumen lightbox mixes narrow-band LEDs to generate different SPDs. Similarly, we can simulate different camera responses by linearly combining raw images captured under individual LEDs as shown in Figure~\ref{fig:data_simulation}. Using this approach, we generated the full dataset and compared the results against training on real data. As shown in Figure~\ref{fig:data_simulation}, training the CCM-MLP on simulated data resulted in only a negligible change in angular error, with a maximum difference of $0.06 \degree$.

\begin{figure*}[t!]
  \centering
  \includegraphics[width=\linewidth]{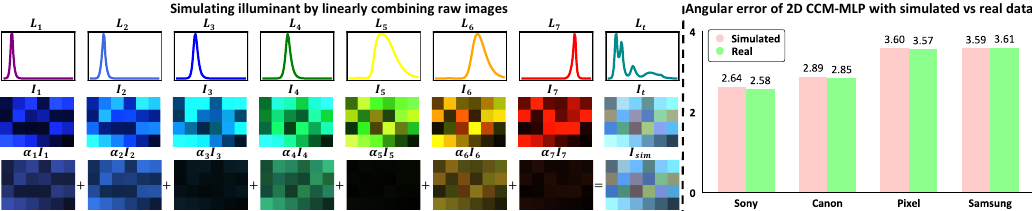}
  \caption{\textbf{Validation of the RAW data synthesis pipeline}. \textbf{Left:} A schematic of the simulation process, where a novel RAW image response $I_{sim}$ under a target illuminant $L_t$ is synthesized by linearly combining 7 pre-captured RAW images $I_i$, each illuminated by a single narrow-band LED $L_i$. The weights $\alpha_i$ correspond to the contribution of each LED required to create the target spectral power distribution. \textbf{Right:} A bar chart validating this approach by comparing the mean angular error of the 2D CCM-MLP model when trained on data from the synthesis pipeline \textit{Simulated} versus data captured sequentially under each illuminant \textit{Real}. The nearly identical performance across all four camera sensors (Sony, Canon, Pixel, Samsung) confirms the validity of the data synthesis technique, enabling the rapid generation of a large-scale, off-locus illuminant dataset.}
  \label{fig:data_simulation}
\end{figure*}

\paragraph{Simulation details.} Formally, our lightbox contains 7 narrow-band LEDs, each with a spectral power distribution (SPD) denoted by \( L_i(y) \), where \( i \in [0, 7] \). We capture 7 images \( I^i \) of the color chart, each illuminated by a single LED $i\in[1,7]$ with intensity \( \alpha_i \), where \( \alpha_i = 0 \) indicates the LED is off and \( \alpha_i = 1 \) indicates maximum brightness. The simulated spectra in the lightbox is given by 

\begin{equation}
\label{eq:light_sum_supp}
L_t(y) = \sum_{i=0}^{7} \alpha_i L_i(y).
 \end{equation}

 Under the traditional image formation model, the raw image $I$ at a location $x$ is given by

 \begin{equation}
 \label{eq:camera_basic_model}
I_t(x) = \int_\gamma L(y) R(x, y) S(x, y) \, dy,
\end{equation}

where $R(x,y)$ is the scene reflectance, and $S(x,y)$ is the sensor sensitivity. The raw color response is given by integrating across all wavelengths $y$ in the visible range $\gamma$. 

We can now substitute the definition provided in Equation~\ref{eq:light_sum_supp} into the image formation model given in Equation~\ref{eq:camera_basic_model},

\begin{equation}
\begin{aligned}
I_{sim}(x) 
&= \int_\gamma \left( \sum_{i=0}^{7} \alpha_i L_i(y)\right) R(x, y) S(x, y) \, dy \\
&= \sum_{i=0}^{7} \alpha_i \int_\gamma L_i(y) R(x, y) S(x, y) \, dy \\
&= \sum_{i=0}^{7} \alpha_i I_i(x),
\end{aligned}
\end{equation}

showing the lightbox image is equivalent to a weighted combination of the individual LED images. We note that sensor effects such as noise are not captured under this model, however, we capture at a low noise level (ISO 100).

\subsection{Look-Up-Table (LUT) Conversion}
Our 2D CCM-MLP can be converted into a look-up-table (LUT), enabling CCM prediction on mobile devices without evaluating the MLP. We perform this conversion by uniformly sampling the MLP with a 2D grid and storing the predicted CCM at each location. At test time, given a 2D chromaticity coordinate, we utilize bilinear interpolation to generate the output CCM. We try various LUT sizes, as shown in Table~\ref{tab:luts}, and find that a $20\times 20$ LUT that uses 14 KB is nearly identical to our MLP's performance. We note that the MACs remain identical across all methods because the dominant computational cost is in applying the predicted CCM rather than computing it.

\begin{table}[h]
\centering
\small
\setlength{\tabcolsep}{2pt}

\caption{\textbf{Converting 2D CCM-MLP into LUTs.} We convert our 2D CCM-MLP trained on the Pixel camera dataset into LUTs of different resolutions and find that a small $20\times 20$ LUT is comparable in performance to our 2D CCM-MLP.}

\scalebox{0.93}{

\begin{tabular}{lcccccc}
\toprule

\multirow[b]{2}{*}{\textbf{Method}\rule{0pt}{3.1em}}
  &
\multirow[b]{2}{*}{\textbf{\shortstack{\textbf{Size}\\\textbf{(KB)}}}\rule{0pt}{3.1em}}
 &

 \multirow[b]{2}{*}{\textbf{\shortstack{\textbf{MACs}\\\textbf{(M)}}}\rule{0pt}{3.1em}}
&
\multicolumn{2}{c}{\textbf{Lightbox}} &
\multicolumn{2}{c}{\textbf{In-the-wild}} \\
\cmidrule(lr){4-5} \cmidrule(lr){6-7}
&&&

\textbf{Ang.  ($\boldsymbol{^\circ}$)} & \textbf{$\boldsymbol{\Delta E}$ } &
\textbf{Ang.  ($\boldsymbol{^\circ}$)} & \textbf{$\boldsymbol{\Delta E}$ } \\
\midrule
\cellcolor{blue!30}2D CCM-MLP &  \cellcolor{tabfirst}1.54 &  \cellcolor{tabfirst}0.01 &  \cellcolor{tabfirst}3.57 &  \cellcolor{tabfirst}7.91 & \cellcolor{tabsecond}4.47 & \cellcolor{tabsecond}8.29 \\
$10\times10$ LUT             & \cellcolor{tabsecond}3.52 &  \cellcolor{tabfirst}0.01 & \cellcolor{tabsecond}3.80 &  \cellcolor{tabthird}8.37 &  \cellcolor{tabthird}4.53 &  \cellcolor{tabthird}8.33 \\
$20\times20$ LUT             &  \cellcolor{tabthird}14.06 &  \cellcolor{tabfirst}0.01 &  \cellcolor{tabfirst}3.57 & \cellcolor{tabsecond}7.98 &  \cellcolor{tabfirst}4.43 &  \cellcolor{tabfirst}8.24 \\

\bottomrule
\end{tabular}
}

\label{tab:luts}
\end{table}

\subsection{Ablations}
\label{sec:supp_model_abation}
This section presents a series of ablation studies. First, we evaluate the robustness of both our model and baselines when utilizing white points predicted by an off-the-shelf illumination estimation method. Next, we compare our model’s performance to baseline methods across different input types and examine performance under varying levels of noise. Finally, we also assess the impact of network architecture modifications on our proposed method. 

\paragraph{Utilizing white points estimated by an illumination estimation network.} We extend the white point error ablation done in Section~\ref*{sec:results} of the main paper to in-the-wild scenes by using white points derived from a cross-camera illumination estimation network, CCMNet~\cite{ccmnet}. Since this model is not fine-tuned to our camera, it represents a worst-case illumination estimation network. We directly apply it to our in-the-wild dataset to derive white points, which are then used as input to a CCM method. Observing Table~\ref{tab:awb_ablation}, we find that when using estimated white points, our proposed 2D parameterization outperforms 1D alternatives and traditional CCT-based interpolation. Like other CCT-based methods, our approach depends on the quality of the illumination estimate; and, aside from cases of complete white point estimation failure---where all methods exhibit large colorimetric errors (see 75th percentile in Table~\ref{tab:awb_ablation})---our method shows improved performance.

\begin{table*}[]
\centering
\small
\setlength{\tabcolsep}{4pt}

\caption{\textbf{Performance when using white points estimated by an illumination estimation method.} 
We evaluate performance on the Pixel in-the-wild dataset using white points predicted by the off-the-shelf illumination estimation model~\cite{ccmnet}. 
Results are reported using angular error and $\Delta E_{2000}$. In addition to the standard mean and percentile statistics, we report a trimmed mean (Tri) where the top 10\% of errors are removed to reduce the influence of scenes with large illumination estimation errors. 
We again observe that our proposed 2D chromaticity parametrization is advantageous when using estimated white points. Additionally, these results represent a worst-case illumination estimation scenario, as the method is cross-camera and not a camera-specific model. }

\resizebox{\textwidth}{!}{
\begin{tabular}{c c ccccc ccccc}
\toprule
& & \multicolumn{5}{c}{\textbf{Angular Error ($^\circ$)}} 
& \multicolumn{5}{c}{$\boldsymbol{\Delta E_{2000}}$} \\
\cmidrule(lr){3-7} \cmidrule(lr){8-12}
\textbf{Method} & \textbf{Dim}
& Mean & Tri & 25\% & 50\% & 75\%
& Mean & Tri & 25\% & 50\% & 75\% \\
\midrule
2-CCM~\cite{AdobeDNG}                         & 1D & 6.23 & 5.20 & 2.62 & 4.58 & 7.78 & \cellcolor{tabsecond}10.51 & \cellcolor{tabsecond}9.12 & 5.40 & \cellcolor{tabsecond}7.91 & \cellcolor{tabsecond}12.89 \\
3-CCM~\cite{Hakki_CVPR18}                     & 1D & 6.14 & 5.12 & 2.53 & 4.52 & \cellcolor{tabthird}7.69 & \cellcolor{tabfirst}10.42 & \cellcolor{tabfirst}9.04 & \cellcolor{tabsecond}5.25 & \cellcolor{tabfirst}7.88 & \cellcolor{tabfirst}12.83 \\
NN~\cite{nearestneighbors}                    & 1D & 7.47 & 6.42 & 3.14 & 5.60 & 9.64 & 12.69 & 11.41 & 6.65 & 10.30 & 16.00 \\
CCM-MLP                                       & 1D & 6.09 & 5.10 & 2.50 & 4.30 & 7.84 & 11.01 & 9.75 & 5.71 & 8.70 & 13.94 \\
CCM-MLP(RP)                                   & 1D & \cellcolor{tabthird}6.04 & \cellcolor{tabsecond}4.94 & \cellcolor{tabsecond}2.34 & \cellcolor{tabthird}4.17 & \cellcolor{tabsecond}7.56 & 11.37 & 10.11 & 5.92 & 9.08 & 14.51 \\
\cellcolor{blue!30}NN~\cite{nearestneighbors} & \cellcolor{blue!30}2D & 7.45 & 6.19 & 2.72 & 5.31 & 9.63 & 12.61 & 11.16 & 6.25 & 9.94 & 15.99 \\
\cellcolor{blue!30}CCM-MLP                    & \cellcolor{blue!30}2D & \cellcolor{tabsecond}6.01 & \cellcolor{tabthird}4.97 & \cellcolor{tabthird}2.35 & \cellcolor{tabsecond}4.05 & 7.75 & 10.88 & 9.56 & 5.38 & 8.56 & 13.83 \\
\cellcolor{blue!30}CCM-MLP(RP)                & \cellcolor{blue!30}2D & \cellcolor{tabfirst}5.88 & \cellcolor{tabfirst}4.75 & \cellcolor{tabfirst}2.19 & \cellcolor{tabfirst}3.93 & \cellcolor{tabfirst}7.41 & \cellcolor{tabthird}10.79 & \cellcolor{tabthird}9.43 & \cellcolor{tabthird}5.37 & \cellcolor{tabthird}8.34 & \cellcolor{tabthird}13.48 \\
\bottomrule
\end{tabular}
}


\vspace{-5pt}
\label{tab:awb_ablation}
\end{table*}

\paragraph{Input coordinate ablation.} We evaluate the performance of our method using three different coordinate types: 1D (CCT), 2D (raw), and 2D (xy) in Table~\ref{tab:noise_augment_ablation}. Our results indicate that the 2D xy chromaticity space is the most informative, consistently improving performance across all models. Furthermore, we observe that our noise augmentation improves generalization to in-the-wild illuminants, while direct pixel-prediction methods show a decline in performance under the same augmentation.

\begin{table}[t!]
\captionsetup{skip=4pt}
\setlength{\belowcaptionskip}{15pt}
\setlength{\abovecaptionskip}{2pt}
\caption{\textbf{Ablation of input coordinates.} We evaluate the performance of our CCM-MLP as well as the CCCNN and EXPINV  methods using 1D and 2D input coordinates with a noise augmentation level of $\sigma=0.05$. Results shows us that our CCM-MLP performs best with 2D chromaticity input.}
\centering
\small
\setlength{\tabcolsep}{2pt}
\renewcommand{\arraystretch}{1.05}

\scalebox{0.93}{

\begin{tabular}{l@{\hskip 2pt}cccccc}
\toprule
\multirow[b]{2}{*}{\textbf{Method}\rule{0pt}{3.1em}}&
\multirow[b]{2}{*}{\textbf{\shortstack{Input \\ coord.}}\rule{0pt}{3.1em}}&
\multirow[b]{2}{*}{\textbf{\shortstack{Noise \\ aug.}}\rule{0pt}{3.1em}}&
\multicolumn{2}{c}{\textbf{Lightbox}} &
\multicolumn{2}{c}{\textbf{In-the-wild}} \\
\cmidrule(lr){4-5}\cmidrule(lr){6-7}

& & & \textbf{Ang. ($\boldsymbol{\degree}$)} & \textbf{$\boldsymbol{\Delta E_{2000}}$} &
\textbf{Ang. ($\boldsymbol{\degree}$)} & \textbf{$\boldsymbol{\Delta E_{2000}}$} \\

\midrule
\multirow{5}{*}{CCM-MLP}              & 1D (CCT) & \xmark &                      4.03 &                      8.35 &                      4.86 &                      8.48 \\
             & 2D (raw) & \xmark &  \cellcolor{tabthird}3.58 &                      7.99 &                      5.39 &                      9.64 \\
             & 2D (raw) & \cmark &                      3.70 & \cellcolor{tabsecond}7.91 &                      4.92 &                      8.79 \\
             & 2D (xy) & \xmark &  \cellcolor{tabfirst}3.53 &  \cellcolor{tabthird}7.92 &                      5.04 &                      9.12 \\
             & 2D (xy) & \cmark & \cellcolor{tabsecond}3.57 & \cellcolor{tabsecond}7.91 &                      4.47 &                      8.29 \\
\midrule
\multirow{5}{*}{CCCNN}   & 1D (CCT) & \xmark &                      4.41 &                      8.30 &                      4.87 &                      8.54 \\
  & 2D (raw) & \xmark &                      3.90 &  \cellcolor{tabthird}7.92 &                      4.69 &                      8.40 \\
  & 2D (raw) & \cmark &                      4.18 & \cellcolor{tabsecond}7.91 &                      4.44 & \cellcolor{tabsecond}7.88 \\
  & 2D (xy) & \xmark &                      4.00 &                      8.07 &                      4.63 &                      8.19 \\
  & 2D (xy) & \cmark &                      4.14 &                      8.08 &                      4.51 &                      8.14 \\
\midrule
\multirow{5}{*}{EXPINV}
& 1D (CCT) & \xmark &                      4.29 &                      8.18 &                      4.91 &                      8.50 \\
& 2D (raw) & \xmark &                      3.78 &  \cellcolor{tabfirst}7.77 &  \cellcolor{tabfirst}4.33 &  \cellcolor{tabfirst}7.80 \\
& 2D (raw) & \cmark &                      4.24 &                      8.12 &                      4.44 &  \cellcolor{tabthird}7.92 \\
& 2D (xy) & \xmark &                      3.84 &  \cellcolor{tabfirst}7.77 & \cellcolor{tabsecond}4.35 &                      7.97 \\
& 2D (xy) & \cmark &                      3.96 &                      8.11 &  \cellcolor{tabthird}4.39 &                      8.05 \\
\bottomrule
\end{tabular}
}

\label{tab:noise_augment_ablation}
\end{table}

\paragraph{Noise augmentation ablation.} We evaluate our CCM-MLP method across varying levels of input noise applied during training and find that utilizing higher noise levels improves generalization to in-the-wild scenes at the cost of reduced accuracy on the original lightbox illuminants. Table~\ref{tab:full_noise_ablation} presents these results, highlighting the importance of our noise augmentation for generalization.

\paragraph{Nodes and layers ablation.} We conduct a detailed ablation study to determine the effect of our MLP's architecture on its performance. In this study, we vary the number of nodes per layer from 8 to 256 and the number of hidden layers from 1 to 3. The comprehensive results, including both angular error and $\Delta E_{2000}$ for the Sony, Canon, Pixel, and Samsung cameras, are presented in Table~\ref{tab:ablation_mlp_layers_full}.

The results show that performance is remarkably consistent across different network sizes. While larger models with more nodes and layers occasionally yield marginal improvements, these gains are not significant enough to justify the increase in model complexity. Given this trade-off between size and performance, we adopt the single-layer, 32-node architecture for all our experiments, as it provides a strong balance of efficiency and accuracy.

\paragraph{Calibration point ablation.}
Our paper assumes that CCT is a meaningful 1D parameterization only when calibration points are near the Planckian locus,  as in the 2-CCM and 3-CCM methods~\cite{AdobeDNG, Hakki_CVPR18}. To test this assumption, we evaluate two alternatives that use our 7 lightbox LEDs as calibration points: 7-CCM, which applies 1D CCT interpolation over the 7 LEDs, and 7-RBF, which interpolates over
2D xy space with an RBF kernel. Both perform substantially worse than baseline and proposed methods (7-CCM: $28.29^\circ$, 7-RBF: $17.76^\circ$, 3-CCM: $4.53^\circ$, 2D CCM-MLP: $3.60^\circ$; ~\cref{tab:main_results}, Pixel). This confirms that expanding to a 2D parameterization and learning a CCM-MLP is a better approach when calibration points are far from the Planckian locus.

\begin{table}[t!]
\captionsetup{skip=4pt}
\setlength{\belowcaptionskip}{5pt}
\setlength{\abovecaptionskip}{2pt}
\centering
\small
\setlength{\tabcolsep}{4pt}
\renewcommand{\arraystretch}{1.05}

\caption{\textbf{Ablation of noise augmentation.} We evaluate the performance of the 2D CCM-MLP using xy chromaticity input on the Pixel camera lightbox dataset under varying noise augmentation levels applied during training.}

\scalebox{0.93}{
\begin{tabular}{cccccc}
\toprule
\multirow[b]{2}{*}{\textbf{Method}\rule{0pt}{3.1em}}
&
\multirow[b]{2}{*}{\textbf{\shortstack{Noise aug. \\ level ($\sigma$)}}\rule{0pt}{3.1em}}
 &
\multicolumn{2}{c}{\textbf{Lightbox}} &
\multicolumn{2}{c}{\textbf{In-the-wild}} \\
\cmidrule(lr){3-4}\cmidrule(lr){5-6}

& & \textbf{Ang. ($\boldsymbol{\degree}$)} & \textbf{$\boldsymbol{\Delta E}$} &
\textbf{Ang. ($\boldsymbol{\degree}$)} & \textbf{$\boldsymbol{\Delta E}$} \\
\midrule
\multirow{4}{*}{CCM-MLP}& 0.00 &  \cellcolor{tabfirst}3.53 &  \cellcolor{tabfirst}7.92 &                      5.04 &                      9.12 \\
& 0.05 & \cellcolor{tabsecond}3.57 & \cellcolor{tabsecond}7.91 & \cellcolor{tabsecond}4.47 & \cellcolor{tabsecond}8.29 \\
& 0.10 &  \cellcolor{tabthird}3.80 &  \cellcolor{tabthird}8.17 &  \cellcolor{tabfirst}4.26 &  \cellcolor{tabfirst}7.89 \\
& 0.20 &                      3.97 &                      8.33 &  \cellcolor{tabthird}4.55 &  \cellcolor{tabthird}8.25 \\
\bottomrule
\end{tabular}
}
\label{tab:full_noise_ablation}
\end{table}

\begin{table*}[]
\centering
\vspace{-12pt}
\small
\setlength{\tabcolsep}{1pt} 

\caption{\textbf{Ablation of CCM-MLP with varying node and layer sizes.} Our 2D CCM-MLP with xy chromaticity input and noise augmentation is evaluated on camera data from lightbox scenes with varying node and layer counts. Our chosen configuration (highlighted in blue) uses 32 nodes and one layer, and is lightweight yet still competitive in performance.}

\resizebox{\textwidth}{!}{
\begin{tabular}{c@{\hskip 2pt}c@{\hskip 2pt}c cccc cccc cccc cccc}
\toprule
\multicolumn{3}{c}{} & \multicolumn{16}{c}{\textbf{Angular Error ($\boldsymbol{\degree}$)}} \\
\cmidrule(lr){4-19}
& & & \multicolumn{4}{c}{\textbf{Sony}} & \multicolumn{4}{c}{\textbf{Canon}} & \multicolumn{4}{c}{\textbf{Pixel}} & \multicolumn{4}{c}{\textbf{Samsung}} \\
\cmidrule(lr){4-7} \cmidrule(lr){8-11} \cmidrule(lr){12-15} \cmidrule(lr){16-19}
\cmidrule(lr){4-7} \cmidrule(lr){8-11}
\cmidrule(lr){12-15} \cmidrule(lr){16-19}
\textbf{Nodes} & \textbf{Layers} & \textbf{Size (KB)}
& Mean & 25\% & 50\%  & 90\%
& Mean & 25\% & 50\% & 90\%
& Mean & 25\% & 50\% & 90\%
& Mean & 25\% & 50\%  & 90\% \\
\midrule
 8 & 1 &  \cellcolor{tabfirst}0.41 &                      2.70 &  \cellcolor{tabfirst}0.78 &  \cellcolor{tabthird}1.68 &                      6.31 &                      3.24 &                      1.03 &                      1.98 &                      7.56 &                      3.83 &                      1.32 &                      2.52 &                      8.67 &                      3.80 &                      1.30 &                      2.47 &                      8.64 \\
8 & 2 & \cellcolor{tabsecond}0.69 &                      2.65 & \cellcolor{tabsecond}0.80 &                      1.69 &                      6.20 &                      2.93 &                      0.87 &                      1.91 &                      6.76 &                      3.69 &                      1.28 &                      2.41 &                      8.30 &                      3.70 &                      1.26 &                      2.42 &                      8.38 \\
8 & 3 &                      0.97 &                      2.66 &  \cellcolor{tabthird}0.81 &                      1.71 &                      6.14 &                      2.96 &                      0.89 &                      1.89 &                      6.80 &                      3.63 &                      1.25 &                      2.42 &                      8.28 &                      3.70 &                      1.26 &                      2.45 &                      8.46 \\
16 & 1 &  \cellcolor{tabthird}0.79 &                      2.73 &                      0.86 &                      1.73 &                      6.33 &                      2.93 &                      0.87 &                      1.84 &  \cellcolor{tabthird}6.64 &                      3.65 & \cellcolor{tabsecond}1.19 &                      2.41 &                      8.30 &                      3.68 &                      1.24 &                      2.43 &                      8.46 \\
16 & 2 &                      1.85 &                      2.65 &                      0.83 &                      1.72 &                      6.08 & \cellcolor{tabsecond}2.88 &                      0.87 &                      1.85 & \cellcolor{tabsecond}6.59 &                      3.59 & \cellcolor{tabsecond}1.19 &                      2.41 &                      8.04 &                      3.64 & \cellcolor{tabsecond}1.20 &                      2.45 &                      8.23 \\
16 & 3 &                      2.91 &  \cellcolor{tabthird}2.64 &                      0.85 &  \cellcolor{tabthird}1.68 & \cellcolor{tabsecond}6.00 & \cellcolor{tabsecond}2.88 &  \cellcolor{tabthird}0.85 &                      1.83 & \cellcolor{tabsecond}6.59 &                      3.60 &                      1.24 &  \cellcolor{tabthird}2.39 &                      8.20 &                      3.61 &                      1.24 & \cellcolor{tabsecond}2.38 &                      8.21 \\
32 & 1 &                      1.54 &                      2.66 &                      0.82 &                      1.70 &                      6.14 &  \cellcolor{tabthird}2.89 &                      0.87 &                      1.85 &                      6.67 &                      3.60 &                      1.22 & \cellcolor{tabsecond}2.38 &                      8.18 &                      3.60 & \cellcolor{tabsecond}1.20 &                      2.43 &                      8.23 \\
32 & 2 &                      5.66 & \cellcolor{tabsecond}2.63 &                      0.83 &                      1.70 &  \cellcolor{tabfirst}5.98 &                      2.94 & \cellcolor{tabsecond}0.84 &                      1.87 &                      6.79 & \cellcolor{tabsecond}3.56 &  \cellcolor{tabfirst}1.17 &  \cellcolor{tabfirst}2.35 &                      7.99 &                      3.60 &  \cellcolor{tabfirst}1.19 &  \cellcolor{tabthird}2.40 &                      8.28 \\
32 & 3 &                      9.79 &                      2.65 &                      0.84 &                      1.72 & \cellcolor{tabsecond}6.00 &  \cellcolor{tabfirst}2.85 & \cellcolor{tabsecond}0.84 &  \cellcolor{tabfirst}1.80 &  \cellcolor{tabfirst}6.56 &                      3.73 &                      1.25 &  \cellcolor{tabthird}2.39 &                      8.21 &                      3.68 &                      1.24 &                      2.43 &                      8.16 \\
64 & 1 &                      3.04 &                      2.66 &                      0.85 &                      1.72 &                      6.09 &  \cellcolor{tabthird}2.89 &                      0.86 &                      1.86 &                      6.78 &                      3.58 &  \cellcolor{tabthird}1.20 &                      2.45 &                      7.98 &                      3.60 &  \cellcolor{tabthird}1.21 &                      2.48 &                      8.12 \\
64 & 2 &                      19.29 &  \cellcolor{tabfirst}2.62 &                      0.82 & \cellcolor{tabsecond}1.64 &                      6.11 &  \cellcolor{tabthird}2.89 &  \cellcolor{tabfirst}0.83 &  \cellcolor{tabfirst}1.80 &                      6.74 &  \cellcolor{tabfirst}3.55 &                      1.26 &  \cellcolor{tabthird}2.39 & \cellcolor{tabsecond}7.88 &                      3.64 &                      1.24 &                      2.44 &                      8.05 \\
64 & 3 &                      35.54 &                      2.65 &                      0.82 &  \cellcolor{tabthird}1.68 &  \cellcolor{tabthird}6.06 &  \cellcolor{tabthird}2.89 &  \cellcolor{tabthird}0.85 &  \cellcolor{tabthird}1.82 &                      6.74 &                      3.63 &  \cellcolor{tabthird}1.20 &  \cellcolor{tabthird}2.39 &                      8.13 &  \cellcolor{tabfirst}3.52 & \cellcolor{tabsecond}1.20 &  \cellcolor{tabfirst}2.37 & \cellcolor{tabsecond}7.87 \\
128 & 1 &                      6.04 &  \cellcolor{tabthird}2.64 &                      0.84 &  \cellcolor{tabthird}1.68 &                      6.09 & \cellcolor{tabsecond}2.88 &  \cellcolor{tabfirst}0.83 &                      1.83 &                      6.74 &  \cellcolor{tabthird}3.57 & \cellcolor{tabsecond}1.19 &                      2.42 &  \cellcolor{tabthird}7.90 &                      3.64 &                      1.23 &                      2.43 &                      8.21 \\
128 & 2 &                      70.54 &  \cellcolor{tabthird}2.64 &                      0.84 &  \cellcolor{tabfirst}1.63 &                      6.14 & \cellcolor{tabsecond}2.88 & \cellcolor{tabsecond}0.84 & \cellcolor{tabsecond}1.81 &                      6.69 & \cellcolor{tabsecond}3.56 &                      1.24 &                      2.44 &  \cellcolor{tabfirst}7.83 &  \cellcolor{tabthird}3.56 &                      1.23 &                      2.45 &  \cellcolor{tabthird}8.00 \\
128 & 3 &                      135.04 &                      2.67 &  \cellcolor{tabthird}0.81 &                      1.69 &                      6.10 &  \cellcolor{tabthird}2.89 &  \cellcolor{tabthird}0.85 &  \cellcolor{tabthird}1.82 &                      6.67 & \cellcolor{tabsecond}3.56 &                      1.25 &  \cellcolor{tabfirst}2.35 &                      8.03 & \cellcolor{tabsecond}3.55 &                      1.25 &                      2.42 &  \cellcolor{tabfirst}7.85 \\
\bottomrule
\end{tabular}
}
\vspace{-5pt}
\label{tab:ablation_mlp_layers_full}
\end{table*}

\subsection{Limitations}
\label{sec:supp_limitations}
Metamers, unique spectral distributions that produce identical camera responses, pose a fundamental challenge for all color constancy methods~\cite{Logvinenko2015}. They arise naturally from projecting continuous spectral power distributions into three-channel color spaces~\cite{wyszecki1982color}. This phenomenon is especially pronounced under narrow-band or highly saturated illuminants, where the limited range of reflected wavelengths increases the likelihood that distinct spectra produce identical camera responses~\cite{tedla2022}. This makes metamerism a source of ambiguity that can not be resolved with traditional color constancy techniques~\cite{Logvinenko2015}. An illustration of this behavior can be seen in Figure~\ref{fig:metamers}.

\begin{figure*}

    \centering
    \includegraphics[width=0.8\linewidth]{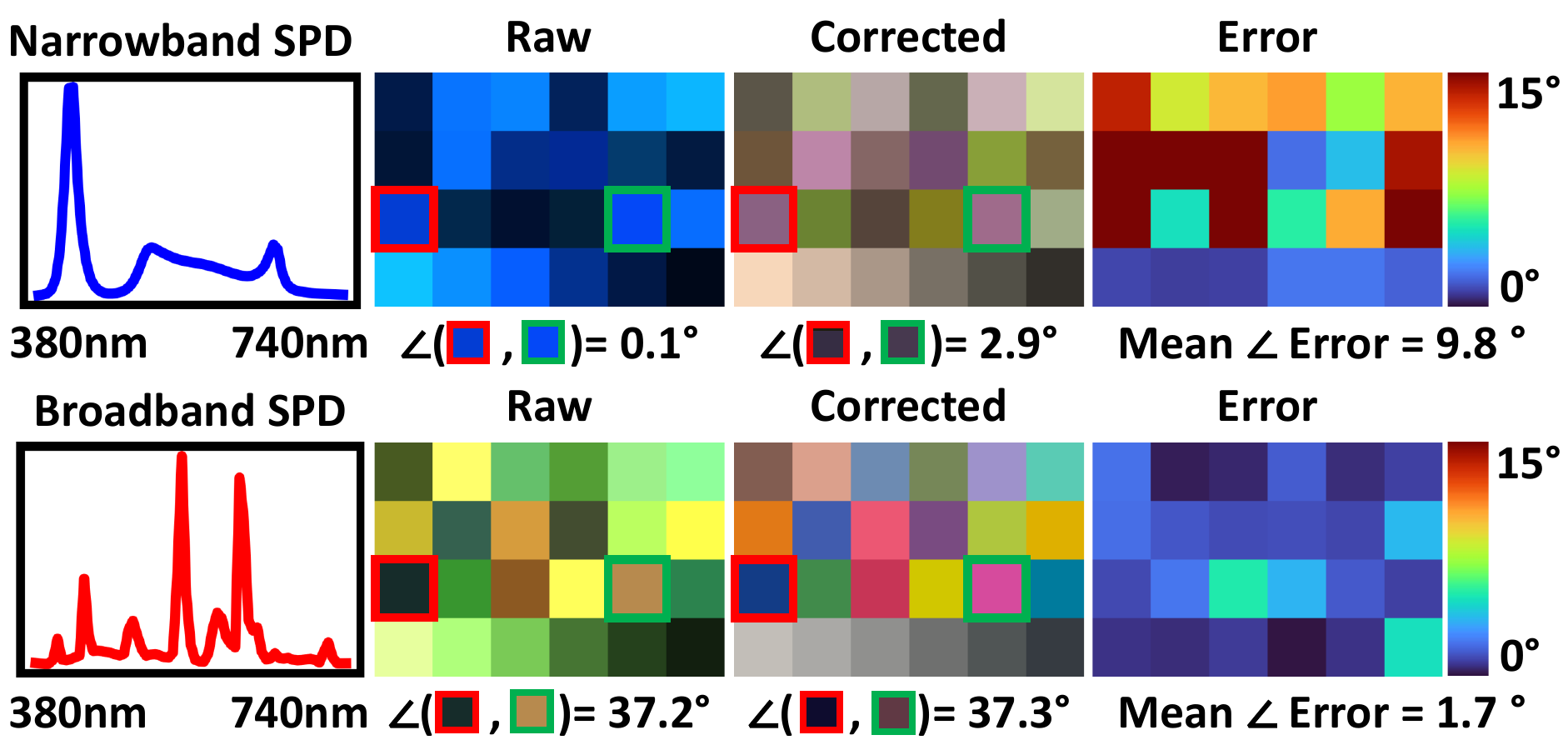}
    \captionof{figure}{\textbf{Limitations from metameric sensor responses.} 
  Metamers are distinct spectral distributions that produce identical camera responses. Metamers can occur under all illuminants, but are more common under narrowband illumination (top). In this scenario, two patches with different reflectances (highlighted in red and green) yield nearly identical raw responses, resulting in colorimetric failure for our 2D CCM-MLP. Although our method fails, this is an inherent limitation for all color constancy methods~\cite{Logvinenko2015}. Color differences are quantified using angular error.}

    \label{fig:metamers}
\end{figure*}




\section{Our Lightbox and In-The-Wild Dataset} 

We provide additional descriptions and results on our lightbox and in-the-wild dataset. Additionally, please see the supplementary material for a montage of the raw captures from these datasets. 

\label{sec:lightbox_supp}

\begin{figure*}[t]
    \centering
    \includegraphics[width=\linewidth]{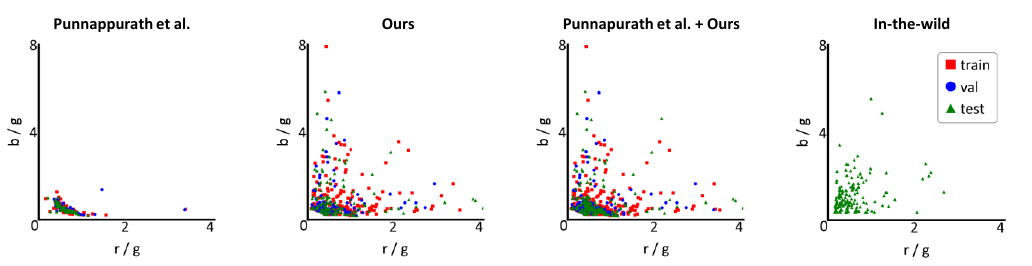}
    \caption{We visualize the chromaticities of four illuminant sets: those from Punnapurath et al.~\cite{Hoang}, those we sample from a Dirichlet distribution, the combined training set (Punnapurath et al.~and  ours), and the in-the-wild testing set. Our lightbox illuminants span a broader chromaticity region that contains the in-the-wild illuminants.} 
    
    \label{fig:chroma_plots}
\end{figure*}

\subsection{Dataset Details}

\paragraph{Lightbox dataset.} The lightbox dataset contains 790 illuminants sourced from two datasets. The first consists of 390 real illuminants from Punnappurath et al.~\cite{Hoang}, with 190 training, 78 validation, and 117 test samples as originally defined. The second source comprises 400 synthetic illuminants generated by linearly combining the seven Telelumen illuminant SPDs. We generate these illuminants by sampling 400 seven-dimensional weight vectors from a Dirichlet distribution~\cite{dirichlet}, where each element of a vector corresponds to the contribution of one illuminant to the synthetic SPD. To reduce bias toward clustered LEDs, the Dirichlet concentration parameters are adjusted and normalized to sum to 1, ensuring that the illuminants span a larger portion of the raw chromaticity space, as seen in Figure \ref{fig:chroma_plots}. These weighted vectors are then used to compute the final synthetic illuminants via a linear combination of the lightbox LED illuminant SPDs. Finally, we split the synthetic dataset into 200 training, 80 validation, and 120 test illuminants. We combine both datasets, resulting in 395 training, 158 validation, and 237 test illuminants. Please see the supplementary montage for a preview of the raw images.

\subsection{Data capture} 
\label{sec:supp_data_capture}

\paragraph{In-the-wild dataset.}

As described in the main paper, we further evaluated our method on a small in-the-wild dataset captured using mobile phones (Pixel and Samsung). We captured images with ISO level 50 on Samsung and ISO 21 on Pixel as well as using low exposure to ensure that no color patches were clipped. Our final dataset comprises 150 images per camera, containing identical scenes and lighting conditions.
\subsection{Additional Results} 

\paragraph{Lightbox dataset.} We provide additional qualitative results for all cameras in our lightbox dataset in Figure~\ref{fig:lightbox_supp1}. Our method generalizes well across a wide range of illuminants, and 2D chromaticity-based approaches consistently outperform 1D inputs.

\paragraph{In-the-wild dataset.} Our 2D CCM-MLP generalizes well to various in-the-wild scenes as can be seen qualitatively in Figures~\ref{fig:wild_supp1}-\ref{fig:wild_supp2} and quantitatively in Table~\ref{tab:wild_supp}. CCM-MLP performs worse than our proposed 2D EXPINV variant on color chart evaluations in the wild. However, we observe that EXPINV often introduces color casts in full-size images, likely because it is a spatially varying per-pixel transform that doesn't enforce global consistency. Please see the supplementary montage to view the raw captured images.

\begin{table*}[]
\centering
\small
\setlength{\tabcolsep}{1pt}

\caption{\textbf{Quantitative evaluation of in-the-wild dataset.}
We compare angular error and $\Delta E_{2000}$ of traditional 2-CCM/3-CCM interpolation~\cite{AdobeDNG, Hakki_CVPR18}, NN~\cite{nearestneighbors}, CCCNN~\cite{cccnn}, EXPINV~\cite{expinv}, and our proposed CCM-MLP. For NN, CCCNN, EXPINV, and CCM-MLP, we evaluate both 1D CCT-based and 2D chromaticity-based variants (highlighted in blue). We find using 2D chromaticity coordinates leads to lower error for all methods.}
\scalebox{0.93}{
\begin{tabular}{ l@{\hskip 2pt}c cc cccc cccc }
\toprule
\multicolumn{4}{ c }{} & \multicolumn{8}{c}{\textbf{Angular Error ($\boldsymbol{\degree}$)}} \\
\cmidrule{5-12}
\multirow[b]{2}{*}{\textbf{Method}} & \multirow[b]{2}{*}{\textbf{Dim}} 
& \multirow[b]{2}{*}{\shortstack{\textbf{Size} \\ \textbf{(KB)}}} 
& \multirow[b]{2}{*}{\shortstack{\textbf{MACs} \\ \textbf{(M)}}}
& \multicolumn{4}{c}{\textbf{Pixel}} 
& \multicolumn{4}{c}{\textbf{Samsung}} \\
\cmidrule(lr){5-8}\cmidrule(lr){9-12}

& & & 
& Mean & 25\% & 50\% & 90\%
& Mean & 25\% & 50\% & 90\% \\
\hline
Oracle                                        & 1D &                      0.00 &                      0.01 &                      1.89 &                      0.47 &                      1.12 &                      4.31 &                      2.07 &                      0.53 &                      1.25 &                      4.53 \\
2-CCM~\cite{AdobeDNG}                         & 1D &  \cellcolor{tabfirst}0.07 &  \cellcolor{tabfirst}0.01 &                      3.86 &                      1.25 &                      3.05 &                      8.04 &                      3.80 &                      1.26 &                      2.94 &                      8.19 \\
3-CCM~\cite{Hakki_CVPR18}                     & 1D & \cellcolor{tabsecond}0.11 &  \cellcolor{tabfirst}0.01 &                      3.81 &                      1.25 &                      2.97 &                      7.91 &                      3.81 &                      1.33 &                      2.93 &                      8.14 \\
NN~\cite{nearestneighbors}                    & 1D &                      15.43 &  \cellcolor{tabfirst}0.01 &                      5.43 &                      1.64 &                      4.14 &                      11.94 &                      5.41 &                      1.77 &                      3.80 &                      12.00 \\
CCCNN~\cite{cccnn}                            & 1D &                      68.51 & \cellcolor{tabsecond}13.39 &                      3.58 &                      1.35 &                      2.47 &                      7.70 &                      3.67 &                      1.37 &                      2.58 &                      8.05 \\
EXPINV~\cite{expinv}                          & 1D &                      68.53 & \cellcolor{tabsecond}13.39 &  \cellcolor{tabthird}3.35 &  \cellcolor{tabfirst}1.04 & \cellcolor{tabsecond}2.15 &                      7.81 &                      3.88 &                      1.42 &                      2.76 &                      8.47 \\
CCM-MLP                                       & 1D &  \cellcolor{tabthird}1.41 &  \cellcolor{tabfirst}0.01 &                      3.68 &                      1.43 &                      2.71 &                      7.91 &                      3.58 &  \cellcolor{tabfirst}1.18 &                      2.43 &                      8.35 \\
CCM-MLP (RP)                                  & 1D &                      2.57 &  \cellcolor{tabfirst}0.01 &                      3.45 &                      1.34 &                      2.52 &  \cellcolor{tabthird}7.01 & \cellcolor{tabsecond}3.32 &  \cellcolor{tabfirst}1.18 &  \cellcolor{tabfirst}2.24 & \cellcolor{tabsecond}7.15 \\
\cellcolor{blue!30}NN~\cite{nearestneighbors} & \cellcolor{blue!30}2D &                      16.97 &  \cellcolor{tabfirst}0.01 &                      5.06 &                      1.49 &                      3.35 &                      11.24 &                      5.15 &                      1.56 &                      3.45 &                      11.19 \\
\cellcolor{blue!30}CCCNN~\cite{cccnn}         & \cellcolor{blue!30}2D &                      69.01 &  \cellcolor{tabthird}13.48 &                      3.47 & \cellcolor{tabsecond}1.10 &                      2.28 &                      7.90 &                      3.56 &  \cellcolor{tabthird}1.21 &                      2.63 &                      7.83 \\
\cellcolor{blue!30}EXPINV~\cite{expinv}       & \cellcolor{blue!30}2D &                      69.03 &  \cellcolor{tabthird}13.48 &  \cellcolor{tabfirst}3.07 &  \cellcolor{tabfirst}1.04 &  \cellcolor{tabfirst}1.98 & \cellcolor{tabsecond}6.89 &                      3.64 &                      1.26 &                      2.62 &                      8.15 \\
\cellcolor{blue!30}CCM-MLP                    & \cellcolor{blue!30}2D &                      1.54 &  \cellcolor{tabfirst}0.01 & \cellcolor{tabsecond}3.32 &                      1.28 &                      2.42 &                      7.06 &  \cellcolor{tabthird}3.40 & \cellcolor{tabsecond}1.19 &  \cellcolor{tabthird}2.40 &  \cellcolor{tabthird}7.68 \\
\cellcolor{blue!30}CCM-MLP (RP)               & \cellcolor{blue!30}2D &                      2.70 &  \cellcolor{tabfirst}0.01 &  \cellcolor{tabfirst}3.07 &  \cellcolor{tabthird}1.12 &  \cellcolor{tabthird}2.27 &  \cellcolor{tabfirst}6.14 &  \cellcolor{tabfirst}3.18 &  \cellcolor{tabthird}1.21 & \cellcolor{tabsecond}2.25 &  \cellcolor{tabfirst}6.44 \\
\cmidrule{5-12}
\multicolumn{4}{ c }{} & \multicolumn{8}{c }{\textbf{$\boldsymbol{\Delta E_{2000}}$}} \\
\cmidrule{5-12}
Oracle                                        & 1D &                      0.00 &                      0.01 &                      4.74 &                      2.53 &                      3.99 &                      8.46 &                      4.90 &                      2.65 &                      4.13 &                      8.71 \\
2-CCM~\cite{AdobeDNG}                         & 1D &  \cellcolor{tabfirst}0.07 &  \cellcolor{tabfirst}0.01 &                      7.04 &                      4.05 &                      5.84 &                      12.24 &  \cellcolor{tabthird}7.07 &  \cellcolor{tabthird}4.00 & \cellcolor{tabsecond}5.84 &  \cellcolor{tabthird}12.04 \\
3-CCM~\cite{Hakki_CVPR18}                     & 1D & \cellcolor{tabsecond}0.11 &  \cellcolor{tabfirst}0.01 &                      7.00 &                      3.98 &                      5.81 &                      12.22 &                      7.10 &                      4.08 &  \cellcolor{tabthird}5.87 &  \cellcolor{tabthird}12.04 \\
NN~\cite{nearestneighbors}                    & 1D &                      15.43 &  \cellcolor{tabfirst}0.01 &                      9.53 &                      5.07 &                      7.81 &                      18.16 &                      9.52 &                      5.10 &                      7.78 &                      17.56 \\
CCCNN~\cite{cccnn}                            & 1D &                      68.51 & \cellcolor{tabsecond}13.39 &                      7.06 &                      4.10 &                      6.00 &                      12.36 &                      7.29 &                      4.27 &                      6.32 &                      12.34 \\
EXPINV~\cite{expinv}                          & 1D &                      68.53 & \cellcolor{tabsecond}13.39 &                      6.82 & \cellcolor{tabsecond}3.76 & \cellcolor{tabsecond}5.74 &                      12.16 &                      7.34 &                      4.38 &                      6.39 &                      12.26 \\
CCM-MLP                                       & 1D &  \cellcolor{tabthird}1.41 &  \cellcolor{tabfirst}0.01 &                      7.05 &                      4.14 &                      6.02 &                      12.00 &                      7.15 &                      4.07 &                      6.22 & \cellcolor{tabsecond}12.02 \\
CCM-MLP (RP)                                  & 1D &                      2.57 &  \cellcolor{tabfirst}0.01 &                      7.63 &                      4.50 &                      6.70 &                      13.06 &                      7.80 &                      4.76 &                      6.82 &                      13.03 \\
\cellcolor{blue!30}NN~\cite{nearestneighbors} & \cellcolor{blue!30}2D &                      16.97 &  \cellcolor{tabfirst}0.01 &                      9.13 &                      4.71 &                      7.15 &                      17.17 &                      9.39 &                      4.83 &                      7.44 &                      17.62 \\
\cellcolor{blue!30}CCCNN~\cite{cccnn}         & \cellcolor{blue!30}2D &                      69.01 &  \cellcolor{tabthird}13.48 &                      6.79 &  \cellcolor{tabthird}3.80 &                      5.77 &                      11.96 &  \cellcolor{tabfirst}6.81 &  \cellcolor{tabfirst}3.89 &  \cellcolor{tabfirst}5.79 &                      12.15 \\
\cellcolor{blue!30}EXPINV~\cite{expinv}       & \cellcolor{blue!30}2D &                      69.03 &  \cellcolor{tabthird}13.48 &  \cellcolor{tabfirst}6.42 &  \cellcolor{tabfirst}3.53 &  \cellcolor{tabfirst}5.33 &  \cellcolor{tabfirst}11.39 & \cellcolor{tabsecond}6.98 & \cellcolor{tabsecond}3.96 &                      6.06 &                      12.40 \\
\cellcolor{blue!30}CCM-MLP                    & \cellcolor{blue!30}2D &                      1.54 &  \cellcolor{tabfirst}0.01 & \cellcolor{tabsecond}6.71 &                      3.92 &  \cellcolor{tabthird}5.75 &  \cellcolor{tabthird}11.64 & \cellcolor{tabsecond}6.98 &                      4.06 &                      6.13 &  \cellcolor{tabfirst}11.56 \\
\cellcolor{blue!30}CCM-MLP (RP)               & \cellcolor{blue!30}2D &                      2.70 &  \cellcolor{tabfirst}0.01 &  \cellcolor{tabthird}6.77 &                      3.99 &                      5.94 & \cellcolor{tabsecond}11.57 &                      7.47 &                      4.45 &                      6.54 &                      12.79 \\
\bottomrule
\end{tabular}
}

\vspace{-12pt}
\label{tab:wild_supp}
\end{table*}

\section{LSMI Dataset}
\label{sec:lsmi_supp}
In this section, we provide an overview of the LSMI dataset and the preprocessing steps we utilized. We then explain how we extend CCM-MLP to multi-illuminant scenes using blending maps and present complementary quantitative and qualitative results.

\subsection{Dataset Details}

The Large-Scale Multi-Illuminant (LSMI) dataset~\cite{lsmi} comprises 2,700 scenes captured using a Samsung Galaxy Note 20 Ultra (Galaxy), Sony $\alpha$9 (Sony), and Nikon D810 (Nikon). Each scene contains multiple color charts and two or three distinct illuminants, covering both natural and indoor conditions. Each scene provides both single- and multi-illuminant images, enabling evaluation of algorithms under controlled and complex lighting scenarios.

\paragraph{Data processing.}\label{subsec:lsmiproc}
To enable evaluation, we process the scenes and organize the dataset, ensuring consistency across all splits. We extract the color chart values following the procedure detailed in Section~\ref*{sec:model} of the main paper, using a smaller $3 \times 3$ pixel region to accommodate small color charts that are far from the camera. We then partition the dataset by scene, ensuring that each scene is assigned to the training, validation, or testing sets to prevent leakage between splits. These same splits are maintained across single- and multi-illuminant experiments.

\paragraph {Single-illuminant data.} 
The single-illuminant subset is constructed from the multi-illuminant captures of each scene by applying pairwise subtraction, allowing recovery of multiple single-illuminant images per scene. We discard images with color charts containing pixels with near-zero or maximum white-level values. Applying this procedure yields 1,504 images from the Galaxy, 720 from the Nikon, and 1,414 from the Sony. Each camera's images are then split into training, validation, and test sets following the previously described partitioning, resulting in splits of 756/299/449 for Galaxy, 360/142/218 for Nikon, and 709/283/422 for Sony.

\paragraph{Multi-illuminant data.} 
To build the multi-illuminant subset, we apply the same quality-filtering criteria used in the single-illuminant subset. Each scene provides per-pixel illuminant maps $W \in \mathbb{R}^{H \times W \times 2}$ and blending maps $B_i \in \mathbb{R}^{H\times W}$ where \(B_i\) encodes the spatial contribution of illuminant \(i\) to the final image. Using the same scene splits, we build the training, validation, and test sets. For the Galaxy, this results in 358/143/216 images; for the Nikon, 176/70/106; and for the Sony, 325/130/195.

\begin{table*}[]
\centering

\small
\setlength{\tabcolsep}{1pt}

\caption{\textbf{Quantitative evaluation on LSMI dataset.} We compare 2-CCM \cite{AdobeDNG} and NN~\cite{nearestneighbors} methods against our proposed CCM-MLP using angular error and $\Delta E_{2000}$ on single- and multi-illuminant scenes. Methods using 2D chromaticity inputs are highlighted in blue.}
\resizebox{\textwidth}{!}{
\begin{tabular}{l@{\hskip 2pt}ccccccccccccccc}
\toprule
\rowcolor{blue!10}
\multicolumn{16}{c}{\textbf{Single-Illuminant}} \\
\midrule
\multicolumn{4}{c}{} & \multicolumn{12}{c}{\textbf{Angular Error ($\boldsymbol{\degree}$)}} \\
\cmidrule{5-16}
\multirow[b]{2}{*}{\textbf{Method}} & \multirow[b]{2}{*}{\textbf{Dim}} 
& \multirow[b]{2}{*}{\shortstack{\textbf{Size} \\ \textbf{(KB)}}} 
& \multirow[b]{2}{*}{\shortstack{\textbf{MACs} \\ \textbf{(M)}}}
& \multicolumn{4}{c}{\textbf{Galaxy}} 
& \multicolumn{4}{c}{\textbf{Sony}} 
& \multicolumn{4}{c}{\textbf{Nikon}} \\
\cmidrule(lr){5-8}\cmidrule(lr){9-12}\cmidrule(lr){13-16}
& & & 
& Mean & 25\% & 50\% & 90\%
& Mean & 25\% & 50\% & 90\%
& Mean & 25\% & 50\% & 90\% \\


\midrule
Oracle                                        & 1D &                      0.00 &                      0.01 &                      2.38 &                      0.61 &                      1.33 &                      5.30 &                      1.59 &                      0.55 &                      1.10 &                      3.40 &                      1.37 &                      0.48 &                      0.96 &                      3.00 \\
2-CCM~\cite{AdobeDNG}                         & 1D &  \cellcolor{tabfirst}0.07 &  \cellcolor{tabfirst}0.01 &                      5.00 &                      2.11 &                      4.15 &                      9.24 &                      8.40 &                      3.78 &                      8.77 &                      15.22 &                      8.16 &                      3.91 &                      8.49 &                      14.65 \\
NN~\cite{nearestneighbors}                    & 1D &                      29.53 &  \cellcolor{tabfirst}0.01 &                      4.46 &                      1.61 &                      3.20 &                      9.24 &                      3.07 &                      1.13 &                      2.11 &                      6.47 &                      3.27 &                      1.16 &                      2.21 &                      7.23 \\
CCM-MLP                                       & 1D & \cellcolor{tabsecond}1.41 &  \cellcolor{tabfirst}0.01 &                      3.77 &  \cellcolor{tabthird}1.19 &                      2.36 &                      8.49 &                      2.70 &                      0.97 &                      1.82 &                      5.81 &                      2.57 &                      0.90 &                      1.71 &                      5.90 \\
CCM-MLP (RP)                                  & 1D &                      2.57 &  \cellcolor{tabfirst}0.01 &  \cellcolor{tabthird}3.60 & \cellcolor{tabsecond}1.16 &  \cellcolor{tabthird}2.24 &  \cellcolor{tabthird}7.93 & \cellcolor{tabsecond}2.37 & \cellcolor{tabsecond}0.88 & \cellcolor{tabsecond}1.61 & \cellcolor{tabsecond}4.84 & \cellcolor{tabsecond}2.47 &  \cellcolor{tabthird}0.87 &  \cellcolor{tabthird}1.69 & \cellcolor{tabsecond}5.37 \\
\cellcolor{blue!30}NN~\cite{nearestneighbors} & \cellcolor{blue!30}2D &                      32.48 &  \cellcolor{tabfirst}0.01 &                      3.98 &                      1.30 &                      2.60 &                      8.65 &  \cellcolor{tabthird}2.48 &                      0.94 &                      1.77 &                      5.36 &                      3.06 &                      1.05 &                      2.07 &                      6.56 \\
\cellcolor{blue!30}CCM-MLP                    & \cellcolor{blue!30}2D &  \cellcolor{tabthird}1.54 &  \cellcolor{tabfirst}0.01 & \cellcolor{tabsecond}3.49 &  \cellcolor{tabfirst}1.11 & \cellcolor{tabsecond}2.20 & \cellcolor{tabsecond}7.45 &                      2.49 &  \cellcolor{tabthird}0.90 &  \cellcolor{tabthird}1.68 &  \cellcolor{tabthird}5.34 &  \cellcolor{tabthird}2.48 & \cellcolor{tabsecond}0.85 & \cellcolor{tabsecond}1.62 &  \cellcolor{tabthird}5.65 \\
\cellcolor{blue!30}CCM-MLP (RP)               & \cellcolor{blue!30}2D &                      2.70 &  \cellcolor{tabfirst}0.01 &  \cellcolor{tabfirst}3.37 &  \cellcolor{tabfirst}1.11 &  \cellcolor{tabfirst}2.12 &  \cellcolor{tabfirst}7.08 &  \cellcolor{tabfirst}2.31 &  \cellcolor{tabfirst}0.84 &  \cellcolor{tabfirst}1.55 &  \cellcolor{tabfirst}4.83 &  \cellcolor{tabfirst}2.39 &  \cellcolor{tabfirst}0.84 &  \cellcolor{tabfirst}1.57 &  \cellcolor{tabfirst}5.30 \\
\cmidrule{5-16}
\multicolumn{4}{c}{} & \multicolumn{12}{c}{\textbf{$\boldsymbol{\Delta E_{2000}}$}} \\
\cmidrule{5-16}
Oracle                                        & 1D &                      0.00 &                      0.01 &                      4.90 &                      2.13 &                      3.55 &                      9.52 &                      3.52 &                      1.82 &                      2.92 &                      6.59 &                      3.36 &                      1.74 &                      2.80 &                      6.36 \\
2-CCM~\cite{AdobeDNG}                         & 1D &  \cellcolor{tabfirst}0.07 &  \cellcolor{tabfirst}0.01 &                      8.43 &                      4.01 &                      6.60 &                      15.51 &                      9.80 &                      6.55 &                      10.46 &                      15.41 &                      9.63 &                      6.73 &                      10.30 &                      14.75 \\
NN~\cite{nearestneighbors}                    & 1D &                      29.53 &  \cellcolor{tabfirst}0.01 &                      7.40 &                      3.46 &                      5.54 &                      14.61 &                      5.19 &                      2.69 &                      4.11 &                      9.16 &                      5.31 &                      2.78 &                      4.25 &                      9.80 \\
CCM-MLP                                       & 1D & \cellcolor{tabsecond}1.41 &  \cellcolor{tabfirst}0.01 &  \cellcolor{tabthird}6.30 & \cellcolor{tabsecond}2.76 & \cellcolor{tabsecond}4.42 &                      12.78 &                      4.55 &  \cellcolor{tabthird}2.37 & \cellcolor{tabsecond}3.57 &                      8.08 & \cellcolor{tabsecond}4.42 & \cellcolor{tabsecond}2.29 & \cellcolor{tabsecond}3.48 & \cellcolor{tabsecond}8.35 \\
CCM-MLP (RP)                                  & 1D &                      2.57 &  \cellcolor{tabfirst}0.01 &                      6.54 &                      3.12 &                      4.97 &                      12.60 &  \cellcolor{tabthird}4.45 &                      2.44 &  \cellcolor{tabthird}3.71 & \cellcolor{tabsecond}7.57 &                      4.98 &                      2.72 &                      4.36 &                      8.66 \\
\cellcolor{blue!30}NN~\cite{nearestneighbors} & \cellcolor{blue!30}2D &                      32.48 &  \cellcolor{tabfirst}0.01 &                      6.56 &                      3.08 &                      4.91 &  \cellcolor{tabthird}12.57 & \cellcolor{tabsecond}4.39 & \cellcolor{tabsecond}2.36 &  \cellcolor{tabthird}3.71 &                      8.04 &                      5.25 &  \cellcolor{tabthird}2.65 &  \cellcolor{tabthird}4.16 &                      9.97 \\
\cellcolor{blue!30}CCM-MLP                    & \cellcolor{blue!30}2D &  \cellcolor{tabthird}1.54 &  \cellcolor{tabfirst}0.01 &  \cellcolor{tabfirst}5.82 &  \cellcolor{tabfirst}2.67 &  \cellcolor{tabfirst}4.21 &  \cellcolor{tabfirst}11.17 &  \cellcolor{tabfirst}4.22 &  \cellcolor{tabfirst}2.24 &  \cellcolor{tabfirst}3.40 &  \cellcolor{tabfirst}7.47 &  \cellcolor{tabfirst}4.33 &  \cellcolor{tabfirst}2.25 &  \cellcolor{tabfirst}3.40 &  \cellcolor{tabfirst}8.12 \\
\cellcolor{blue!30}CCM-MLP (RP)               & \cellcolor{blue!30}2D &                      2.70 &  \cellcolor{tabfirst}0.01 & \cellcolor{tabsecond}6.12 &  \cellcolor{tabthird}3.06 &  \cellcolor{tabthird}4.75 & \cellcolor{tabsecond}11.31 &                      4.50 &                      2.43 &                      3.77 &  \cellcolor{tabthird}7.78 &  \cellcolor{tabthird}4.89 &  \cellcolor{tabthird}2.65 &                      4.25 &  \cellcolor{tabthird}8.44 \\
\midrule

\rowcolor{blue!10}
\multicolumn{16}{c}{\textbf{Multi-Illuminant}} \\
\midrule
\multicolumn{4}{c}{} & \multicolumn{11}{c}{\textbf{Angular Error ($\boldsymbol{\degree}$)}} \\
\cmidrule{5-16}
\multirow[b]{2}{*}{\textbf{Method}} & \multirow[b]{2}{*}{\textbf{Dim}} 
& \multirow[b]{2}{*}{\shortstack{\textbf{Size} \\ \textbf{(KB)}}} 
& \multirow[b]{2}{*}{\shortstack{\textbf{MACs} \\ \textbf{(M)}}}
& \multicolumn{4}{c}{\textbf{Galaxy}} 
& \multicolumn{4}{c}{\textbf{Sony}} 
& \multicolumn{4}{c}{\textbf{Nikon}} \\
\cmidrule(lr){5-8}\cmidrule(lr){9-12}\cmidrule(lr){13-16}
& & & 
& Mean & 25\% & 50\% & 90\%
& Mean & 25\% & 50\% & 90\%
& Mean & 25\% & 50\% & 90\% \\
\midrule
Oracle                                        & 1D &                      0.00 &                      0.01 &                      2.06 &                      0.79 &                      1.49 &                      4.17 &                      1.44 &                      0.60 &                      1.11 &                      2.91 &                      1.45 &                      0.60 &                      1.08 &                      2.96 \\
2-CCM~\cite{AdobeDNG}                         & 1D &  \cellcolor{tabfirst}0.07 &  \cellcolor{tabfirst}0.01 &                      3.88 &                      1.76 &                      3.49 &                      7.12 &                      8.14 &                      3.64 &                      8.51 &                      14.49 &                      7.87 &                      3.66 &                      8.26 &                      14.01 \\
NN~\cite{nearestneighbors}                    & 1D &                      29.53 &  \cellcolor{tabfirst}0.01 &                      3.23 &                      1.32 &                      2.47 &                      6.52 &                      2.43 &                      0.97 &                      1.73 &                      5.12 &                      2.58 &                      1.01 &                      1.83 &                      5.55 \\
CCM-MLP                                       & 1D & \cellcolor{tabsecond}1.41 &  \cellcolor{tabfirst}0.01 &                      2.52 &                      0.97 &                      1.79 &  \cellcolor{tabthird}5.20 &                      1.88 &  \cellcolor{tabthird}0.79 &                      1.41 &                      3.87 &                      1.86 &                      0.77 &                      1.46 &                      3.70 \\
CCM-MLP (RP)                                  & 1D &                      2.57 &  \cellcolor{tabfirst}0.01 &  \cellcolor{tabthird}2.35 &  \cellcolor{tabthird}0.94 &  \cellcolor{tabthird}1.68 & \cellcolor{tabsecond}4.68 & \cellcolor{tabsecond}1.74 & \cellcolor{tabsecond}0.75 & \cellcolor{tabsecond}1.34 & \cellcolor{tabsecond}3.39 &  \cellcolor{tabthird}1.78 &  \cellcolor{tabthird}0.74 &  \cellcolor{tabthird}1.38 & \cellcolor{tabsecond}3.62 \\
\cellcolor{blue!30}NN~\cite{nearestneighbors} & \cellcolor{blue!30}2D &                      32.48 &  \cellcolor{tabfirst}0.01 &                      2.83 &                      1.10 &                      2.10 &                      5.69 &                      1.93 &                      0.82 &                      1.46 &                      3.91 &                      2.25 &                      0.91 &                      1.65 &                      4.65 \\
\cellcolor{blue!30}CCM-MLP                    & \cellcolor{blue!30}2D &  \cellcolor{tabthird}1.54 &  \cellcolor{tabfirst}0.01 & \cellcolor{tabsecond}2.26 & \cellcolor{tabsecond}0.88 & \cellcolor{tabsecond}1.61 & \cellcolor{tabsecond}4.68 &  \cellcolor{tabthird}1.82 &  \cellcolor{tabthird}0.79 &  \cellcolor{tabthird}1.39 &  \cellcolor{tabthird}3.64 & \cellcolor{tabsecond}1.77 & \cellcolor{tabsecond}0.73 & \cellcolor{tabsecond}1.34 &  \cellcolor{tabthird}3.63 \\
\cellcolor{blue!30}CCM-MLP (RP)               & \cellcolor{blue!30}2D &                      2.70 &  \cellcolor{tabfirst}0.01 &  \cellcolor{tabfirst}2.14 &  \cellcolor{tabfirst}0.85 &  \cellcolor{tabfirst}1.52 &  \cellcolor{tabfirst}4.30 &  \cellcolor{tabfirst}1.69 &  \cellcolor{tabfirst}0.72 &  \cellcolor{tabfirst}1.27 &  \cellcolor{tabfirst}3.28 &  \cellcolor{tabfirst}1.69 &  \cellcolor{tabfirst}0.67 &  \cellcolor{tabfirst}1.29 &  \cellcolor{tabfirst}3.41 \\
\cmidrule{5-16}
\multicolumn{4}{c}{} & \multicolumn{11}{c}{\textbf{$\boldsymbol{\Delta E_{2000}}$ }} \\
\cmidrule{5-16}
Oracle                                        & 1D &                      0.00 &                      0.01 &                      3.92 &                      2.14 &                      3.19 &                      6.63 &                      3.01 &                      1.77 &                      2.64 &                      5.17 &                      2.99 &                      1.72 &                      2.59 &                      5.16 \\
2-CCM~\cite{AdobeDNG}                         & 1D &  \cellcolor{tabfirst}0.07 &  \cellcolor{tabfirst}0.01 &                      6.87 &                      3.31 &                      5.20 &                      13.50 &                      9.08 &                      6.19 &                      10.04 &                      14.26 &                      8.98 &                      6.27 &                      9.96 &                      13.67 \\
NN~\cite{nearestneighbors}                    & 1D &                      29.53 &  \cellcolor{tabfirst}0.01 &                      5.42 &                      2.83 &                      4.38 &                      9.81 &                      4.26 &                      2.27 &                      3.42 &                      7.37 &                      4.22 &                      2.37 &                      3.55 &                      7.65 \\
CCM-MLP                                       & 1D & \cellcolor{tabsecond}1.41 &  \cellcolor{tabfirst}0.01 &  \cellcolor{tabthird}4.28 & \cellcolor{tabsecond}2.27 & \cellcolor{tabsecond}3.37 &                      7.60 & \cellcolor{tabsecond}3.40 & \cellcolor{tabsecond}1.99 & \cellcolor{tabsecond}2.90 & \cellcolor{tabsecond}5.72 & \cellcolor{tabsecond}3.35 & \cellcolor{tabsecond}1.94 & \cellcolor{tabsecond}2.82 & \cellcolor{tabsecond}5.81 \\
CCM-MLP (RP)                                  & 1D &                      2.57 &  \cellcolor{tabfirst}0.01 &                      4.62 &                      2.61 &                      3.87 &  \cellcolor{tabthird}7.59 &  \cellcolor{tabthird}3.55 &                      2.09 &  \cellcolor{tabthird}3.12 &  \cellcolor{tabthird}5.92 &                      4.04 &                      2.20 &                      3.62 &                      6.79 \\
\cellcolor{blue!30}NN~\cite{nearestneighbors} & \cellcolor{blue!30}2D &                      32.48 &  \cellcolor{tabfirst}0.01 &                      4.76 &                      2.57 &                      3.87 &                      8.14 &                      3.61 &  \cellcolor{tabthird}2.08 &  \cellcolor{tabthird}3.12 &                      6.18 &                      4.00 &                      2.22 &  \cellcolor{tabthird}3.34 &                      7.02 \\
\cellcolor{blue!30}CCM-MLP                    & \cellcolor{blue!30}2D &  \cellcolor{tabthird}1.54 &  \cellcolor{tabfirst}0.01 &  \cellcolor{tabfirst}3.88 &  \cellcolor{tabfirst}2.08 &  \cellcolor{tabfirst}3.12 &  \cellcolor{tabfirst}6.61 &  \cellcolor{tabfirst}3.28 &  \cellcolor{tabfirst}1.90 &  \cellcolor{tabfirst}2.82 &  \cellcolor{tabfirst}5.60 &  \cellcolor{tabfirst}3.24 &  \cellcolor{tabfirst}1.88 &  \cellcolor{tabfirst}2.75 &  \cellcolor{tabfirst}5.51 \\
\cellcolor{blue!30}CCM-MLP (RP)               & \cellcolor{blue!30}2D &                      2.70 &  \cellcolor{tabfirst}0.01 & \cellcolor{tabsecond}4.20 &  \cellcolor{tabthird}2.36 &  \cellcolor{tabthird}3.57 & \cellcolor{tabsecond}6.72 &                      3.63 &                      2.13 &                      3.22 &                      6.00 &  \cellcolor{tabthird}3.90 &  \cellcolor{tabthird}2.19 &                      3.47 &  \cellcolor{tabthird}6.50 \\
\bottomrule

\end{tabular}
}
\label{tab:lsmi_main}
\end{table*}

 \subsection{Applying CCM-MLP to Multi-Illuminant Scenes} We train CCM-MLP on single-illuminant scenes and apply it to multi-illuminant scenes given the raw image, illumination map, and blending maps. We assume all illumination information is known, because our work, CCM-MLP, focuses only on the CCM stage of the colorimetric mapping process.

First, we compute the white-balanced image \( I_{wb} \) by performing a per-pixel diagonal correction of the raw image using the illumination map $W$. Given the white point \( w_i \) (derived from the white patch of the single-illuminant image under illumination \( i \)), we then compute the corresponding CCM \( T_i \) using our CCM-MLP. The final corrected color at a pixel $x$ is obtained by blending the per-illuminant transformations:
\begin{equation}
I_{xyz}(x) = I_{wb}(x) \sum_i B_i(x) T_i,
\end{equation}
where each CCM \( T_i \) contributes proportionally to its blending weight \( B_i \). This formulation requires computing and applying a per-pixel CCM, which is computationally expensive. To reduce computation, we adopt an equivalent formulation that applies each illuminant's CCM globally to the white-balanced image and blends the results:
\begin{equation}
I_{xyz} = \sum_i B_i [I_{wb}T_i ].
\end{equation}
We adopt this formulation for our method and all baselines.

\subsection{Additional Results}

We present a comprehensive evaluation of our proposed CCM-MLP method on the LSMI dataset, comparing it with traditional 2-CCM~\cite{AdobeDNG} interpolation and NN~\cite{nearestneighbors} across both single- and multi-illuminant scenarios. The results, summarized in Table~\ref{tab:lsmi_main}, demonstrate the effectiveness of our approach using both traditional 1D and our proposed 2D chromaticity input. Performance is quantified using both angular error and $\Delta E_{2000}$.

For a more detailed and qualitative analysis, we provide a series of visual comparisons. Figures \ref{fig:lsmi_single_supp2}-\ref{fig:lsmi_single_supp1} showcase the colorimetric mapping results on the single-illuminant subset of the LSMI dataset for the Sony, Galaxy, and Nikon cameras, respectively. These figures illustrate the qualitative impact of our method on color accuracy.

Similarly, Figures \ref{fig:lsmi_mixed_supp1}-\ref{fig:lsmi_mixed_supp3} present the outcomes for the multi-illuminant scenes from the LSMI dataset, again for the Galaxy, Sony, and Nikon cameras. These visualizations, paired with the quantitative error metrics, highlight the superior performance of our CCM-MLP approach in multi-illuminant lighting conditions. Together, these tables and figures offer a thorough validation of our method's ability to achieve state-of-the-art colorimetric mapping.

\begin{table*}[]
\centering

\small
\setlength{\tabcolsep}{1.05pt} 

\caption{\textbf{Quantitative evaluation on the NUS dataset, full error distribution.} We compare the 2-CCM \cite{AdobeDNG} and NN \cite{nearestneighbors} methods against our proposed CCM-MLP  using angular error and $\Delta E_{2000}$, reporting the mean together with the 25th, 50th, and 90th percentiles; \cref{tab:nus} of the main paper reports means only. We find that the difference between 1D and 2D CCM-MLPs is small, likely because the dataset’s light sources do not strongly deviate from the Planckian Locus. Methods using our proposed 2D chromaticity input are highlighted in blue.}
\resizebox{\textwidth}{!}{
\begin{tabular}{l@{\hskip 2pt}ccccccccccccccccccc}
\toprule
\multicolumn{4}{c}{} & \multicolumn{16}{c}{\textbf{Angular Error ($\boldsymbol{\degree}$)}} \\
\cmidrule{5-20}
\multirow[b]{2}{*}{\textbf{Method}} & \multirow[b]{2}{*}{\textbf{Dim}} 
& \multirow[b]{2}{*}{\shortstack{\textbf{Size} \\ \textbf{(KB)}}} 
& \multirow[b]{2}{*}{\shortstack{\textbf{MACs} \\ \textbf{(M)}}}
& \multicolumn{4}{c}{\textbf{Nikon}} 
& \multicolumn{4}{c}{\textbf{Olympus}} 
& \multicolumn{4}{c}{\textbf{Panasonic}} 
& \multicolumn{4}{c}{\textbf{Samsung}} \\
\cmidrule(lr){5-8}\cmidrule(lr){9-12}\cmidrule(lr){13-16}\cmidrule(lr){17-20}
& & & 
& Mean & 25\% & 50\% & 90\%
& Mean & 25\% & 50\% & 90\%
& Mean & 25\% & 50\% & 90\%
& Mean & 25\% & 50\% & 90\% \\
\midrule
Oracle                                        & 1D &                      0.00 &                      0.01 &                      1.00 &                      0.34 &                      0.67 &                      2.18 &                      1.16 &                      0.38 &                      0.78 &                      2.61 &                      1.21 &                      0.46 &                      0.89 &                      2.80 &                      0.92 &                      0.30 &                      0.61 &                      2.07 \\
2-CCM~\cite{AdobeDNG}                         & 1D &  \cellcolor{tabfirst}0.07 &  \cellcolor{tabfirst}0.01 &                      1.65 &                      0.64 &                      1.16 &                      3.69 &                      2.10 &                      0.89 &                      1.62 &                      4.08 &                      2.11 &                      0.76 &                      1.61 &                      4.72 &                      1.69 &                      0.78 &                      1.27 &                      3.46 \\
NN~\cite{nearestneighbors}                    & 1D &                      3.67 &  \cellcolor{tabfirst}0.01 &                      1.38 &                      0.50 &                      0.93 &                      2.98 &                      1.67 &                      0.55 &                      1.15 &                      3.65 &                      1.53 &  \cellcolor{tabthird}0.56 &                      1.12 &                      3.42 &                      1.31 &                      0.51 &                      0.94 &                      2.76 \\
CCM-MLP                                       & 1D & \cellcolor{tabsecond}1.41 &  \cellcolor{tabfirst}0.01 &                      1.29 &                      0.52 &                      0.93 &  \cellcolor{tabthird}2.68 &                      1.47 &  \cellcolor{tabthird}0.53 &  \cellcolor{tabthird}1.05 &  \cellcolor{tabthird}2.99 &                      1.45 &                      0.60 &                      1.11 &                      3.10 &  \cellcolor{tabthird}1.13 &  \cellcolor{tabthird}0.48 &  \cellcolor{tabthird}0.81 &  \cellcolor{tabthird}2.33 \\
CCM-MLP (RP)                                  & 1D &                      2.57 &  \cellcolor{tabfirst}0.01 & \cellcolor{tabsecond}1.12 & \cellcolor{tabsecond}0.45 & \cellcolor{tabsecond}0.78 &  \cellcolor{tabfirst}2.38 & \cellcolor{tabsecond}1.21 & \cellcolor{tabsecond}0.42 &  \cellcolor{tabfirst}0.78 &  \cellcolor{tabfirst}2.56 & \cellcolor{tabsecond}1.18 & \cellcolor{tabsecond}0.49 & \cellcolor{tabsecond}0.86 & \cellcolor{tabsecond}2.49 & \cellcolor{tabsecond}1.02 &  \cellcolor{tabfirst}0.41 &  \cellcolor{tabfirst}0.73 & \cellcolor{tabsecond}2.10 \\
\cellcolor{blue!30}NN~\cite{nearestneighbors} & 2D &                      4.04 &  \cellcolor{tabfirst}0.01 &                      1.41 &  \cellcolor{tabthird}0.49 &                      0.96 &                      3.08 &                      1.64 &                      0.54 &                      1.17 &                      3.50 &                      1.52 &  \cellcolor{tabthird}0.56 &                      1.11 &                      3.46 &                      1.37 &                      0.49 &                      1.01 &                      3.02 \\
\cellcolor{blue!30}CCM-MLP                    & 2D &  \cellcolor{tabthird}1.54 &  \cellcolor{tabfirst}0.01 &  \cellcolor{tabthird}1.28 &                      0.51 &  \cellcolor{tabthird}0.90 &                      2.70 &  \cellcolor{tabthird}1.46 &                      0.54 &                      1.06 & \cellcolor{tabsecond}2.95 &  \cellcolor{tabthird}1.44 &                      0.59 &  \cellcolor{tabthird}1.10 &  \cellcolor{tabthird}3.03 &                      1.15 &                      0.49 &                      0.82 &                      2.40 \\
\cellcolor{blue!30}CCM-MLP (RP)               & 2D &                      2.70 &  \cellcolor{tabfirst}0.01 &  \cellcolor{tabfirst}1.11 &  \cellcolor{tabfirst}0.44 &  \cellcolor{tabfirst}0.77 & \cellcolor{tabsecond}2.39 &  \cellcolor{tabfirst}1.20 &  \cellcolor{tabfirst}0.39 & \cellcolor{tabsecond}0.79 &  \cellcolor{tabfirst}2.56 &  \cellcolor{tabfirst}1.16 &  \cellcolor{tabfirst}0.47 &  \cellcolor{tabfirst}0.82 &  \cellcolor{tabfirst}2.47 &  \cellcolor{tabfirst}1.01 & \cellcolor{tabsecond}0.44 & \cellcolor{tabsecond}0.74 &  \cellcolor{tabfirst}2.04 \\
\cmidrule{5-20}
\multicolumn{4}{c}{} & \multicolumn{16}{c}{\textbf{$\boldsymbol{\Delta E_{2000}}$ }} \\

\cmidrule{5-20}
Oracle                                        & 1D &                      0.00 &                      0.01 &                      2.29 &                      1.21 &                      1.94 &                      3.99 &                      2.60 &                      1.41 &                      2.21 &                      4.74 &                      2.67 &                      1.41 &                      2.26 &                      4.61 &                      2.08 &                      1.20 &                      1.79 &                      3.57 \\
2-CCM~\cite{AdobeDNG}                         & 1D &  \cellcolor{tabfirst}0.07 &  \cellcolor{tabfirst}0.01 &                      2.88 &                      1.67 &                      2.42 &                      5.14 &                      3.50 &                      2.20 &                      3.08 &                      5.62 &                      3.38 &                      2.03 &                      3.03 &                      5.67 &                      2.83 &                      1.87 &                      2.55 &                      4.80 \\
NN~\cite{nearestneighbors}                    & 1D &                      3.67 &  \cellcolor{tabfirst}0.01 &  \cellcolor{tabthird}2.57 &                      1.45 &  \cellcolor{tabthird}2.19 &  \cellcolor{tabthird}4.64 &                      3.07 &                      1.79 &                      2.66 &                      5.46 & \cellcolor{tabsecond}2.91 & \cellcolor{tabsecond}1.58 &  \cellcolor{tabthird}2.46 &                      4.98 &                      2.42 &                      1.49 &                      2.12 &                      4.06 \\
CCM-MLP                                       & 1D & \cellcolor{tabsecond}1.41 &  \cellcolor{tabfirst}0.01 & \cellcolor{tabsecond}2.47 &  \cellcolor{tabfirst}1.39 & \cellcolor{tabsecond}2.11 &  \cellcolor{tabfirst}4.24 &  \cellcolor{tabthird}2.89 &  \cellcolor{tabthird}1.70 &  \cellcolor{tabthird}2.50 &  \cellcolor{tabfirst}5.03 &  \cellcolor{tabfirst}2.80 &  \cellcolor{tabfirst}1.56 &  \cellcolor{tabfirst}2.40 & \cellcolor{tabsecond}4.87 &  \cellcolor{tabthird}2.22 &  \cellcolor{tabthird}1.38 &  \cellcolor{tabthird}1.93 &                      3.77 \\
CCM-MLP (RP)                                  & 1D &                      2.57 &  \cellcolor{tabfirst}0.01 &                      2.84 &                      1.58 &                      2.52 &                      4.87 &                      3.20 &                      1.92 &                      2.79 &                      5.54 &                      3.28 &                      1.92 &                      2.90 &                      5.71 & \cellcolor{tabsecond}2.05 &  \cellcolor{tabfirst}1.21 & \cellcolor{tabsecond}1.74 & \cellcolor{tabsecond}3.55 \\
\cellcolor{blue!30}NN~\cite{nearestneighbors} & 2D &                      4.04 &  \cellcolor{tabfirst}0.01 &                      2.62 &  \cellcolor{tabthird}1.44 &                      2.23 &                      4.76 &                      3.06 &  \cellcolor{tabthird}1.70 &                      2.53 &                      5.60 & \cellcolor{tabsecond}2.91 &                      1.62 &                      2.47 &  \cellcolor{tabthird}4.89 &                      2.47 &                      1.46 &                      2.16 &                      4.31 \\
\cellcolor{blue!30}CCM-MLP                    & 2D &  \cellcolor{tabthird}1.54 &  \cellcolor{tabfirst}0.01 &  \cellcolor{tabfirst}2.46 & \cellcolor{tabsecond}1.40 &  \cellcolor{tabfirst}2.10 & \cellcolor{tabsecond}4.28 & \cellcolor{tabsecond}2.87 & \cellcolor{tabsecond}1.65 & \cellcolor{tabsecond}2.44 & \cellcolor{tabsecond}5.08 &  \cellcolor{tabfirst}2.80 &  \cellcolor{tabthird}1.60 & \cellcolor{tabsecond}2.42 &  \cellcolor{tabfirst}4.82 &                      2.23 &                      1.41 &                      1.95 &  \cellcolor{tabthird}3.71 \\
\cellcolor{blue!30}CCM-MLP (RP)               & 2D &                      2.70 &  \cellcolor{tabfirst}0.01 &                      3.17 &                      1.62 &                      2.68 &                      5.77 &  \cellcolor{tabfirst}2.86 &  \cellcolor{tabfirst}1.54 &  \cellcolor{tabfirst}2.34 &  \cellcolor{tabthird}5.21 &  \cellcolor{tabthird}3.25 &                      1.89 &                      2.92 &                      5.74 &  \cellcolor{tabfirst}2.03 & \cellcolor{tabsecond}1.23 &  \cellcolor{tabfirst}1.72 &  \cellcolor{tabfirst}3.54 \\
\bottomrule

\end{tabular}
}
\label{tab:nus_supp}

\end{table*}

\section{NUS Dataset}
\label{sec:nus_supp}
We additionally compare on Nikon D5200 (Nikon), Olympus E-PL6 (Olympus), Panasonic Lumix DMC-GX1 (Lumix), and Samsung NX2000 (Samsung) cameras from the NUS~\cite{nus} dataset. For this dataset, we again do not have access to the cameras' spectral sensitivities, so we are unable to compare with CCCNN~\cite{cccnn} or EXPINV~\cite{expinv}. We report quantitative results for four cameras in Table~\ref{tab:nus_supp}. Similarly to our lightbox dataset, the 2D CCM-MLP outperforms the 2-CCM baseline but performs equally well as the 1D CCM-MLP. We believe this is due to the illumination distribution of the NUS dataset, which is dominated by illuminants that can be effectively modeled with CCT.

\section{CCT-Based Interpolation Methods for Traditional Planckian Illuminants}
\label{sec:cct_interpolation}

The interpolation-based colorimetric mapping approaches in classical camera processing rely on accurate estimation of the scene illuminant's correlated color temperature (CCT), specifically designed for traditional light sources that approximate the Planckian locus. We outline here the computational procedures for both two-point and three-point interpolation methods, which convert observed neutral gray responses (white points) in camera RGB space to CIE xy chromaticity coordinates. While these approaches are effective for illuminants that are near the Planckian locus, they become fundamentally problematic for modern light sources that deviate significantly from the Planckian locus.

These estimation procedures address a circularity inherent in traditional camera color processing: computing the optimal color correction matrix requires knowledge of the illuminant's CCT, yet determining this CCT necessitates a preliminary mapping from camera space to CIE-xy coordinates before the final CCM is established. However, this circular dependency assumes that all illuminants can be meaningfully characterized by a single CCT value\textemdash an assumption that breaks down for non-Planckian sources where CCT becomes ambiguous or undefined.

\subsection{Unified CCT Estimation Algorithm}
Both methods utilize factory-calibrated transforms and employ an iterative refinement strategy based on Adobe's DNG processing workflow. These approaches were designed under the assumption that scene illuminants follow the blackbody radiator curve. The key difference lies in the interpolation strategy: Two-point methods use fixed calibration points at 2500 K and 6500 K~\cite{AdobeDNG}, and Karaimer and Brown's three-point extension~\cite{Hakki_CVPR18} adds a 5000 K calibration point with nearest-neighbor pair selection. The former interpolates across the full 4000 K span between calibration points, leading to maximum errors around 5000 K where interpolated transforms deviate most from optimal values. The latter reduces maximum interpolation distance to 2500 K through adaptive pair selection, improving accuracy in the problematic mid-range temperatures. Adobe subsequently adopted the three-point approach in their DNG SDK standard, recognizing its practical benefits for traditional illuminant processing. The algorithm is presented in Algorithm~\ref{alg:find_CCT}.

\subsection{Limitations}
Despite the improvements, traditional 1D interpolation approaches have conceptual limitations when confronted with modern lighting. These fundamental constraints arise from their reliance on CCT-based characterization. For LED illuminants with chromaticity coordinates significantly displaced from the Planckian locus, CCT calculations become mathematically ill-defined. Multiple CCT values may correspond to the same chromaticity point, or no meaningful CCT may exist. Readers are referred to Ohno~\cite{Ohno} for more details.  

The prevalence of LEDs and non-Planckian illuminants in modern lighting demands a shift from interpolation-based approaches to methods capable of handling the full multidimensional complexity of modern illuminants. This motivation drives our proposed neural network architecture, which abandons CCT-based characterization in favor of learning in chromaticity space for more accurate colorimetric mapping under arbitrary illumination conditions.

\begin{algorithm}
\caption{Unified CCT Estimation: Two-Point and Three-Point Interpolation}
\label{alg:find_CCT}
\begin{algorithmic}[1]
\Require Observed neutral gray in raw-RGB: $\mathbf{n}_{RGB}$
\Require Interpolation mode: $mode \in \{TwoPoint, ThreePoint\}$
\Require Pre-calibrated CCMs: $\mathbf{T}_{2500 K}$, $\mathbf{T}_{6500 K}$, $[\mathbf{T}_{5000 K}]$ \Comment{Third CCM For three-point only}
\Require CCT look-up-function: $CCT(x,y) \rightarrow$ correlated color temperature
\Ensure Estimated CCT of scene illuminant

\State $x_{last} \leftarrow 0.34$, $y_{last} \leftarrow 0.35$ \Comment{Initial coordinates}
\State $maxIter \leftarrow 30$, $tolerance \leftarrow 10^{-7}$
\State $threshold \leftarrow 5000 K$ \Comment{Pair selection threshold For three-point}

\For{$iter = 1$ to $maxIter$}
    \State $CCT_{current} \leftarrow CCT(x_{last}, y_{last})$
    
    \If{$mode = TwoPoint$} \Comment{Traditional two-point interpolation}
        \State $g \leftarrow \frac{CCT_{current}^{-1} - 6500^{-1}}{2500^{-1} - 6500^{-1}}$
        \State $\mathbf{T}_{interpolated} \leftarrow g \cdot \mathbf{T}_{2500 K} + (1-g) \cdot \mathbf{T}_{6500 K}$
    \Else \Comment{Three-point nearest-neighbor interpolation}
        \If{$CCT_{current} < threshold$} \Comment{Use warm-neutral pair}
            \State $g \leftarrow \frac{CCT_{current}^{-1} - 5000^{-1}}{2500^{-1} - 5000^{-1}}$
            \State $\mathbf{T}_{interpolated} \leftarrow g \cdot \mathbf{T}_{2500 K} + (1-g) \cdot \mathbf{T}_{5000 K}$
        \Else \Comment{Use neutral-cool pair}
            \State $g \leftarrow \frac{CCT_{current}^{-1} - 6500^{-1}}{5000^{-1} - 6500^{-1}}$
            \State $\mathbf{T}_{interpolated} \leftarrow g \cdot \mathbf{T}_{5000 K} + (1-g) \cdot \mathbf{T}_{6500 K}$
        \EndIf
    \EndIf
    
    \State $\mathbf{XYZ} \leftarrow \mathbf{T}_{interpolated} \cdot \mathbf{n}_{RGB}$ \Comment{Apply interpolated CCM}
    \State $(x_{new}, y_{new}) \leftarrow$ Convert $\mathbf{XYZ}$ to chromaticity coordinates
    
    \If{$|x_{new} - x_{last}| + |y_{new} - y_{last}| < tolerance$}
        \State \textbf{break} \Comment{Convergence achieved}
    \EndIf
    
    \State $x_{last} \leftarrow x_{new}$, $y_{last} \leftarrow y_{new}$
\EndFor

\If{$iter = maxIter$} \Comment{Fallback averaging If no convergence}
    \State $x_{last} \leftarrow (x_{last} + x_{new}) / 2$
    \State $y_{last} \leftarrow (y_{last} + y_{new}) / 2$
\EndIf

\Return $CCT(x_{last}, y_{last})$
\end{algorithmic}
\end{algorithm}

\begin{figure*}[t]
  \centering
  \vspace{-12pt}
  \includegraphics[width=1.0\linewidth]{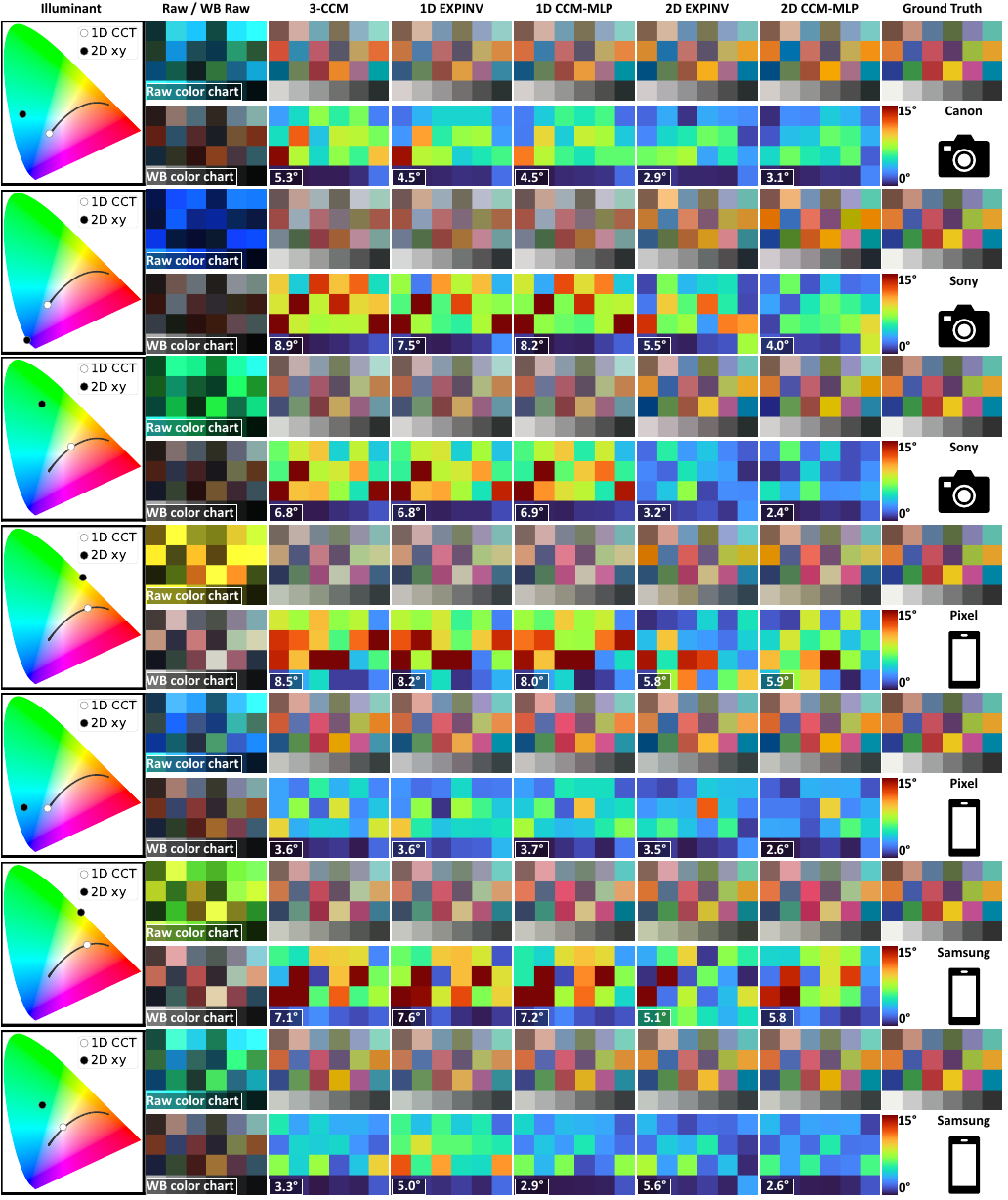}
\caption{\textbf{Colorimetric mapping results on lightbox dataset.} Results for seven different non-Planckian LED illuminant and camera combinations: two examples for Sony, two for Pixel, two for Samsung, and one for Canon. Each of the seven blocks starts with the illuminant's CCT and xy chromaticity and contains two rows: the top row shows the raw-RGB color chart, and the display-referred renderings for each method; the bottom row shows the white-balanced color chart and corresponding angular error maps. Methods: 2-CCM~\cite{Hakki_CVPR18}, 1D EXPINV~\cite{expinv}, our 1D CCM-MLP, 2D EXPINV~\cite{expinv}, our 2D CCM-MLP, and ground truth. \textbf{Key findings:} (1) Traditional 1D methods exhibit substantial errors for off-locus illuminants. (2) 2D chromaticity representation improves existing methods (compare 1D vs. 2D for EXPINV and CCM-MLP). } 
\label{fig:lightbox_supp1}
\end{figure*}

\begin{figure*}[t]
  \centering
  \includegraphics[width=1.0\linewidth]{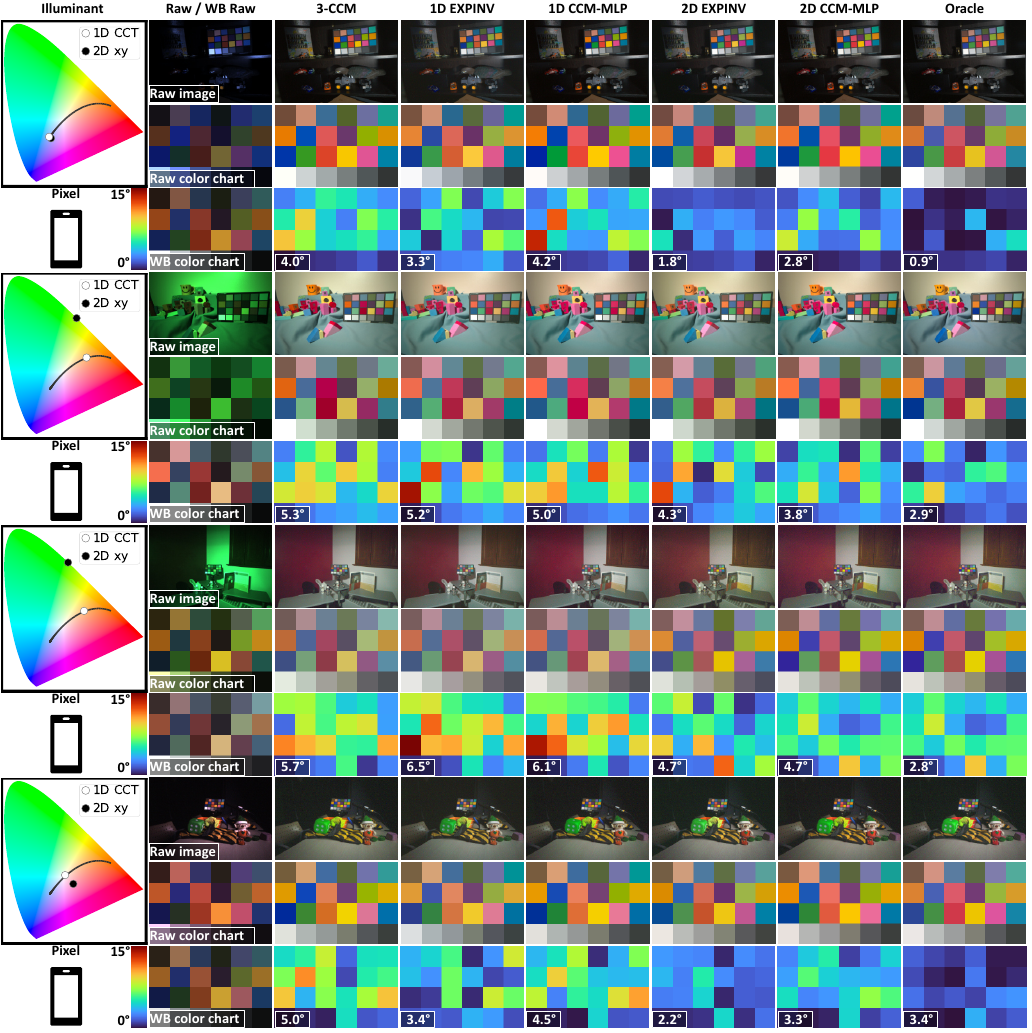}
\caption{\textbf{Colorimetric mapping results on in-the-wild captures (Pixel camera).} This figure provides an expanded set of results, supplementing the two examples shown in the main paper. Results are shown for four different light sources captured using the Pixel camera. Each section corresponds to a different illuminant, starting with its CCT and xy chromaticity, and is organized into three rows: the top row shows the raw-RGB image and display-referred renderings; the middle row shows the raw-RGB color chart and display-referred renderings; the bottom row shows the white-balanced color chart and angular error maps. Methods: 3-CCM~\cite{Hakki_CVPR18}, 1D EXPINV~\cite{expinv}, our 1D CCM-MLP, 2D EXPINV~\cite{expinv}, our 2D CCM-MLP, and Oracle. \textbf{Key findings:} (1) 2D methods outperform 1D methods under these in-the-wild lighting conditions. (2) Per-pixel methods such as EXPINV~\cite{expinv} can suffer from color casts. 
} 
\label{fig:wild_supp1}
\end{figure*}

\begin{figure*}[t]
  \centering
  \includegraphics[width=1.0\linewidth]{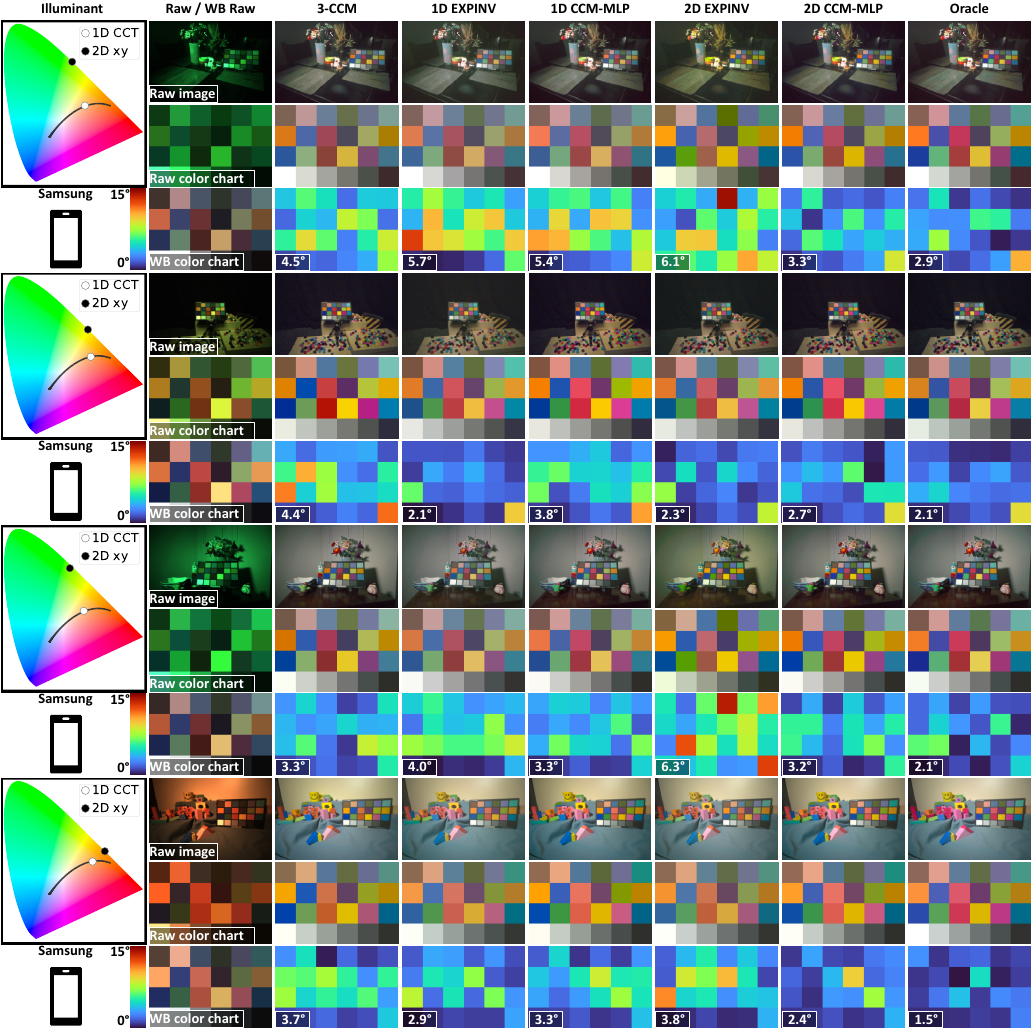}
\caption{\textbf{Colorimetric mapping results on in-the-wild captures (Samsung camera).} This figure provides an expanded set of results, supplementing the two examples shown in the main paper. Results are shown for four different light sources captured using the Samsung camera. Each section corresponds to a different illuminant, starting with its CCT and xy chromaticity, and is organized into three rows: the top row shows the raw-RGB image and display-referred renderings; the middle row shows the raw-RGB color chart and display-referred renderings; the bottom row shows the white-balanced color chart and angular error maps. Methods: 3-CCM~\cite{Hakki_CVPR18}, 1D EXPINV~\cite{expinv}, our 1D CCM-MLP, 2D EXPINV~\cite{expinv}, our 2D CCM-MLP, and Oracle. \textbf{Key findings:} (1) 2D methods outperform 1D methods under these in-the-wild lighting conditions. (2) Per-pixel methods such as EXPINV~\cite{expinv} can suffer from color casts.}
\label{fig:wild_supp2}
\end{figure*}

\begin{figure*}[t]
  \centering
  \includegraphics[width=1.0\linewidth]{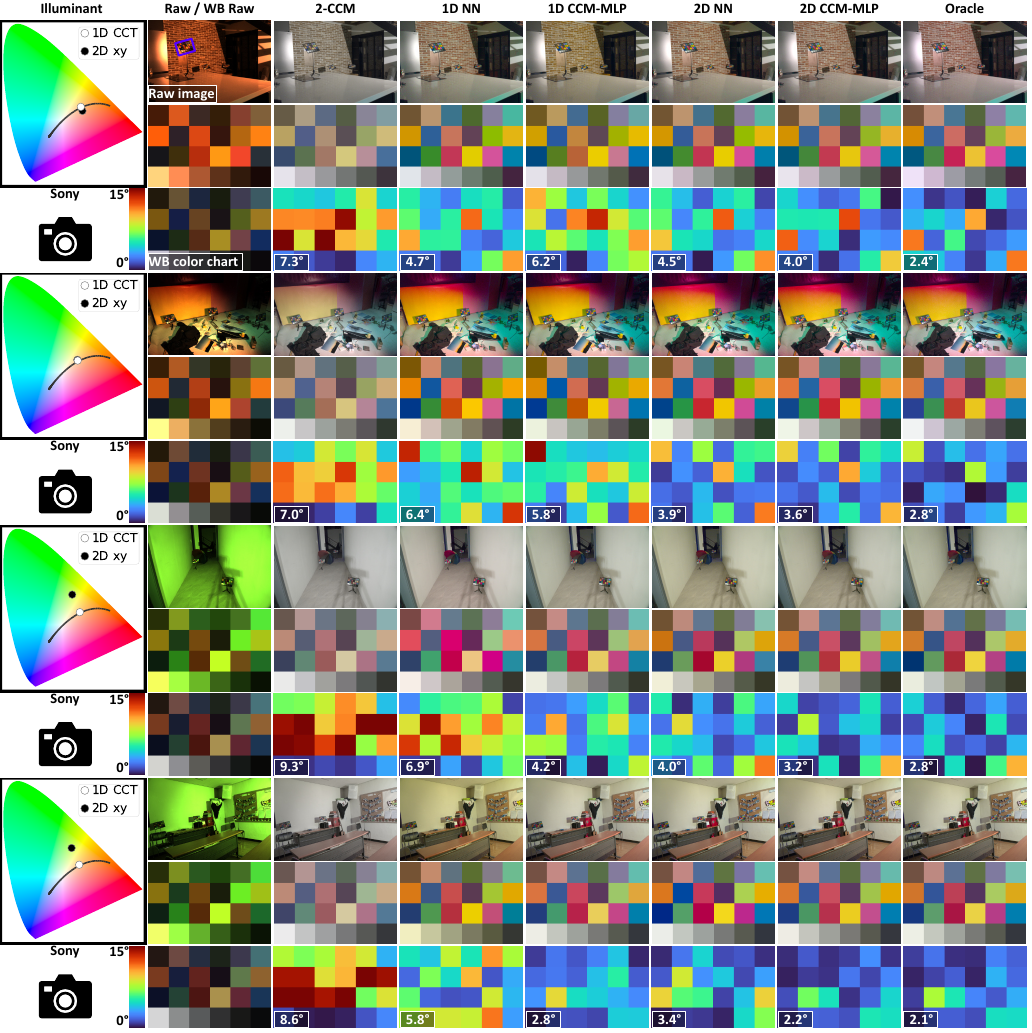}

\caption{\textbf{Colorimetric mapping results on single-illuminant data from LSMI~\cite{lsmi} (Sony camera).} 
While the main paper focused on multi-illuminant scenes due to space constraints, this figure presents a detailed evaluation on single-illuminant data. Each section corresponds to a different illuminant, starting with its CCT and xy chromaticity, and is organized into three rows: The top row displays the full raw-RGB image alongside the display-referred renderings; the middle row shows the raw-RGB color chart and its rendered versions; and the bottom row presents the white-balanced color chart with angular error maps. \textbf{Methods (from left to right):} 2-CCM~\cite{AdobeDNG}, 1D NN~\cite{nearestneighbors}, our 1D CCM-MLP, 2D NN~\cite{nearestneighbors}, our 2D CCM-MLP, and Oracle. \textbf{Analysis:} The visual evidence across these examples reinforces our quantitative results, consistently demonstrating the superior color reproduction of our 2D CCM-MLP.}

\label{fig:lsmi_single_supp2}
\end{figure*}

\begin{figure*}[t]
  \centering
  \includegraphics[width=1.0\linewidth]{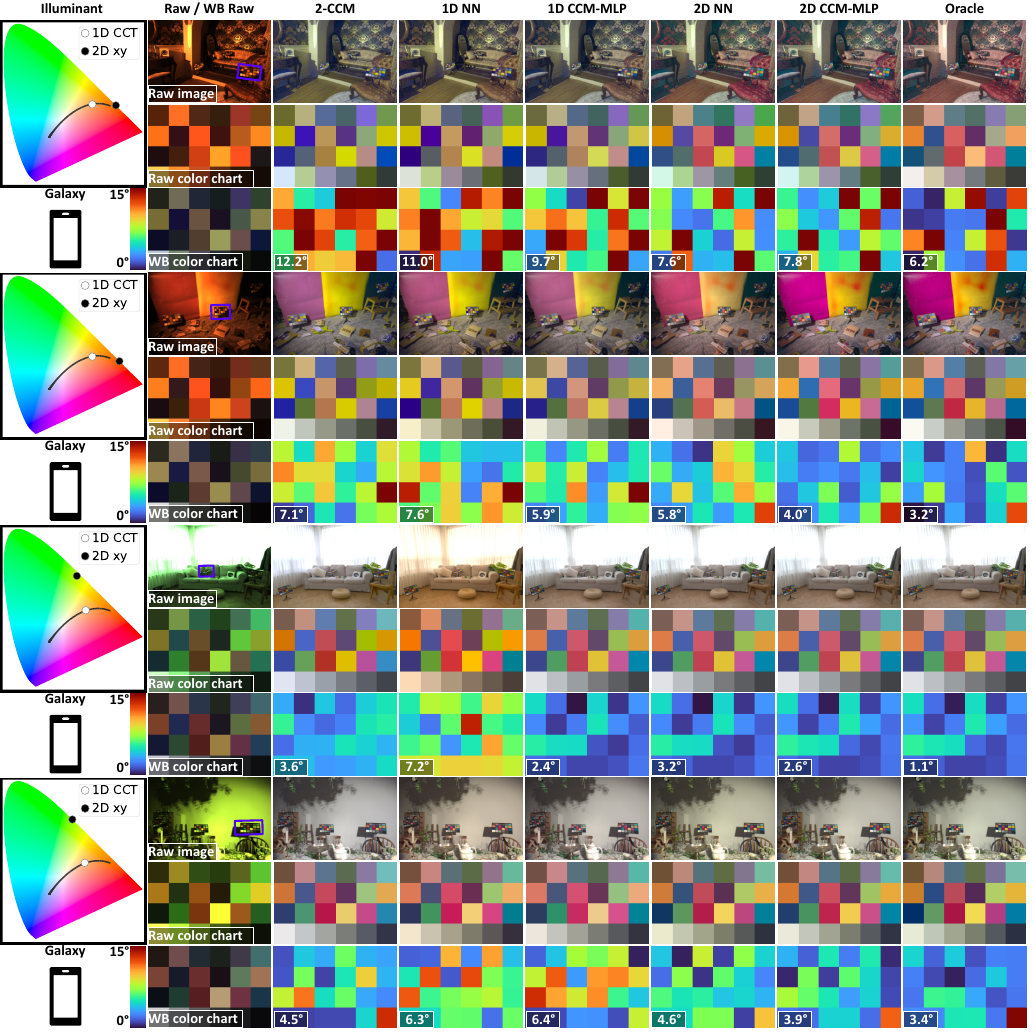}

\caption{\textbf{Colorimetric mapping results on single-illuminant data from LSMI~\cite{lsmi} (Galaxy camera).} To complement the multi-illuminant analysis in the main paper, we now show results on single-illuminant scenes, which were omitted for brevity. Each section corresponds to a different illuminant, starting with its CCT and xy chromaticity, and is organized into three rows. The first row provides full scene context with final renderings. The subsequent rows offer detailed views, with the second focusing on the raw color chart and its renderings, and the third displaying the white-balanced chart and corresponding angular error maps. \textbf{Methods (from left to right):} 2-CCM~\cite{AdobeDNG}, 1D NN~\cite{nearestneighbors}, our 1D CCM-MLP, 2D NN~\cite{nearestneighbors}, our 2D CCM-MLP, and Oracle. \textbf{Analysis:} These results visually affirm the limitations of 1D methods even in less complex scenes, where our 2D approach produces noticeably more accurate color.}

\label{fig:lsmi_single_supp3}
\end{figure*}

\begin{figure*}[t]
  \centering
  \includegraphics[width=1.0\linewidth]{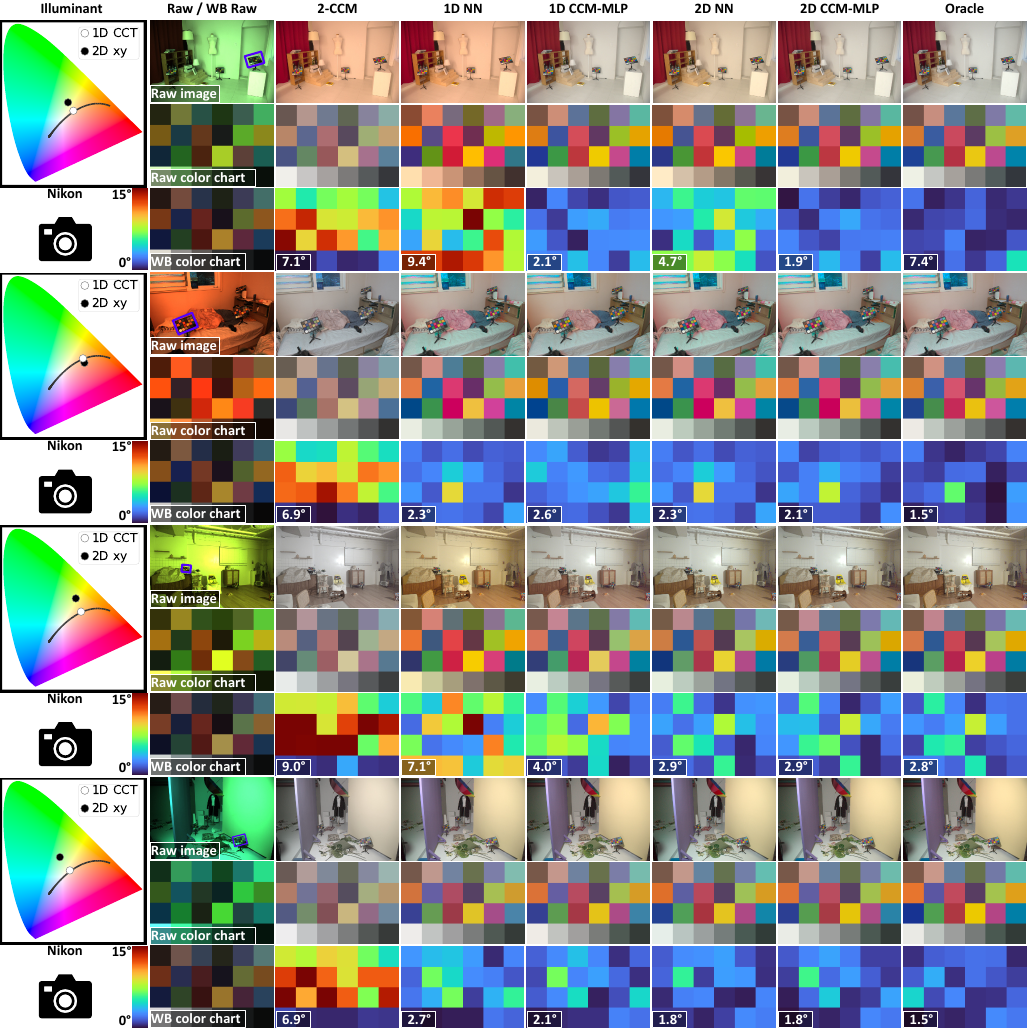}

\caption{\textbf{Colorimetric mapping results on single-illuminant data from LSMI~\cite{lsmi} (Nikon camera).} The main paper prioritized an analysis of challenging multi-illuminant scenes. Here, we provide the corresponding single-illuminant results to offer a complete performance picture. Each section corresponds to a different illuminant, starting with its CCT and xy chromaticity, and is organized into three rows. The rows display: (1) the full raw scene and final outputs, (2) the raw color chart with its display-referred renderings, and (3) the white-balanced chart paired with visualizations of angular error. \textbf{Methods (from left to right):} 2-CCM~\cite{AdobeDNG}, 1D NN~\cite{nearestneighbors}, our 1D CCM-MLP, 2D NN~\cite{nearestneighbors}, our 2D CCM-MLP, and Oracle. \textbf{Analysis:} Across these diverse lighting conditions, the error maps and rendered charts visually corroborate our central claim: our 2D CCM-MLP consistently provides more accurate color corrections than its 1D counterparts. }

\label{fig:lsmi_single_supp1}
\end{figure*}

\begin{figure*}[t]
  \centering
  \includegraphics[width=1.0\linewidth]{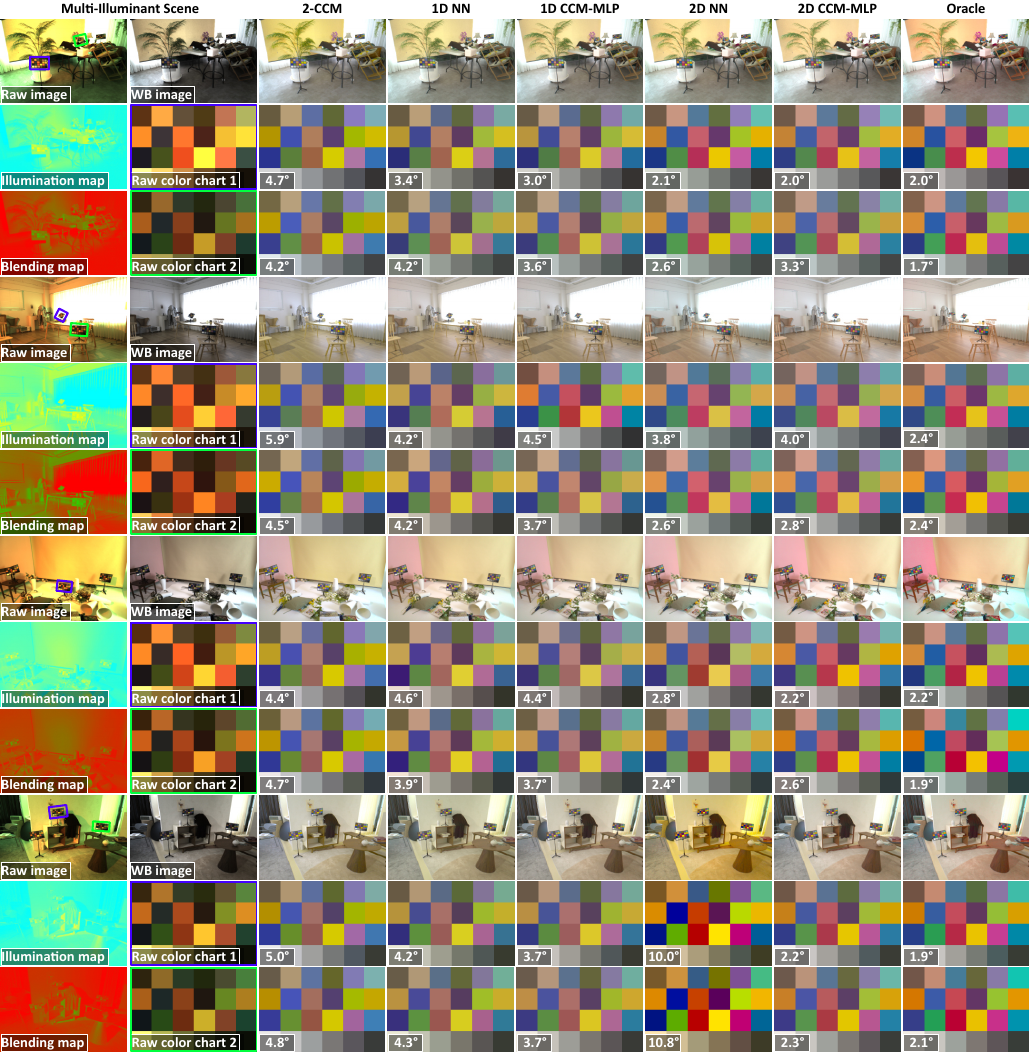}
\caption{\textbf{Colorimetric mapping results on multi-illuminant data from the LSMI dataset (Galaxy camera).} This figure presents an expanded qualitative evaluation, showcasing four spatially-varying multi-illuminant scenes to supplement the two shown in the main paper. \textbf{Layout for each scene:} The top row displays the input raw-RGB image, the white-balanced image, and the final display-referred renderings. The middle row visualizes the contrast-stretched illumination map, the first raw-RGB color chart, and its corrected renderings. The bottom row shows the per-pixel blending map, the second raw-RGB color chart, and its corresponding renderings. \textbf{Methods (from left to right):} 2-CCM~\cite{AdobeDNG}, 1D NN~\cite{nearestneighbors}, our 1D CCM-MLP, 2D NN~\cite{nearestneighbors}, our 2D CCM-MLP, and Oracle. Mean angular errors are provided at the bottom left of each rendered chart. \textbf{Observations:} These additional results further validate our method's effective extension to complex scenes and confirm the consistent performance advantage of 2D chromaticity-based approaches over 1D methods.}

\label{fig:lsmi_mixed_supp1}
\end{figure*}

\begin{figure*}[t]
  \centering
  \includegraphics[width=1.0\linewidth]{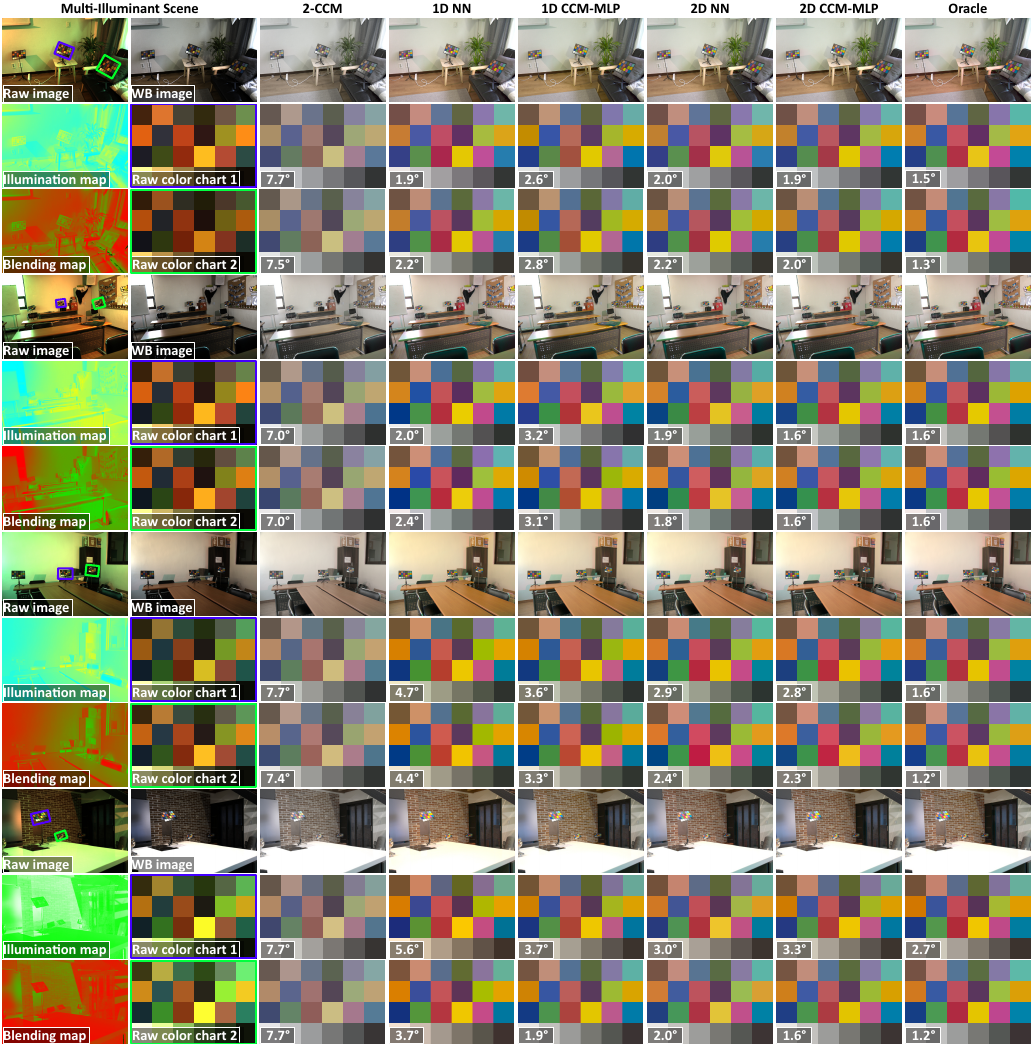}
\caption{\textbf{Colorimetric mapping results on multi-illuminant data from the LSMI dataset (Sony camera).} We provide expanded colorimetric mapping results for the Sony camera, showcasing four multi-illuminant scenes from the LSMI dataset~\cite{lsmi}. These build upon the two examples presented in the main paper. \textbf{Layout for each scene:} Each of the four scenes is detailed across three rows. Row 1: Original raw-RGB input, the white-balanced version, and final rendered outputs. Row 2: The illumination map, the first raw-RGB color chart, and its rendered results. Row 3: The blending map, the second raw-RGB color chart, and its rendered results. \textbf{Methods (from left to right):} 2-CCM~\cite{AdobeDNG}, 1D NN~\cite{nearestneighbors}, our 1D CCM-MLP, 2D NN~\cite{nearestneighbors}, our 2D CCM-MLP, and Oracle. Angular errors are displayed on each rendered color chart. \textbf{Observations:} These comprehensive visualizations reinforce the robustness of our 2D CCM-MLP in handling spatially-varying illumination and its superior accuracy compared to 1D techniques across different camera sensors.}
\label{fig:lsmi_mixed_supp2}
\end{figure*}

\begin{figure*}[t]
  \centering
  \includegraphics[width=1.0\linewidth]{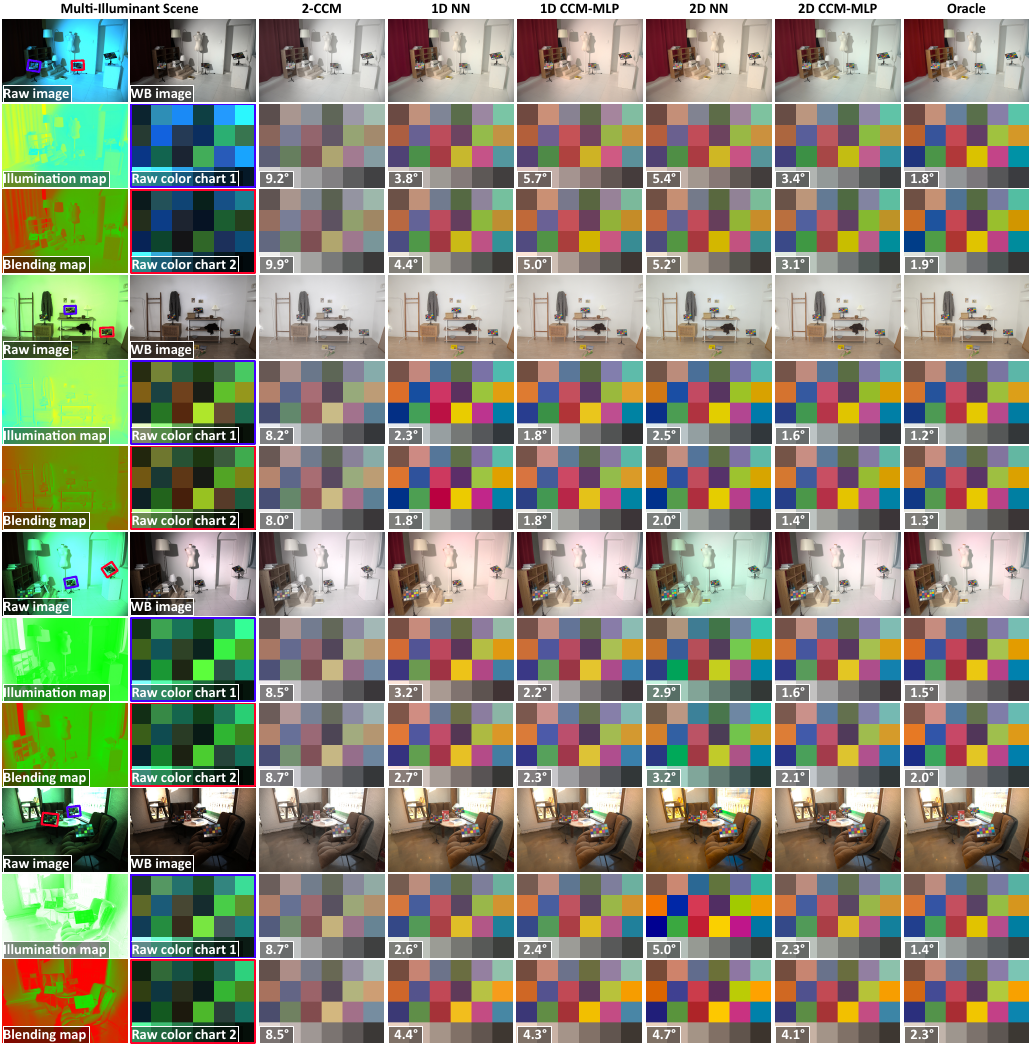}
\caption{\textbf{Colorimetric mapping results on multi-illuminant data from the LSMI dataset (Nikon camera).} This figure offers further qualitative validation with four challenging multi-illuminant scenes from the LSMI dataset~\cite{lsmi}, providing a more comprehensive view than the two scenes in the main paper. \textbf{Layout for each scene:} For each of the four scenes, the visualization is organized as follows: (top row) the input raw image, the result after white balancing, and final corrected outputs from all methods; (middle row) the illumination map and its corrected renderings of the first color chart; (bottom row) the blending map and its corresponding renderings of the second color chart. \textbf{Methods (from left to right):} 2-CCM~\cite{AdobeDNG}, 1D NN~\cite{nearestneighbors}, our 1D CCM-MLP, 2D NN~\cite{nearestneighbors}, our 2D CCM-MLP, and Oracle. Quantitative angular errors are noted on each rendered chart. \textbf{Observations:} As with other cameras, these extensive results for the Nikon sensor underscore the capability of our 2D CCM-MLP to manage complex, mixed-lighting conditions, outperforming methods that rely on a single-dimensional illuminant representation.}
\label{fig:lsmi_mixed_supp3}
\end{figure*}

\section{Implementation Details}
We include additional details for our implementation of the Oracle method with a least-squares solver, and CCM-MLP with TinyCUDA Neural Networks (tinycadann)~\cite{tinycudann1,tinycudann2}.
\paragraph{Oracle.}
Our Oracle baseline uses the L-BFGS-B least-squares solver~\cite{L-BFGS-B}. The solver is optimized to learn a $3\times3$ transformation by minimizing a cosine loss. We consider this solver our ``Oracle'', as it achieves best-case results for a  $3\times3$ mapping.

\paragraph{TinyCUDA Neural Networks.}
TinyCUDA Neural Networks (tinycudann) is a lightweight framework designed for training highly efficient neural networks~\cite{tinycudann1,tinycudann2}. We leverage this framework to train our CCM-MLP, enabling fast, memory-efficient optimization while maintaining high performance on color correction tasks.

\end{document}